\pdfoutput=1

\documentclass{article}
\usepackage[final, nonatbib]{neurips_2019} % produce camera ready copy
\usepackage{paper}
\author{
	Jiefeng Chen\,\footnotemark[1]\ \ $^{1}$\hspace{4mm}
	Xi Wu\,\thanks{\footnotesize Equal contribution.}\ \ $^{2}$\hspace{4mm}
	Vaibhav Rastogi\,\thanks{\footnotesize Work done while at UW-Madison.\vspace{2mm}} $^{2}$ \hspace{4mm}
	Yingyu Liang\,$^{1}$ \hspace{4mm}
	Somesh Jha\,$^{1, 3}$ \\
	\vspace{1mm}
	$^{1}$\,University of Wisconsin-Madison \hspace{5mm}
	$^{2}$\,Google \hspace{5mm}
	$^{3}$\,XaiPient \\
}

\newcommand{\change}[1]{{#1}}
\newcommand\blfootnote[1]{%
  \begingroup
  \renewcommand\thefootnote{}\footnote{#1}%
  \addtocounter{footnote}{-1}%
  \endgroup
}

\title{ Robust Attribution Regularization }

\begin{document}
\maketitle

\begin{abstract}
  An emerging problem\blfootnote{
    Due to lack of space and for completeness,
    we put some definitions (such as coupling) to Section~\ref{sec:additional-defs}.
    Code for this paper is publicly available at the following repository:
    \url{https://github.com/jfc43/robust-attribution-regularization}
  } in trustworthy machine learning is to train models
  that produce robust interpretations for their predictions.
  We take a step towards solving this problem through the lens of axiomatic attribution
  of neural networks. Our theory is grounded in the recent work,
  \emph{Integrated Gradients} ($\IG$)~\cite{STY17}, in
  \emph{axiomatically attributing} a neural network's \emph{output change} to
  \emph{its input change}. We propose training objectives in classic
  robust optimization models to achieve robust $\IG$ attributions.
  Our objectives give principled generalizations of previous objectives
  designed for robust predictions, and they naturally degenerate to classic
  soft-margin training for one-layer neural networks. We also generalize
  previous theory and prove that the objectives for different robust
  optimization models are closely related. Experiments demonstrate the
  effectiveness of our method, and also point to intriguing problems which
  hint at the need for better optimization techniques or better neural network
  architectures for robust attribution training.
\end{abstract}
%%% Local Variables:
%%% mode: latex
%%% TeX-master: t
%%% End:

\section{Introduction}
\label{sec:introduction}
Trustworthy machine learning has received considerable attention
in recent years. An emerging problem to tackle in this domain is to train
models that produce reliable interpretations for their predictions.
For example, a pathology prediction model may predict certain images
as containing malignant tumor. Then one would hope that under visually
indistinguishable perturbations of an image, similar sections of the image,
instead of entirely different ones, can account for the prediction.
However, as Ghorbani, Abid, and Zou~\cite{GAZ17} convincingly demonstrated,
for existing models, one can generate minimal perturbations that substantially
change model interpretations, \emph{while keeping their predictions intact}.
Unfortunately, while the \emph{robust prediction} problem of machine learning
models is well known and has been extensively studied in recent years
(for example, \cite{MMSTV17,SND18,WK18}, and also the tutorial
by Madry and Kolter~\cite{KM-tutorial}), there has only been limited progress
on the problem of \emph{robust interpretations}.

\begin{figure}[htb]
  \centering 
  \begin{minipage}{\linewidth}
    \centering
    NATURAL \hspace{2.5cm} IG-NORM \hspace{2.5cm} IG-SUM-NORM
  \end{minipage} 
  \begin{minipage}{\textwidth}
    \centering
    \begin{subfigure}[b]{.32\textwidth}
      \centering
      \includegraphics[width=0.48\linewidth,bb=0 0 449 464]{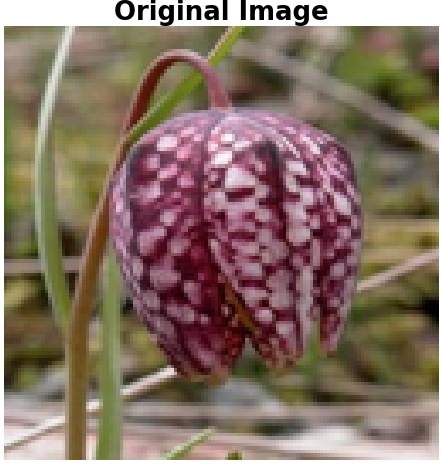} 
      \includegraphics[width=0.48\linewidth,bb=0 0 449 464]{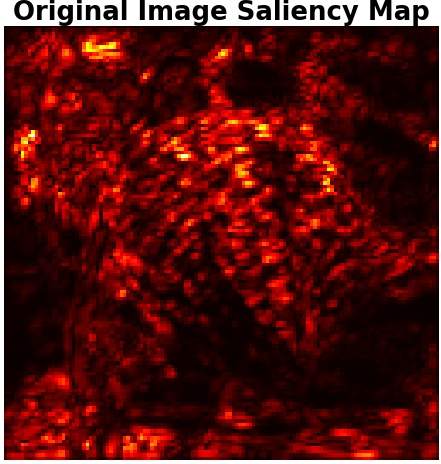} \\
      \includegraphics[width=0.48\linewidth,bb=0 0 449 464]{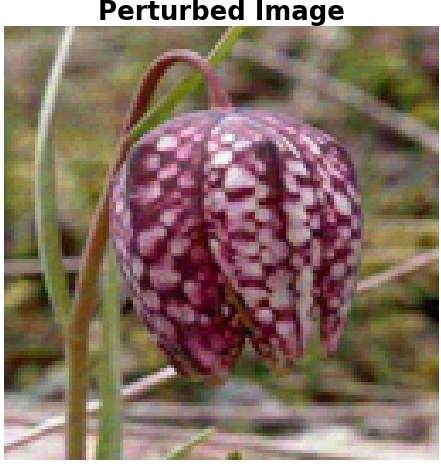} 
      \includegraphics[width=0.48\linewidth,bb=0 0 449 464]{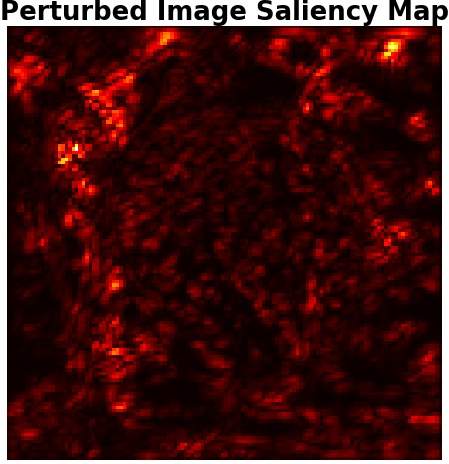} 
      \captionsetup{justification=centering}
      \caption*{Top-1000 Intersection: 0.1\% \\  Kendall's Correlation: 0.2607}
    \end{subfigure}
    \begin{subfigure}[b]{.32\textwidth}
      \centering
      \includegraphics[width=0.48\linewidth,bb=0 0 449 464]{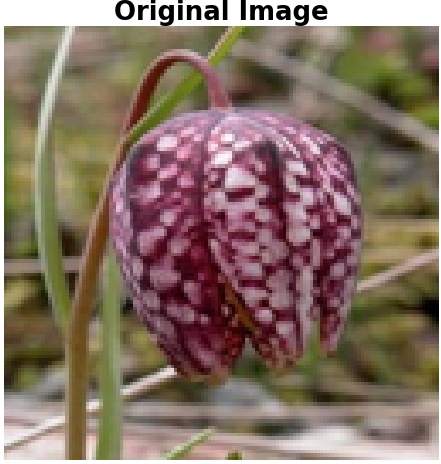} 
      \includegraphics[width=0.48\linewidth,bb=0 0 449 464]{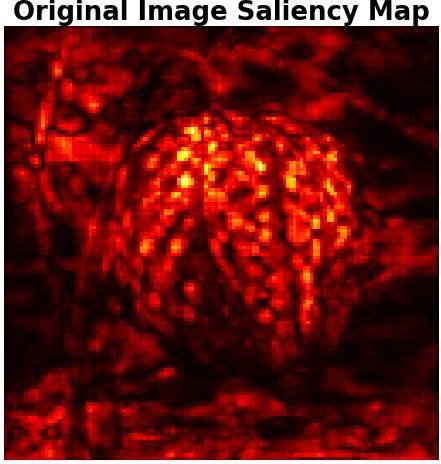} \\
      \includegraphics[width=0.48\linewidth,bb=0 0 449 464]{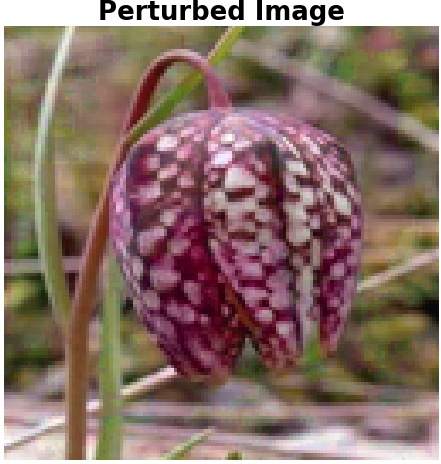} 
      \includegraphics[width=0.48\linewidth,bb=0 0 449 464]{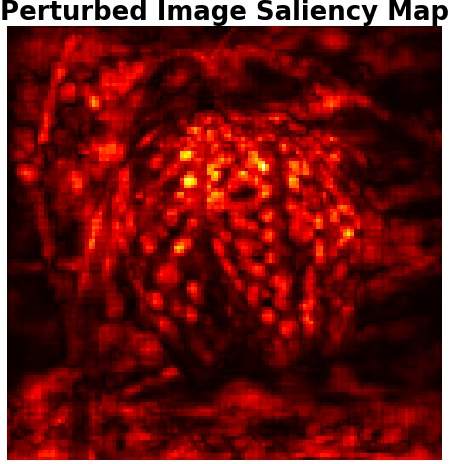} 
      \captionsetup{justification=centering}
      \caption*{Top-1000 Intersection: 58.8\% \\ Kendall's Correlation: 0.6736}
    \end{subfigure}
    \begin{subfigure}[b]{.32\textwidth}
      \centering
      \includegraphics[width=0.48\linewidth,bb=0 0 449 464]{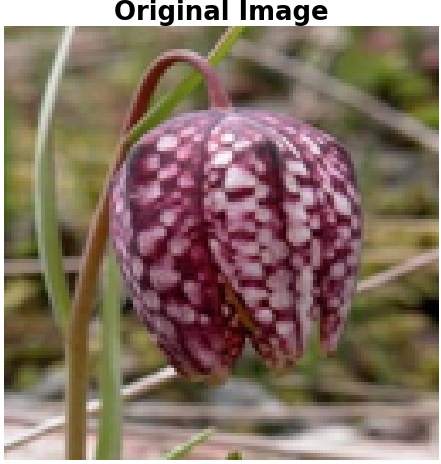} 
      \includegraphics[width=0.48\linewidth,bb=0 0 449 464]{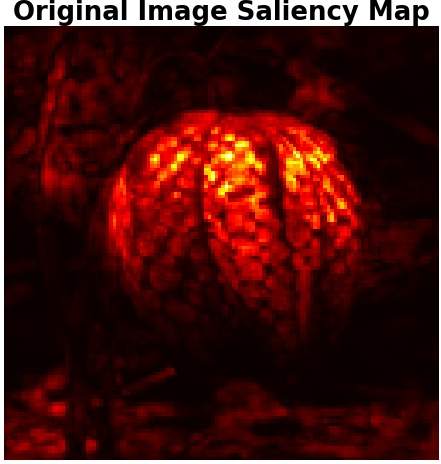} \\
      \includegraphics[width=0.48\linewidth,bb=0 0 449 464]{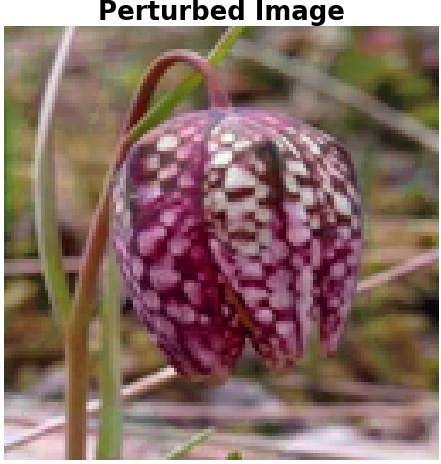} 
      \includegraphics[width=0.48\linewidth,bb=0 0 449 464]{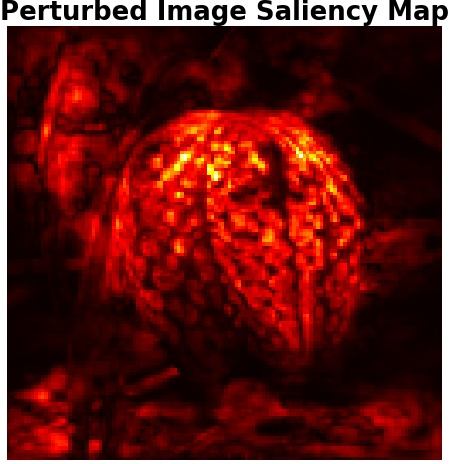} 
      \captionsetup{justification=centering}
      \caption*{Top-1000 Intersection: 60.1\% \\ Kendall's Correlation: 0.6951}
    \end{subfigure}
  \end{minipage}
  \caption{\textbf{Attribution robustness comparing different models.}
    Top-1000 Intersection and Kendall's Correlation are rank correlations
    between original and perturbed saliency maps.
    NATURAL is the naturally trained model,
    IG-NORM and IG-SUM-NORM are models trained using our robust attribution method.
    We use attribution attacks described in~\cite{GAZ17} to perturb the attributions
    while keeping predictions intact. For all images,
    the models give \emph{correct} prediction --  Windflower.
    However, the saliency maps (also called feature importance maps),
    computed via IG, show that attributions of the naturally trained model
    are very fragile, either visually or quantitatively as measured by correlation analyses,
    while models trained using our method are much more robust in their attributions.}
  \label{fig:ex}
\end{figure}

In this paper we take a step towards solving this problem
by viewing it through the lens of axiomatic attribution of neural networks,
and propose Robust Attribution Regularization.
Our theory is grounded in the recent work,
\emph{Integrated Gradients} ($\IG$)~\cite{STY17},
in \emph{axiomatically attributing} a neural network's \emph{output change}
to \emph{its input change}. Specifically, given a model $f$,
two input vectors $\bfx, \bfx'$, and an input coordinate $i$,
$\IG^f_{i}(\bfx, \bfx')$ defines a path integration (parameterized by a curve
from $\bfx$ to $\bfx'$) that assigns a number to the $i$-th input as
its ``contribution'' to the change of the model's output from $f(\bfx)$
to $f(\bfx')$. $\IG$ enjoys several natural theoretical properties (such as
the Axiom of Completeness\footnote{Axiom of Completeness says that summing
  up attributions of all components should give $f(\bfx') - f(\bfx)$.})
that other related methods violate.

We briefly overview our approach. Given a loss function $\ell$ and
a data generating distribution $P$, our Robust Attribution Regularization
objective contains two parts: {\bf (1)} Achieving a small loss over
the distribution $P$, and {\bf (2)} The $\IG$ attributions of the loss $\ell$
over $P$ are ``close'' to the $\IG$ attributions over $Q$,
if distributions $P$ and $Q$ are close to each other.
We can naturally encode these two goals in two classic robust optimization models:
{\bf (1)} In the \emph{uncertainty set model}~\cite{BGN09-robust-optimization}
where we treat sample points as ``nominal'' points, and assume that true sample
points are from certain vicinity around them, which gives:
\begin{alignat*}{2}
  &\minimize_\theta \Exp_{(\bfx, y) \sim P}[\rho(\bfx, y; \theta)] \\
  &\text{where\ \ }\rho(\bfx, y; \theta) = \ell(\bfx, y; \theta) +
  \lambda\max_{\bfx'\in N(\bfx, \varepsilon)}
  s(\IG^{\ell_y}_{\bfh}(\bfx, \bfx'; r))
\end{alignat*}
where $\IG^{\ell_y}_{\bfh}(\cdot)$ is the attribution w.r.t. neurons
in an intermediate layer $\bfh$, and $s(\cdot)$ is a size function
(e.g., $\|\cdot\|_2$) measuring the size of $\IG$,
and {\bf (2)} In the \emph{distributional robustness} model~\cite{SND18,EK15},
where closeness between $P$ and $Q$ is measured using metrics
such as Wasserstein distance, which gives:
\begin{align*}
  &\minimize_\theta \Exp_P[\ell(P; \theta)] +
    \lambda\sup_{Q; M \in \prod(P, Q)} \Big\{
  \Exp_{Z,Z'}[d_{\IG}(Z, Z')]
    \text{ s.t. } \Exp_{Z,Z'}[c(Z, Z')] \le \rho \Big\},
\end{align*}

In this formulation, $\prod(P, Q)$ is the set of couplings of $P$ and $Q$,
and $M=(Z, Z')$ is one coupling. $c(\cdot, \cdot)$ is a metric,
such as $\|\cdot\|_2$, to measure the cost of an adversary perturbing $z$ to $z'$.
$\rho$ is an upper bound on the expected perturbation cost,
thus constraining $P$ and $Q$ to be ``close'' with each together.
$d_{\IG}$ is a metric to measure the change of attributions from $Z$ to $Z'$,
where we want a large $d_{\IG}$-change under a small $c$-change.
The supremum is taken over $Q$ and $\prod(P, Q)$.

We provide theoretical characterizations of our objectives.
First, we show that they give principled generalizations of previous objectives
designed for \emph{robust predictions}. Specifically, under \emph{weak}
instantiations of size function $s(\cdot)$, and how we estimate $\IG$
computationally, we can leverage axioms satisfied by $\IG$ to recover
the robust prediction objective of~\cite{MMSTV17},
the input gradient regularization objective of~\cite{RDV18},
and also the distributional robust prediction objective of~\cite{SND18}.
These results provide theoretical evidence that
robust prediction training can provide some control over robust interpretations.
Second, for one-layer neural networks, we prove that
instantiating $s(\cdot)$ as 1-norm coincides with the instantiation of
$s(\cdot)$ as $\sf sum$, and further coincides with classic soft-margin training,
which implies  that for generalized linear classifiers,
soft-margin training will robustify both predictions and interpretations.
Finally, we generalize previous theory on
distributional robust prediction~\cite{SND18} to our objectives,
and show that they are closely related.

Through detailed experiments we study the effect of our method
in robustifying attributions.
On MNIST, Fashion-MNIST, GTSRB and Flower datasets, we report encouraging improvement in attribution robustness.
Compared with naturally trained models, we show significantly improved
attribution robustness, as well as prediction robustness.
Compared with Madry et al.'s model~\cite{MMSTV17} trained for robust predictions,
we demonstrate \emph{comparable} prediction robustness (\emph{sometimes even better}),
while \emph{consistently} improving attribution robustness.
We observe that even when our training stops, the attribution regularization term
remains much more significant compared to the natural loss term.
We discuss this problem and point out that current optimization techniques
may not have effectively optimized our objectives.
These results hint at the need for better optimization techniques or
new neural network architectures that are more amenable to
robust attribution training.

The rest of the paper is organized as follows:
Section~\ref{sec:preliminaries} briefly reviews necessary background.
Section~\ref{sec:robust-attribution-regularization} presents our framework
for robustifying attributions, and proves theoretical characterizations.
Section~\ref{sec:optimization} presents instantiations of our method
and their optimization,
and we report experimental results in Section~\ref{sec:experiments}.
Finally, Section~\ref{sec:conclusion} concludes with a discussion on
\change{future directions}.

%%% Local Variables:
%%% mode: latex
%%% TeX-master: "make"
%%% End:

\section{Preliminaries}
\label{sec:preliminaries}
\noindent\textbf{Axiomatic attribution and Integrated Gradients}
Let $f:\Real^d \mapsto \Real$ be a real-valued function, and $\bfx$ and $\bfx'$ be two
input vectors. Given that function values changes from $f(\bfx)$ to $f(\bfx')$, a basic
question is:
\emph{``How to attribute the function value change to the input variables?''}
A recent work by Sundararajan, Taly and Yan~\cite{STY17} provides an
\emph{axiomatic answer} to this question. Formally, let $r: [0, 1] \mapsto \Real^d$ be a
curve such that $r(0) = \bfx$, and $r(1) = \bfx'$, Integrated Gradients ($\IG$) for input
variable $i$ is defined as the following integral:
\begin{align}
  \label{eq:IG}
  \IG^f_i(\bfx, \bfx'; r) = \int_0^1\frac{\partial f(r(t))}{\partial\bfx_i}r'_i(t)dt,
\end{align}
which formalizes the contribution of the $i$-th variable as the integration of
the $i$-th partial as we move along curve $r$. Let $\IG^f(\bfx, \bfx'; r)$ be
the vector where the $i$-th component is $\IG^f_i$, then $\IG^f$ satisfies some
natural axioms. For example, the Axiom of Completeness says that summing all coordinates
gives the change of function value:
${\sf sum}(\IG^f(\bfx, \bfx'; r))
=\sum_{i=1}^d\IG^f_i(\bfx, \bfx'; r)
=f(\bfx') - f(\bfx)$.
We refer readers to the paper~\cite{STY17} for other axioms $\IG$ satisfies.

\noindent\textbf{Integrated Gradients for an intermediate layer.}
We can generalize the theory of $\IG$ to an intermediate layer of neurons.
The key insight is to leverage the fact that Integrated Gradients is a
\emph{curve integration}. Therefore, given some hidden layer
$\bfh = [h_1, \dots, h_l]$, computed by a function $h(\bfx)$ induced by
previous layers, one can then naturally view the previous layers as inducing a
\emph{curve} $h \circ r$ which moves from $h(\bfx)$ to $h(\bfx')$, as we move from
$\bfx$ to $\bfx'$ along curve $r$. Viewed this way, we can thus naturally compute
$\IG$ for $\bfh$ in a way that leverages all layers of the network\footnote{
  Proofs are deferred to~\ref{sec:IG-intermediate}.},
\begin{lemma}
  Under curve $r: [0,1] \mapsto \Real^d$ such that $r(0) = \bfx$ and $r(1)=\bfx'$
  for moving $\bfx$ to $\bfx'$, and the function induced by layers before $\bfh$,
  the attribution for $h_i$ for a differentiable $f$ is
  \begin{align}
    \label{eq:intermediate-attribution-integral}
    \IG^f_{h_i}(\bfx, \bfx') = \sum_{j=1}^d \left\{
    \int_0^1\frac{\partial f(h(r(t)))}{\partial h_i}
    \frac{\partial h_i(r(t))}{\partial \bfx_j}r'_j(t)dt\right\}.
  \end{align}
  The corresponding summation approximation is:
  \begin{align}
    \label{eq:intermediate-attribution-sum-approx}
    \IG^f_{h_i}(\bfx, \bfx') =
    \frac{1}{m}\sum_{j=1}^d
    \left\{
    \sum_{k=0}^{m-1}
    \frac{\partial f(h(r(k/m)))}{\partial h_i}
    \frac{\partial h_i(r(k/m))}{\partial \bfx_j}
    r'_j(k/m)
    \right\}
  \end{align}
\end{lemma}
%%% Local Variables:
%%% mode: latex
%%% TeX-master: "make"
%%% End:

\section{Robust Attribution Regularization}
\label{sec:robust-attribution-regularization}
In this section we propose objectives for achieving robust attribution,
and study their connections with existing robust training objectives.
At a high level, given a loss function $\ell$ and a data generating
distribution $P$, our objectives contain two parts:
(1) Achieving a small loss over the data generating distribution $P$,
and (2) The $\IG$ attributions of the loss $\ell$ over $P$ are ``close'' to
the $\IG$ attributions over distribution $Q$, if $P$ and $Q$ are close to each other.
We can naturally encode these two goals in existing robust optimization models.
Below we do so for two popular models: the \emph{uncertainty set model} and
the \emph{distributional robustness} model.
\vspace{-2mm}
\subsection{Uncertainty Set Model}
\label{sec:uncertainty-set-model}
In the uncertainty set model, for any sample $(\bfx, y) \sim P$ for
a data generating distribution $P$, we think of it as a ``nominal'' point and
assume that the real sample comes from a neighborhood around $\bfx$.
In this case, given any intermediate layer $\bfh$, we propose the following
objective function:
\begin{align}
  \label{eq:rar-objective}
  \begin{split}
      &\minimize_\theta \Exp_{(\bfx, y) \sim P}[\rho(\bfx, y; \theta)] \\
      &\text{where\ \ }\rho(\bfx, y; \theta) = \ell(\bfx, y; \theta) +
      \lambda\max_{\bfx'\in N(\bfx, \varepsilon)}
      s(\IG^{\ell_y}_{\bfh}(\bfx, \bfx'; r))
  \end{split}
\end{align}
where $\lambda \ge 0$ is a regularization parameter, $\ell_y$ is the
loss function with label $y$ fixed:
$\ell_y(\bfx; \theta) = \ell(\bfx, y; \theta)$,
$r: [0, 1] \mapsto \Real^d$ is a curve parameterization from $\bfx$ to $\bfx'$,
and $\IG^{\ell_y}$ is the integrated gradients of $\ell_y$, and therefore gives
attribution of changes of $\ell_y$ as we go from $\bfx$ to $\bfx'$. $s(\cdot)$
is a size function that measures the ``size'' of the attribution.\footnote{
  We stress that this regularization term depends on model parameters $\theta$
  through loss function $\ell_y$.}

We now study some particular instantiations of the objective
(\ref{eq:rar-objective}). Specifically, we recover existing robust training
objectives under \emph{weak} instantiations (such as choosing $s(\cdot)$ as
summation function, which is not metric, or use crude approximation of $\IG$),
and also derive new instantiations that are natural extensions to existing ones.

\begin{proposition}[\textbf{Madry et al.'s robust prediction objective}]
  \label{prop:recover-madry}
  If we set $\lambda=1$  , and let $s(\cdot)$ be the $\sf sum$ function
  (sum all components of a vector), then for any curve $r$ and any intermediate
  layer $\bfh$, (\ref{eq:rar-objective}) is exactly the objective proposed
  by Madry et al.~\cite{MMSTV17} where
  $\rho(\bfx, y; \theta) = \max_{\bfx' \in N(\bfx, \varepsilon)}\ell(
  \bfx', y; \theta)$.
\end{proposition}
We note that:
(1) $\sf sum$ is a weak size function which does not give a metric.
(2) As a result, while this robust prediction objective falls
within our framework, and regularizes robust attributions,
it allows a small regularization term where attributions actually change
significantly but they cancel each other in summation.
Therefore, the control over robust attributions can be weak.

\begin{proposition}[\textbf{Input gradient regularization}]
  \label{prop:recover-input-gradient-regularization}
  For any $\lambda' > 0$ and $q \ge 1$, if we set
  $\lambda = \lambda'/\varepsilon^q$, $s(\cdot)=\|\cdot\|_1^q$,
  and use only the first term of summation approximation
  (\ref{eq:intermediate-attribution-sum-approx}) to approximate $\IG$,
  then (\ref{eq:rar-objective}) becomes exactly the input gradient
  regularization of Drucker and LeCun~\cite{DL92}, where we have
  $\rho(\bfx, y; \theta)
  = \ell(\bfx, y; \theta)
  + \lambda \|\nabla_{\bfx} \ell(\bfx, y; \theta)\|_q^q$.
\end{proposition}
In the above we have considered instantiations of a weak size function
(summation function), which recovers Madry et al.'s objective,
and of a weak approximation of $\IG$ (picking the first term),
which recovers input gradient regularization. In the next example,
we pick a nontrivial size function, the 1-norm $\|\cdot\|_1$,
use the precise $\IG$, but then we use a \emph{trivial
  intermediate layer}, the output loss $\ell_y$.
  
\begin{proposition}[\textbf{Regularizing by attribution of the loss output}]
  \label{prop:regularize-by-loss-output}
  Let us set $\lambda=1$, $s(\cdot) = \|\cdot\|_1$, and $\bfh = \ell_y$
  (the output layer of loss function!), then we have
  $\rho(\bfx, y; \theta)
  = \ell_y(\bfx) + \max_{\bfx' \in N(\bfx, \varepsilon)}\{
  |\ell_y(\bfx') - \ell_y(\bfx)|\}$.
\end{proposition}
We note that this loss function is a ``surrogate'' loss function for
Madry et al.'s loss function because
$\ell_y(\bfx) + \max_{\bfx' \in N(\bfx, \varepsilon)}\{|\ell_y(\bfx')
- \ell_y(\bfx)|\}
\ge \ell_y(\bfx) + \max_{\bfx' \in N(\bfx, \varepsilon)}\{(\ell_y(\bfx')
- \ell_y(\bfx))\}
= \max_{\bfx' \in N(\bfx, \varepsilon)}\ell_y(\bfx')$.
Therefore, even at such a trivial instantiation,
robust attribution regularization provides interesting guarantees.
\vspace{-2mm}

\subsection{Distributional Robustness Model}
\label{sec:distributional-robustness-model}
A different but popular model for robust optimization is the
distributional robustness model. In this case we consider a family of
distributions $\calP$, each of which is supposed to be a ``slight variation''
of a base distribution $P$. The goal of robust optimization is then
that certain objective functions obtain stable values over this entire family.
Here we apply the same underlying idea to the distributional robustness
model: One should get a small loss value over the base distribution $P$,
and for any distribution $Q \in \calP$, the $\IG$-based \emph{attributions}
change only a little if we move from $P$ to $Q$. This is formalized as:
\begin{align*}
  \minimize_\theta \Exp_P[\ell(P; \theta)]
  + \lambda\sup_{Q \in \calP} \left\{W_{d_{\IG}}(P, Q) \right\},
\end{align*}
where the $W_{d_{\IG}}(P, Q)$ is the Wasserstein distance between
$P$ and $Q$ under a distance metric
$d_{\IG}$.\footnote{
  For supervised learning problem where $P$ is of the form $Z=(X, Y)$,
  we use the same treatment as in~\cite{SND18} so that cost function is
  defined as $c(z, z') = c_x(x, x') + \infty\cdot{\bf 1}\{y \neq y'\}$.
  All our theory carries over to such $c$ which has range
  $\Real_{+} \cup \{\infty\}$.
}
We use $\IG$ to highlight that this metric is related to integrated gradients.

We propose again $d_{\IG}(\bfz, \bfz') = s(\IG^{\ell}_{\bfh}(\bfz, \bfz'))$.
We are particularly interested in the case where $\calP$ is a  Wasserstein ball
around the base distribution $P$, using ``perturbation'' cost metric $c(\cdot)$.
This gives regularization term
$\lambda \Exp_{W_c(P, Q) \le \rho}\sup\{W_{d_{\IG}}(P, Q)\}$.
An unsatisfying aspect of this objective, as one can observe now, is that
$W_{d_{\IG}}$ and $W_c$ can take two \emph{different} couplings,
while intuitively we want to use only one coupling to transport $P$ to $Q$.
For example, this objective allows us to pick a coupling $M_1$ under which
we achieve $W_{d_{\IG}}$
(recall that Wasserstein distance is an infimum over couplings),
and a different coupling $M_2$ under which we achieve $W_c$,
but under $M_1 = (Z, Z')$, $\Exp_{z, z' \sim M_1}[c(z, z')] > \rho$,
violating the constraint. This motivates the following modification:
\begin{align}
  \label{eq:dist-rar-objective}
  \begin{split}
      &\minimize_\theta \Exp_P[\ell(P; \theta)] +
      \lambda\sup_{Q; M \in \prod(P, Q)} \Big\{
      \Exp_{Z,Z'}[d_{\IG}(Z, Z')]
      \text{ s.t. } \Exp_{Z,Z'}[c(Z, Z')] \le \rho \Big\},
  \end{split}
\end{align}
In this formulation, $\prod(P, Q)$ is the set of couplings of $P$ and $Q$,
and $M=(Z, Z')$ is one coupling. $c(\cdot, \cdot)$ is a metric,
such as $\|\cdot\|_2$, to measure the cost of an adversary perturbing $z$ to $z'$.
$\rho$ is an upper bound on the expected perturbation cost,
thus constraining $P$ and $Q$ to be ``close'' with each together.
$d_{\IG}$ is a metric to measure the change of attributions from $Z$ to $Z'$,
where we want a large $d_{\IG}$-change under a small $c$-change.
The supremum is taken over $Q$ and $\prod(P, Q)$.

\begin{proposition}[\textbf{Wasserstein prediction robustness}]
  \label{prop:recover-wasserstein-prediction-robustness-objective}
  Let $s(\cdot)$ be the summation function and $\lambda=1$, then
  for any curve $\gamma$ and any layer $\bfh$, (\ref{eq:dist-rar-objective})
  reduces to $\sup_{Q: W_c(P, Q) \le \rho}\left\{
    \Exp_Q[\ell(Q; \theta)]\right\}$, which is the objective proposed by
  Sinha, Namhoong, and Duchi~\cite{SND18} for \emph{robust predictions}.
\end{proposition}

\noindent\textbf{Lagrange relaxation}.
For any $\gamma \ge 0$, the Lagrange relaxation of
(\ref{eq:dist-rar-objective}) is
\begin{align}
  \label{eq:dist-rar-objective-lagrange}
  \begin{split}
    \minimize_\theta \bigg\{ \Exp_P[\ell(P; \theta)]
    + \lambda \sup_{Q; M \in \prod(P, Q)}\Big\{
    \Exp_{M=(Z, Z')}\big[
    d_{\IG}(Z, Z') - \gamma c(Z, Z')\big]\Big\} \bigg\}
  \end{split}
\end{align}
where the supremum is taken over $Q$ (unconstrained) and all couplings of $P$
and $Q$, and we want to find a coupling under which $\IG$ attributions change a lot,
while the perturbation cost from $P$ to $Q$ with respect to $c$ is small.
Recall that $g: \Real^d \times \Real^d \rightarrow \Real$
is a \emph{normal integrand} if for each $\alpha$, the mapping
$z \rightarrow \{ z' | g(z, z') \le \alpha \}$
is closed-valued and measurable~\cite{rockafellar2009variational}.
%(any continuous function is a normal integrand, so $d_{\IG}$ is since it is a path integration and so it is differentiable and thus continuous)

Our next two theorems generalize the duality theory in~\cite{SND18} to
a much larger, but natural, class of objectives.
\begin{theorem}
  \label{thm:dist-rar-duality2}
  Suppose $c(z,z)=0$ and $d_{\IG}(z,z)=0$ for any $z$,
  and suppose $\gamma c(z, z') - d_{\IG}(z, z')$ is a normal integrand.
  Then, $\sup_{Q; M \in \prod(P, Q)}\{
  \Exp_{M=(Z, Z')}[d^\gamma_{\IG}(Z, Z')]\} =
  \Exp_{z \sim P}[\sup_{z'}\{d^\gamma_{\IG}(z, z')\}].$
Consequently, we have (\ref{eq:dist-rar-objective-lagrange}) to be equal to the following:
  \begin{align}
    \label{eq:dist-rar-objective-lagrange-2}
    \begin{split}
      \minimize_\theta
      \Exp_{z \sim P}\Big[\ell(z; \theta) +
      \lambda\sup_{z'}\{d_{\IG}(z, z') - \gamma c(z, z')\} \Big]
    \end{split}
  \end{align}
\end{theorem}

The assumption $d_{\IG}(z,z)=0$ is true for what we propose,
and $c(z, z) = 0$ is true for any typical cost such as $\ell_p$ distances.
The normal integrand assumption is also very weak,
e.g., it is satisfied when $d_{\IG}$ is continuous and $c$ is closed convex.

Note that (\ref{eq:dist-rar-objective-lagrange-2}) and (\ref{eq:rar-objective}) are very similar,
and so we use (\ref{eq:rar-objective}) for the rest the paper.
Finally, given Theorem~\ref{thm:dist-rar-duality2}, we are also able to connect (\ref{eq:dist-rar-objective})
and (\ref{eq:dist-rar-objective-lagrange-2}) with the following duality result:
\begin{theorem}
  \label{thm:dist-rob-duality}
  Suppose $c(z,z)=0$ and $d_{\IG}(z,z)=0$ for any $z$,
  and suppose $\gamma c(z, z') - d_{\IG}(z, z')$ is a normal integrand.
  For any $\rho > 0$, there exists $\gamma \ge 0$
  such that the optimal solutions of (\ref{eq:dist-rar-objective-lagrange-2})
  are optimal for (\ref{eq:dist-rar-objective}).
\end{theorem}

\subsection{One Layer Neural Networks}
\label{sec:one-layer-neural-networks}
We now consider the special case of one-layer neural networks,
where the loss function takes the form of
$\ell(\bfx, y; \bfw) = g(-y\langle \bfw, \bfx \rangle)$,
$\bfw$ is the model parameters, $\bfx$ is a feature vector, $y$ is a label,
and $g$ is nonnegative. We take $s(\cdot)$ to be $\|\cdot\|_1$, which
corresponds to a strong instantiation that does not allow attributions
to cancel each other. Interestingly, we prove that for natural choices of $g$,
this is however exactly Madry et al.'s objective~\cite{MMSTV17},
which corresponds to $s(\cdot) = {\sf sum}(\cdot)$.
That is, the strong ($s(\cdot)=\|\cdot\|_1$)
and weak instantiations ($s(\cdot)={\sf sum}(\cdot)$) coincide for
one-layer neural networks. This thus says that for generalized linear
classifiers,  ``robust interpretation'' coincides with ``robust predictions,''
and further with classic soft-margin training.

\begin{theorem}
  \label{thm:one-layer-neural-networks}
  Suppose that $g$ is differentiable, non-decreasing, and convex.
  Then for $\lambda=1$, $s(\cdot) = \|\cdot\|_1$,
  and $\ell_\infty$ neighborhood, (\ref{eq:rar-objective})
  reduces to Madry et al.'s objective:
  \begin{align*}
    &\sum_{i=1}^m\max_{\|\bfx'_i - \bfx_i\|_\infty\le \varepsilon}
      g(-y_i\langle \bfw, \bfx'_i \rangle)
    \ \ \text{(Madry et al.'s objective)}\\
    = &\sum_{i=1}^mg(-y_i\langle \bfw, \bfx_i \rangle + \varepsilon\|\bfw\|_1)\ \ \text{(soft-margin)}.
  \end{align*}
\end{theorem}
Natural losses, such as Negative Log-Likelihood and softplus hinge loss,
satisfy the conditions of this theorem.
%%% Local Variables:
%%% mode: latex
%%% TeX-master: "make"
%%% End:

%%% Local Variables:
%%% mode: latex
%%% TeX-master: "make"
%%% End:

\section{Instantiations and Optimizations}
\label{sec:optimization}
In this section we discuss instantiations of (\ref{eq:rar-objective}) and
how to optimize them. We start by presenting two objectives instantiated
from our method: (1) IG-NORM, and (2) IG-SUM-NORM.
Then we discuss how to use gradient descent to optimize these objectives.

\noindent\textbf{IG-NORM}. As our first instantiation, we pick
$s(\cdot) = \|\cdot\|_1$, $\bfh$ to be the input layer, and $r$ to be the
straightline connecting $\bfx$ and $\bfx'$. This gives:
\begin{align*}
  \minimize_\theta \Exp_{(\bfx, y)\sim P}\left[
  \ell(\bfx, y;\theta)
  + \lambda\max_{\bfx'\in N(\bfx, \varepsilon)}\|\IG^{\ell_y}(\bfx,\bfx')\|_1
  \right]
\end{align*}

\noindent\textbf{IG-SUM-NORM}. In the second instantiation we combine the sum
size function and norm size function, and define
$s(\cdot) = {\sf sum}(\cdot) + \beta\|\cdot\|_1$.
Where $\beta \ge 0$ is a regularization parameter.
Now with the same $\bfh$ and $r$ as above, and put $\lambda=1$,
then our method simplifies to:
\begin{align*}
  \minimize_\theta \Exp_{(\bfx, y)\sim P}\left[
  \max_{\bfx'\in N(\bfx, \varepsilon)}\left\{\ell(\bfx', y;\theta)
  + \beta\|\IG^{\ell_y}(\bfx,\bfx')\|_1\right\}
  \right]
\end{align*}
which can be viewed as appending an extra robust IG term to $\ell(\bfx')$.

\noindent\textbf{Gradient descent optimization}.
We propose the following gradient descent framework to optimize the objectives.
The framework is parameterized by an adversary $\calA$ which is supposed to
solve the inner max by finding a point $\bfx^\star$ which changes attribution
significantly. Specifically, given a point $(\bfx, y)$ at time step $t$
during SGD training, we have the following two steps (this can be easily
generalized to mini-batches):

\noindent\textbf{\em Attack step}.
We run $\calA$ on $(\bfx, y)$ to find $\bfx^\star$ that produces
a large inner max term (that is $\|\IG^{\ell_y}(\bfx, \bfx^\star)\|_1$
for IG-NORM, and
$\ell(\bfx^\star) + \beta\|\IG^{\ell_y}(\bfx, \bfx^\star)\|_1$
for IG-SUM-NORM.

\noindent\textbf{\em Gradient step}.
Fixing $\bfx^\star$, we can then compute the gradient of the corresponding
objective with respect to $\theta$, and then update the model.

\noindent\textbf{Important objective parameters.}
In both attack and gradient steps, we need to differentiate $\IG$
(in attack step, $\theta$ is fixed and we differentiate w.r.t. $\bfx$,
while in gradient step, this is reversed), and this induces a set of parameters
of the objectives to tune for optimization, which is summarized in
Table~\ref{table:optimization parameters}.
Differentiating summation approximation of $\IG$ amounts to
compute second partial derivatives. We rely on the auto-differentiation
capability of TensorFlow~\cite{abadi2016tensorflow} to compute second derivatives.
\begin{table}[bth]
  \centering
  \begin{tabular}{p{3cm}|p{8cm}}
    \hline
    Adversary $\calA$ & Adversary to find $\bfx^{\star}$. Note that
                        our goal is simply to maximize the inner term
                        in a neighborhood, thus in this paper we choose
                        Projected Gradient Descent for this purpose.\\
    \hline
    $m$ in the attack step & To differentiate $\IG$ in the attack step,
                             we use summation approximation of $\IG$, and this
                             is the number of segments for apprioximation.\\
    \hline
    $m$ in the gradient step & Same as above, but in the gradient step.
                               We have this $m$ separately due to
                               efficiency consideration.\\
    \hline
    $\lambda$ & Regularization parameter for IG-NORM.\\
    \hline
    $\beta$ & Regularization parameter for IG-SUM-NORM.\\
    \hline
  \end{tabular}
  \vspace{1mm}
  \caption{Optimization parameters.}
  \label{table:optimization parameters}
  \vspace{-10mm}
\end{table}

%%% Local Variables:
%%% mode: latex
%%% TeX-master: "make"
%%% End:

\section{Experiments}
\label{sec:experiments}
We now perform experiments using our method. We ask the following questions:
{\bf (1)} Comparing models trained by our method and naturally trained models at
\emph{test time}, do we maintain the accuracy on unperturbed test inputs?
{\bf (2)} At test time, if we use attribution attacks mentioned in~\cite{GAZ17} to
perturb attributions while keeping predictions intact,
how does the attribution robustness of our models compare with
that of the naturally trained models?
{\bf (3)} Finally, how do we compare attribution robustness of our models
with \emph{weak instantiations} for robust predictions?

To answer these questions, We perform experiments on four classic datasets:
MNIST~\cite{mnist}, Fashion-MNIST~\cite{xiao2017fashion}, GTSRB~\cite{stallkamp2012man},
and Flower~\cite{nilsback2006visual}.
In summary, our findings are the following:
{\bf (1)} Our method results in very small drop in test accuracy
compared with naturally trained models.
{\bf (2)} On the other hand, our method gives signficantly better
attribution robustness, as measured by correlation analyses.
{\bf (3)} Finally, our models yield \emph{comparable} prediction robustness
(sometimes even better), while \emph{consistently} improving attribution robustness.
In the rest of the section we give more details.

\noindent\textbf{Evaluation setup}.
In this work we use $\IG$ to compute attributions (i.e. feature importance map),
which, as demonstrated by~\cite{GAZ17}, is more robust compared to
other related methods (note that, $\IG$ also enjoys other theoretical properties).
To attack attribution while retaining model predictions,
we use Iterative Feature Importance Attacks (IFIA) proposed by~\cite{GAZ17}.
Due to lack of space, we defer details of parameters
and other settings to the appendix.
We use two metrics to measure attribution robustness
(i.e. how similar the attributions are between original and perturbed images):

\textbf{\em Kendall's tau rank order correlation}.
Attribution methods rank all of the features in order of their importance,
we thus use the rank correlation ~\cite{kendall1938new} to compare
similarity between interpretations.

\textbf{\em Top-k intersection}. We compute the size of intersection of the
$k$ most important features before and after perturbation.

Compared with~\cite{GAZ17}, we use Kendall's tau correlation,
instead of Spearman's rank correlation. The reason is that we found that
on the GTSRB and Flower datasets, Spearman's correlation is not consistent
with visual inspection, and often produces too high correlations.
In comparison, Kendall's tau correlation consistently produces lower
correlations and aligns better with visual inspection.
\change{
  Finally, when computing attribution robustness,
  we only consider the test samples that are correctly classified by the model.
}

\begin{figure}[htb]
  \centering
  \begin{subfigure}{0.45\linewidth}
    \centering
    \includegraphics[width=\linewidth,bb=0 0 432 288]{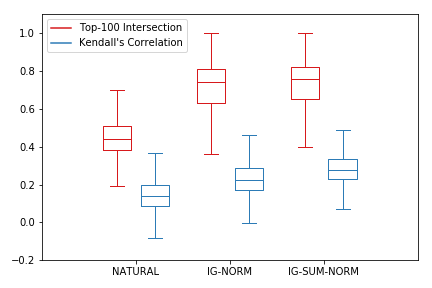}
    \caption{MNIST}
  \end{subfigure}
  \hspace{1cm}
   \begin{subfigure}{0.45\linewidth}
  	\centering
  	\includegraphics[width=\linewidth,bb=0 0 432 288]{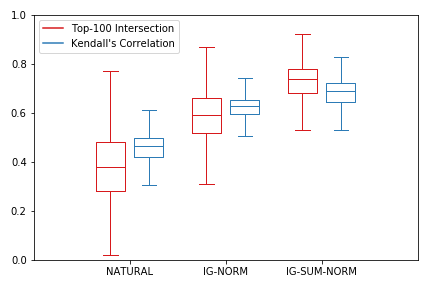}
  	\caption{Fashion-MNIST}
  \end{subfigure}
  
  \begin{subfigure}{0.45\linewidth}
    \centering
    \includegraphics[width=\linewidth,bb=0 0 432 288]{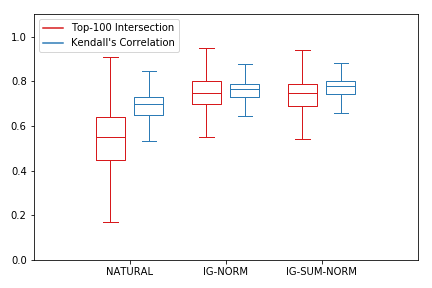}
    \caption{GTSRB}
  \end{subfigure} 
  \hspace{1cm}
  \begin{subfigure}{0.45\linewidth}
    \centering
    \includegraphics[width=\linewidth,bb=0 0 432 288]{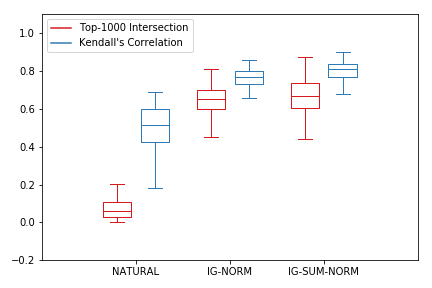}
    \caption{Flower}
  \end{subfigure}
  \caption{Experiment results on MNIST, Fashion-MNIST, GTSRB and Flower.}
  \label{fig:compare-with-natural-models}
  \vspace{-5mm}
\end{figure}

\noindent\textbf{Comparing with natural models}.
\change{
  Figures (a), (b), (c), and (d) in Figure~\ref{fig:compare-with-natural-models}
  show that, compared with naturally trained models,
  robust attribution training gives significant improvements in
  attribution robustness (measured by either median or confidence intervals).
  The exact numbers are recorded in Table~\ref{table:compare-with-madry}:
  Compared with naturally trained models (rows where ``Approach'' is NATURAL),
  robust attribution training has significantly better adversarial accuracy
  and attribution robustness, while having a small drop in natural accuracy (denoted by {\tt Nat Acc.}).
}

\noindent\textbf{Ineffective optimization}.
We observe that even when our training stops,
the attribution regularization term remains much more significant
compared to the natural loss term. For example for IG-NORM,
when training stops on MNIST, $\ell(\bfx)$ typically stays at ~1,
but $\|\IG(\bfx, \bfx')\|_1$ stays at $10 \sim 20$.
This indicates that optimization has not been very effective
in minimizing the regularization term. There are two possible reasons to this:
(1) Because we use summation approximation of $\IG$,
it forces us to compute second derivatives,
which may not be numerically stable for deep networks.
(2) The network architecture may be inherently unsuitable for
robust attributions, rendering the optimization hard to converge.

\noindent\textbf{Comparing with robust prediction models}.
Finally we compare with Madry et al.'s models,
which are trained for robust prediction.
We use {\tt Adv Acc.} to denote adversarial accuracy
(prediction accuracy on perturbed inputs).
Again, {\tt TopK Inter.} denotes the average topK intersection
($K=100$ for MNIST, Fashion-MNIST and GTSRB datasets, $K=1000$ for Flower),
and {\tt Rank Corr.} denotes the average Kendall's rank order correlation.
Table~\ref{table:compare-with-madry} gives the details of the results.
As we can see, our models give comparable adversarial accuracy,
and are sometimes even better (on the Flower dataset). On the other hand,
we are consistently better in terms of attribution robustness.
\begin{table}[htb]
  \centering
  \begin{tabular}{ c | c | c | c | c | c }
    \hline
    Dataset & Approach & {\tt Nat Acc.} & {\tt Adv Acc.} &  {\tt TopK Inter.} & {\tt Rank Corr.} \\ \hline
    \multirow{4}{*}{MNIST} 
            & NATURAL & 99.17\% & 0.00\% & 46.61\% & 0.1758 \\ \cline{2-6}
            & Madry et al. & 98.40\% & 92.47\% & 62.56\% & 0.2422 \\ \cline{2-6}
            & IG-NORM & 98.74\% & 81.43\% & 71.36\% & 0.2841 \\ \cline{2-6}
            & IG-SUM-NORM & 98.34\% & 88.17\% & {\bf 72.45\%} & {\bf 0.3111} \\ \hline \hline
    \multirow{4}{*}{Fashion-MNIST} 
		    & NATURAL & 90.86\% & 0.01\% & 39.01\% & 0.4610 \\ \cline{2-6}
		    & Madry et al. & 85.73\% & 73.01\% & 46.12\% & 0.6251 \\ \cline{2-6}
		    & IG-NORM & 85.13\% & 65.95\% & 59.22\% & 0.6171 \\ \cline{2-6}
		    & IG-SUM-NORM & 85.44\% & 70.26\% & {\bf 72.08\%} & {\bf 0.6747} \\ \hline \hline
    \multirow{4}{*}{GTSRB} 
            & NATURAL & 98.57\% & 21.05\% & 54.16\% & 0.6790 \\ \cline{2-6}
            & Madry et al. & 97.59\% & 83.24\% & 68.85\% & 0.7520 \\ \cline{2-6}
            & IG-NORM & 97.02\% & 75.24\% & {\bf 74.81\%} & 0.7555 \\ \cline{2-6}
            & IG-SUM-NORM & 95.68\%  & 77.12\% & 74.04\% & {\bf 0.7684} \\ \hline \hline
    \multirow{4}{*}{Flower} 
            & NATURAL & 86.76\% & 0.00\% & 8.12\% & 0.4978 \\ \cline{2-6}
            & Madry et al. & 83.82\% & 41.91\% & 55.87\% & 0.7784 \\ \cline{2-6}
            & IG-NORM & 85.29\% & 24.26\% & 64.68\% & 0.7591 \\ \cline{2-6}
            & IG-SUM-NORM & 82.35\% & 47.06\% & {\bf 66.33\%} & {\bf 0.7974} \\ \hline
  \end{tabular}
  \vspace{1mm}
  \caption{\change{Experiment results including prediction accuracy, prediction robustness
    and attribution robustness.}}
  \label{table:compare-with-madry}
  \vspace{-10mm}
\end{table}

%%% Local Variables:
%%% mode: latex
%%% TeX-master: "make"
%%% End:

\section{Discussion and Conclusion}
\label{sec:conclusion}
This paper builds a theory to robustify model interpretations
through the lens of axiomatic attributions of neural networks.
We show that our theory gives principled generalizations of
previous formulations for robust predictions,
and we characterize our objectives for one-layer neural networks.
\change{
  We believe that our work opens many intriguing avenues for future research,
  and we discuss a few topics below.\\
  \noindent\textbf{Why we want robust attributions?} Model attributions are
  \emph{facts} about model behaviors. While robust attribution does not
  necessarily mean that the attribution is correct, a model with
  \emph{brittle attribution} can never be trusted. To this end, it seems
  interesting to examine attribution methods other than Integrated Gradients.\\
  \noindent\textbf{Robust attribution leads to more human-aligned attribution.}
  Note that our proposed training scheme requires both prediction correctness
  and robust attributions, and therefore it encourages to learn \emph{invariant}
  features from data that are also highly predictive. In our experiments, we
  found an intriguing phenomenon that
  \emph{our regularized models produce attributions that are much more
    aligned with human perceptions} (for example, see Figure~\ref{fig:ex}).
  Our results are aligned with the recent work~\cite{TSETM19,engstrom2019adversarial}.\\
  \noindent\textbf{Robust attribution may help tackle spurious correlations.}
  In view of our discussion so far, we think it is plausible that robust
  attribution regularization can help remove spurious correlations because
  intuitively spurious correlations should not be able to be reliably
  attributed to. Future research on this potential connection seems warranted.\\
  \noindent\textbf{Difficulty of optimization.}
  While our experimental results are encouraging,
  we observe that when training stops,   the attribution regularization term
  remains significant (typically around tens to hundreds),
  which indicates ineffective optimization for the objectives.
  To this end, a main problem is network depth, where as depth increases,
  we get very unstable trajectories of gradient descent,
  which seems to be related to the use of \emph{second order information}
  during robust attribution optimization (due to summation approximation,
  we have first order terms in the training objectives).
  Therefore, it is natural to further study better optimization tchniques
  or better architectures for robust attribution training.
}
%%% Local Variables:
%%% mode: latex
%%% TeX-master: "make"
%%% End:

\section{Acknowledgments}
\label{sec:acknowledgments}
This work is partially supported by Air Force Grant FA9550-18-1-0166, the National Science Foundation (NSF) Grants CCF-FMitF-1836978,
SaTC-Frontiers-1804648 and CCF-1652140 and ARO grant number W911NF-17-1-0405.

\newpage

\bibliographystyle{alpha}
\bibliography{paper}

\newpage
\appendix

\section{Code}
\label{sec:code}
Code for this paper is publicly available at the following repository:\\
\url{https://github.com/jfc43/robust-attribution-regularization}

\section{Proofs}
\label{sec:proofs}

\subsection{Additional definitions}
\label{sec:additional-defs}
Let $P, Q$ be two distributions, a coupling $M = (Z, Z')$ is a joint distribution, where,
if we marginalize $M$ to the first component, $Z$, it is identically distributed as $P$,
and if we marginalize $M$ to the second component, $Z'$, it is identically distributed as $Q$.
Let $\prod(P, Q)$ be the set of all couplings of $P$ and $Q$,
and let $c(\cdot, \cdot)$ be a ``cost'' function that maps $(z, z')$ to a real value.
Wasserstein distance between $P$ and $Q$ w.r.t. $c$ is defined as
$$W_c(P, Q) = \inf_{M \in \prod(P, Q)}\left\{
  \Exp_{(z, z') \sim M}\left[c(z, z')\right]\right\}.$$
Intuitively, this is to find the ``best transportation plan'' (a coupling $M$)
to minimize the expected transportation cost
(transporting $z$ to $z'$ where the cost is $c(z, z')$).

\subsection{Integrated Gradients for an Intermediate Layer}
\label{sec:IG-intermediate}
In this section we show how to compute Integrated Gradients for an intermediate layer of
a neural network. Let $h: \Real^d \mapsto \Real^k$ be a function that computes a hidden
layer of a neural network, where we map a $d$-dimensional input vector to a
$k$-dimensional output vector. Given two points $\bfx$ and $\bfx'$ for computing
attribution, again we consider a parameterization
(which is a mapping $r: \Real \mapsto \Real^d$) such that $r(0) = \bfx$,
and $r(1) = \bfx'$.

The key insight is to leverage the fact that Integrated Gradients is a
\emph{curve integration}. Therefore, given some hidden layer, one can then
naturally view the previous layers as inducing a \emph{curve} $h \circ r$
which moves from $h(\bfx)$ to $h(\bfx')$, as we move from $\bfx$ to $\bfx'$
along curve $r$. Viewed this way, we can thus naturally compute $\IG$
for $\bfh$ in a way that leverages all layers of the network.
Specifically, consider another curve $\gamma(t): \Real \mapsto \Real^k$,
defined as $\gamma(t) = h(r(t))$, to compute a curve integral.
By definition we have $f(\bfx) = g(h(\bfx))$
\begin{align*}
  f(\bfx') - f(\bfx)
  &= g(h(\bfx')) - g(h(\bfx)) \\
  &= g(\gamma(1)) - g(\gamma(0)) \\
  &= \int_0^1\sum_{i=1}^k\frac{\partial f(\gamma(t))}{\partial h_i}\gamma_i'(t)dt \\
  &= \sum_{i=1}^k\int_0^1\frac{\partial f(\gamma(t))}{\partial h_i}\gamma_i'(t)dt
\end{align*}
Therefore we can define the attribution of $h_i$ naturally as
\begin{align*}
  \IG^f_{h_i}(\bfx, \bfx')
  = \int_0^1\frac{\partial f(\gamma(t))}{\partial h_i}\gamma_i'(t)dt
\end{align*}
Let's unpack this a little more:
\begin{align*}
  \int_0^1\frac{\partial f(\gamma(t))}{\partial h_i}\gamma_i'(t)dt
  &= \int_0^1\frac{\partial f(h(r(t)))}{\partial h_i}
    \sum_{j=1}^d\frac{\partial h_i(r(t))}{\partial \bfx_j}r_j'(t)dt \\
  &= \int_0^1\frac{\partial f(h(r(t)))}{\partial h_i}
    \sum_{j=1}^d\frac{\partial h_i(r(t))}{\partial \bfx_j}r_j'(t)dt \\
  &= \sum_{j=1}^d \left\{
    \int_0^1\frac{\partial f(h(r(t)))}{\partial h_i}
    \frac{\partial h_i(r(t))}{\partial \bfx_j}r_j'(t)dt\right\}
\end{align*}
This thus gives the lemma
\begin{lemma}
  Under curve $r: \Real \mapsto \Real^d$ where $r(0) = \bfx$ and $r(1) = \bfx'$,
  the attribution for $h_i$ for a differentiable function $f$ is
  \begin{align}
    \label{appendix:eq:intermediate-attribution-integral}
    \IG^f_{h_i}(\bfx, \bfx', r) = \sum_{j=1}^d \left\{
    \int_0^1\frac{\partial f(h(r(t)))}{\partial h_i}
    \frac{\partial h_i(r(t))}{\partial \bfx_j}r'_j(t)dt\right\}
  \end{align}
\end{lemma}
Note that (6) nicely recovers attributions for input layer, in which case $h$ is the
identity function.
\vskip 5pt
\noindent\textbf{Summation approximation.} Similarly, we can approximate the above
Riemann integral using a summation. Suppose we slice $[0, 1]$ into $m$ equal segments,
then (\ref{eq:intermediate-attribution-integral}) can be approximated as:
\begin{align}
  \label{appendix:eq:intermediate-attribution-sum-approx}
  \IG^f_{h_i}(\bfx, \bfx') =
  \frac{1}{m}\sum_{j=1}^d
  \left\{
  \sum_{k=0}^{m-1}
  \frac{\partial f(h(r(k/m)))}{\partial h_i}
  \frac{\partial h_i(r(k/m))}{\partial \bfx_j}
  r'_j(k/m)
  \right\}
\end{align}

\subsection{Proof of Proposition~\ref{prop:recover-madry}}
\label{sec:proof-recover-madry}
If we put $\lambda=1$ and let $s(\cdot)$ be the $\sf sum$ function
(sum all components of a vector), then for any curve $r$ and
any intermediate layer $\bfh$, (\ref{eq:rar-objective}) becomes:
\begin{align*}
  \rho(\bfx, y; \theta)
  &= \ell(\bfx, y; \theta) + \max_{\bfx' \in N(\bfx, \varepsilon)}
    \{{\sf sum}(\IG^{\ell_y}(\bfx, \bfx'; r))\} \\
  &= \ell(\bfx, y; \theta) + \max_{\bfx' \in N(\bfx, \varepsilon)}
    \{\ell(\bfx', y; \theta)
    - \ell(\bfx, y; \theta)\} \\
  &= \max_{\bfx' \in N(\bfx, \varepsilon)}\ell(\bfx',y; \theta)
\end{align*}
where the second equality is due to the Axiom of Completeness of $\IG$.

\subsection{Proof of Proposition~\ref{prop:recover-input-gradient-regularization}}
Input gradient regularization is an old idea proposed by
Drucker and LeCun~\cite{DL92}, and is recently used by
Ross and Doshi-Velez~\cite{RDV18} in adversarial training setting.
Basically, for $q \ge 1$, they propose
$\rho(\bfx, y; \theta)
= \ell(\bfx, y; \theta) + \lambda \|\nabla_{\bfx} \ell(\bfx, y; \theta)\|_q^q,$
where they want small gradient at $\bfx$. To recover this objective
from robust attribution regularization, let us pick $s(\cdot)$ as the
$\|\cdot\|_1^q$ function (1-norm to the $q$-th power),
and consider the simplest curve
$r(t) = \bfx + t(\bfx' - \bfx)$. With the na\"{i}ve summation approximation
of the integral $\IG^{\ell_y}_i$ we have
$\IG^{\ell_y}_i(\bfx, \bfx'; r)
\approx \frac{(\bfx'_i - \bfx_i)}{m}
\sum_{k=1}^m
\frac{
  \partial \ell(\bfx + \frac{k-1}{m}(\bfx'-\bfx), y; \theta)}{\partial \bfx_i}$,
where larger $m$ is, more accurate we approximate the integral.
Now, if we put $m=1$, which is the coarsest approximation, this becomes
$(\bfx'_i - \bfx_i)\frac{\partial\ell(\bfx, y; \theta)}{\partial \bfx_i}$,
and we have $\IG^{\ell_y}(\bfx, \bfx'; \theta)
= (\bfx'-\bfx) \odot \nabla_{\bfx}\ell(\bfx, y; \theta).$
Therefore (\ref{eq:rar-objective}) becomes:
\begin{align*}
  \rho(\bfx, y; \theta)
  =& \ell(\bfx, y; \theta) + \lambda \max_{\bfx' \in N(\bfx, \varepsilon)}
     \{\|\IG^{\ell_y}(\bfx, \bfx'; \theta)\|_1^q\} \\
  \approx &\ell(\bfx, y; \theta)
            + \lambda \max_{\bfx' \in N(\bfx, \varepsilon)}\{
            \|(\bfx'-\bfx) \odot \nabla_{\bfx}\ell(\bfx, y; \theta)\|_1^q\}
\end{align*}
Put the neighborhood as $\|\bfx' - \bfx\|_p \le \varepsilon$
where $p \in [1, \infty]$ and $\frac{1}{p} + \frac{1}{q} = 1$.
By H\"{o}lder's inequality,
$\|(\bfx'-\bfx) \odot \nabla_{\bfx}\ell(\bfx, y; \theta)\|_1^q
\le \|\bfx' - \bfx\|_p^q\|\nabla\ell(\bfx, y;\theta)\|_q^q
\le \varepsilon^q \|\nabla\ell(\bfx, y;\theta)\|_q^q$
which means that
$\max_{\|\bfx'-\bfx\|_p \le \varepsilon}
\{\|(\bfx'-\bfx) \odot \nabla_{\bfx}\ell(\bfx, y; \theta)\|_1^q\}
= \varepsilon^q \|\nabla\ell(\bfx, y;\theta)\|_q^q.$
Thus by putting $\lambda = \lambda'/\varepsilon^q$,
we recover gradient regularization with regularization parameter $\lambda'$.

\subsection{Proof of Proposition~\ref{prop:regularize-by-loss-output}}
Let us put $s(\cdot) = \|\cdot\|_1$, and $\bfh = \ell_y$
(the output layer of loss function!), then we have
\begin{align*}
  \rho(\bfx, y; \theta)
  = &\ell_y(\bfx) + \max_{\bfx' \in N(\bfx, \varepsilon)}\{
      \|\IG^{\ell_y}_{\ell_y}(\bfx, \bfx'; r)\|_1\} \\
  =&\ell_y(\bfx) + \max_{\bfx' \in N(\bfx, \varepsilon)}\{
     |\ell_y(\bfx') - \ell_y(\bfx)|\}    
\end{align*}
where the second equality is because
$\IG^{\ell_y}_{\ell_y}(\bfx, \bfx'; r) = \ell_y(\bfx') - \ell_y(\bfx)$.

\subsection{Proof of
  Proposition~\ref{prop:recover-wasserstein-prediction-robustness-objective}}
Specifically, again, let $s(\cdot)$ be the summation function and
$\lambda=1$, then we have
$\Exp_{Z, Z'}[d_{\IG}(Z, Z')]
= \Exp_{Z, Z'}[{\sf sum}(\IG_{\bfh}^\ell(Z, Z'))]
= \Exp_{Z, Z'}[\ell(Z';\theta) - \ell(Z;\theta)].$
Because $P$ and $Z$ are identically distributed, thus the objective
reduces to
\begin{align*}
  &\sup_{Q;M\in \prod(P,Q)}\Big\{
    \Exp_{Z, Z'}[\ell(Z;\theta) + \ell(Z';\theta) - \ell(Z;\theta)]\\
  &\qquad\qquad\qquad \text{ s.t. } \Exp_{Z,Z'}[c(Z, Z')] \le \rho
    \Big\} \\
  =& \sup_{Q;M\in \prod(P,Q)}\left\{
     \Exp_{Z'}[\ell(Z';\theta)]
     \text{ s.t. } \Exp_{Z,Z'}[c(Z, Z')] \le \rho \right\} \\
  =& \sup_{Q: W_c(P, Q) \le \rho}\left\{\Exp_Q[\ell(Q; \theta)]\right\},
\end{align*}
which is exactly Wasserstein prediction robustness objective.

\subsection{Proof of Theorem~\ref{thm:dist-rar-duality2}}
\label{sec:proof-dist-rar-duality2}

The proof largely follows that for Theorem 5 in~\cite{SND18}, and we provide it here for completeness.
% When $d_{\IG}(z, z')$ is continuous and $c(z, z')$ is closed convex, we know that $- d^\gamma_{\IG}(z, z') = - d_{\IG}(z, z') + \gamma c(z, z')$ is a normal integrand~\cite{rockafellar2009variational}. 
Since we have a joint supremum over $Q$ and $M \in \prod(P, Q)$ we have that
\begin{align*}
  \sup_{Q; M \in \prod(P, Q)}\left\{
  \Exp_{M=(Z, Z')}\big[d^\gamma_{\IG}(Z, Z')\big]\right\} 
  & = \sup_{Q; M \in \prod(P, Q)}
    \int [d_{\IG}(z, z') - \gamma c(z, z')] d M(z, z') \\
  & \le \int \sup_{z'}\{ d_{\IG}(z, z') - \gamma c(z, z')  \} d P(z) \\
  & = \Exp_{z \sim P}\left[\sup_{z'}\{d^\gamma_{\IG}(z, z')\}\right].
\end{align*}
We would like to show equality in the above. 

Let $\mathcal{Q}$ denote the space of regular conditional probabilities from $Z$ to $Z'$. Then
\begin{align*}
  \sup_{Q; M \in \prod(P, Q)} \int [d_{\IG}(z, z') - \gamma c(z, z')]  d M(z, z') 
  \ge \sup_{Q \in \mathcal{Q}}  \int [d_{\IG}(z, z') - \gamma c(z, z') ] d Q(z'|z) d P(z).
\end{align*}
Let $\mathcal{Z'}$ denote all measurable mappings $z \rightarrow z'(z)$ from $Z$ to $Z'$. Using the measurability result in Theorem 14.60 in~\cite{rockafellar2009variational}, we have
\begin{align*}
  \sup_{z' \in \mathcal{Z'}} \int [d_{\IG}(z, z'(z)) - \gamma c(z, z'(z)) ] d P(z) = \int \sup_{z' } [d_{\IG}(z, z') - \gamma c(z, z') ] d P(z)
\end{align*}
since $\gamma c - d_{\IG} $ is a normal integrand. 

Let $z'(z)$ be any measurable function that is $\epsilon$-close to attaining the supremum above, and define the conditional distribution $Q(z'|z)$ to be supported on $z'(z)$. Then 
\begin{align*}
  \sup_{Q; M \in \prod(P, Q)} \int [d_{\IG}(z, z') - \gamma c(z, z')]  d M(z, z')  
  & \ge \int [d_{\IG}(z, z') - \gamma c(z, z') ] d Q(z'|z) d P(z) \\
  & = \int [d_{\IG}(z, z'(z)) - \gamma c(z, z'(z)) ] d P(z) \\
  & \ge \int \sup_{z'} [d_{\IG}(z, z') - \gamma c(z, z') ] d P(z) - \epsilon \\
  & \ge \sup_{Q; M \in \prod(P, Q)} \int [d_{\IG}(z, z') - \gamma c(z, z')]  d M(z, z') - \epsilon.
\end{align*}
Since $\epsilon \ge 0$ is arbitrary, this completes the proof. \qed

\subsection{Proof of Theorem~\ref{thm:dist-rob-duality}:
  Connections Between the Distributional Robustness Objectives}
\label{sec:proof-dist-rob-duality}
Let $\theta^*$ denote an optimal solution of (\ref{eq:dist-rar-objective}) and let $\theta'$ be any non-optimal solution. Let $\gamma(\theta^*)$ denote the corresponding $\gamma$ by Lemma~\ref{lem:duality-inside}, and $\gamma(\theta')$ denote that for $\theta'$. 

Since $\gamma(\theta')$ achieves the infimum, we have
\begin{align}
  & \Exp_{z \sim P}\left[\ell(z; \theta')
    + \lambda \sup_{z'}\{ d_{\IG}(z, z') - \gamma(\theta^*) c(z, z')\} \right]
  \\
  \ge & 
	\Exp_{z \sim P}\left[\ell(z; \theta')
        + \lambda \sup_{z'}\{ d_{\IG}(z, z') - \gamma(\theta') c(z, z')\} \right]
  \\
  > & \Exp_{z \sim P}\left[\ell(z; \theta^*)
      + \lambda \sup_{z'}\{ d_{\IG}(z, z') - \gamma(\theta^*) c(z, z')\} \right].
\end{align}
So $\theta'$ is not optimal for (\ref{eq:dist-rar-objective-lagrange-2}). 
% A similar argument shows that a non-optimal solution of (\ref{eq:dist-rar-objective-lagrange-2}) is not optimal for (\ref{eq:dist-rar-objective}). 
This then completes the proof. \qed

\begin{lemma}
  \label{lem:duality-inside}
  Suppose $c(z,z)=0$ and $d_{\IG}(z,z)=0$ for any $z$, and suppose
  $\gamma c(z, z') - d_{\IG}(z, z')$ is a normal integrand.
  For any $\rho > 0$, there exists $\gamma \ge 0$ such that 
  \begin{align}
    & \sup_{Q; M \in \prod(P, Q)} \left\{
      \Exp_{(Z,Z') \sim M}[d_{\IG}(Z, Z')] \text{ s.t. }
      \Exp_{(Z,Z') \sim M}[c(Z, Z')] \le \rho \right\}
    \\
    = &\inf_{\zeta \ge 0}\Exp_{z \sim P} \left[
        \sup_{z'}\{ d_{\IG}(z, z') - \zeta c(z, z') + \zeta \rho \} \right].
  \end{align}	
  Furthermore, there exists $\gamma \ge 0$ achieving the infimum.
\end{lemma}
This lemma generalizes Theorem 5 in~\cite{SND18}
to a larger, but natural, class of objectives.
\begin{proof} 
  For $Q$ and $M \in \Pi(P,Q)$, let 
  \begin{align}
    \Lambda_{\IG}(Q, M) & := \Exp_{(Z,Z') \sim M}[d_{\IG}(Z, Z')]
    \\
    \Lambda_{c}(Q, M) & := \Exp_{(Z,Z') \sim M}[c(Z, Z')]
  \end{align}

  First, the pair $(Q, M)$ forms a convex set, and $\Lambda_{\IG}(Q, M)$ and $\Lambda_{c}(Q, M)$ are linear functionals over the convex set.
  Set $Q=P$ and set $M$ to the identity coupling (such that $(Z, Z') \sim M$ always has $Z = Z'$). Then $\Lambda_{c}(Q, M) = 0 < \rho$ and thus the Slater's condition holds. Applying standard infinite dimensional duality results (Theorem 8.6.1 in~\cite{luenberger1997optimization}) leads to
  \begin{align}
    & \sup_{Q; M \in \prod(P, Q); \Lambda_c(Q, M) \le \rho}  \Lambda_{\IG}(Q, M)
    \\
    = & 
	\sup_{Q; M \in \prod(P, Q)} 
        \inf_{\zeta \ge 0} \left\{  \Lambda_{\IG}(Q, M) - \zeta \Lambda_c(Q, M) + \zeta \rho \right\}
    \\
    = & 
	\inf_{\zeta \ge 0}  \sup_{Q; M \in \prod(P, Q)} 
        \left\{  \Lambda_{\IG}(Q, M) - \zeta \Lambda_c(Q, M) + \zeta \rho \right\}.
  \end{align}
  Furthermore, there exists $\gamma \ge 0$ achieving the infimum in the last line. 

  Now, it suffices to show that
  \begin{align}
    & \sup_{Q; M \in \prod(P, Q) } 
      \left\{  \Lambda_{\IG}(Q, M) - \gamma \Lambda_c(Q, M) + \gamma \rho \right\}
    \\
    = &  
        \Exp_{z \sim P} \left[ \sup_{z'}\{ d_{\IG}(z, z') - \gamma c(z, z') + \gamma \rho \} \right].
  \end{align}
  This is exactly what Theorem~\ref{thm:dist-rar-duality2} shows.
\end{proof}

\subsection{Proof of Theorem~\ref{thm:one-layer-neural-networks}}
\label{sec:proof-one-layer-neural-networks}
Let us fix any one point $\bfx$, and consider
$g(-y_i\langle \bfw, \bfx \rangle) + \lambda\max_{\bfx' \in N(\bfx, \varepsilon)}\|\IG^{\ell_y}_{\bfx}(
\bfx, \bfx'; \bfw)\|_1$. Due to the special form of $g$, we know that:
\begin{align*}
  \IG^{\ell_y}_i(\bfx, \bfx'; \bfw) = \frac{\bfw_i(\bfx' - \bfx)_i}{\langle \bfw, \bfx'-\bfx \rangle}
  \cdot \big(g(-y\langle \bfw, \bfx'\rangle) - g(-y\langle \bfw, \bfx\rangle)\big)
\end{align*}
Let $\Delta = \bfx'-\bfx$ (which satisfies that $\|\Delta\|_\infty \le \varepsilon)$, therefore its
absolute value (note that we are taking 1-norm):
\begin{align*}
  \frac{\big|g(-y\langle \bfw, \bfx\rangle - y\langle \bfw, \Delta \rangle)
  - g(-y\langle \bfw, \bfx\rangle)\big|)}{|\langle \bfw, \Delta \rangle|}
  \cdot |\bfw_i\Delta_i|
\end{align*}
Let $z = -y\langle \bfw, \bfx \rangle$ and $\delta = -y\langle \bfw, \Delta\rangle$, this is further
simplified as $\frac{|g(z+\delta)-g(z)|}{|\delta|}|\delta_i|$. Because $g$ is non-decreasing,
so $g' \ge 0$, and so this is indeed $\frac{g(z+\delta) - g(z)}{\delta}$, which is the slope of the
secant from $(z, g(z))$ to $(z+\delta, g(z+\delta))$. Because $g$ is convex so the secant slopes are
non-decreasing, so we can simply pick $\Delta_i=-y\sign(\bfw_i)\varepsilon$,
and so $\delta = \|\bfw\|_1\varepsilon$, and so that $\|\IG\|_1$ becomes
\begin{align*}
  |g(z+\varepsilon\|\bfw\|_1) - g(z)| \cdot \frac{\sum_i|\bfw_i\Delta_i|}{|\delta|}
  &= |g(z+\varepsilon\|\bfw\|_1) - g(z)| \cdot \frac{\sum_i|\bfw_i|\varepsilon}{\|\bfw\|_1\varepsilon} \\
  &= |g(z+\varepsilon\|\bfw\|_1) - g(z)| \\
  &= g(z+\varepsilon\|\bfw\|_1) - g(z)
\end{align*}
where the last equality follows because $g$ is nondecreasing. Therefore the objective simplifies to
$\sum_{i=1}^mg(-y_i\langle \bfw, \bfx_i \rangle + \varepsilon\|\bfw\|_1)$,
which is exactly Madry et al.'s objective under $\ell_\infty$ perturbations.\qed

Let us consider two examples:

\noindent\emph{Logistic Regression}.
Let $g(z)=\ln(1+\exp(z))$. Then $g(-y\langle \bfw, \bfx \rangle)$
recovers the Negative Log-Likelihood loss for logistic regression. Clearly $g$ is nondecreasing
and $g'$ is also nondecreasing. As a result, adversarial training for logistic regression is exactly
``robustifying'' attributions/explanations.

\noindent\emph{Softplus hinge loss}. Alternatively, we can let $g(z) = \ln(1+\exp(1+z))$,
and therefore $g(-y\langle \bfw, \bfx \rangle) = \ln(1+\exp(1-y\langle \bfw, \bfx\rangle))$
is the softplus version of the hinge loss function. Clearly this $g$ also satisfy our requirements,
and therefore adversarial training for softplus hinge loss function is also exactly about
``robustifying'' attributions/explanations.

\section{More Details of Experiments}
\label{sec:experiments-details}

\subsection{Experiment Settings}
\label{sec:experiment-settings}

We perform experiments on four datasets: MNIST, Fashion-MNIST, GTSRB and Flower.
Robust attribution regularization training requires extensive computing power. 
We conducted experiments in parallel over multiple NVIDIA Tesla V100 and NVDIA GeForce RTX 2080Ti GPUs both on premises and on cloud. Detailed experiment settings for each dataset are described below.

\subsubsection{MNIST}
\noindent\textbf{Data}.
The MNIST dataset~\cite{mnist} is a large dataset of handwritten digits. Each digit has 5,500 training images and 1,000 test images. Each image is a $28 \times 28$ grayscale. We normalize the range of pixel values to $[0, 1]$. 

\noindent\textbf{Model}.
We use a network consisting of two convolutional layers with 32 and 64 filters respectively, each followed by $2 \times 2$ max-pooling, and a fully connected layer of size 1024. Note that we use the same MNIST model as \cite{madry2017towards}. % We compare our RAR model with their naturally trained model and adversarially trained model. 

\noindent\textbf{Training hyper-parameters}.
The hyper-parameters to train different models are listed below:

\noindent\emph{NATURAL}.
We set learning rate as $10^{-4}$, batch size as 50, training steps as 25,000, and use Adam Optimizer.

\noindent\emph{Madry et al.}.
We set learning rate as $10^{-4}$, batch size as 50, training steps as 100,000, and use Adam Optimizer. We use PGD attack as adversary with random start, number of steps of 40, step size of 0.01, and adversarial budget $\epsilon$ of 0.3. 

\noindent\emph{IG-NORM}.
We set $\lambda=1$, $m=50$ for gradient step, learning rate as $10^{-4}$, batch size as 50, training steps as 100,000, and use Adam Optimizer. We use PGD attack as adversary with random start, number of steps of 40, step size of 0.01, $m=10$ for attack step, and adversarial budget $\epsilon=0.3$. 

\noindent\emph{IG-SUM-NORM}.
We set $\beta$ as 0.1, $m$ in the gradient step as 50, learning rate as $10^{-4}$, batch size as 50, training steps as 100,000, and use Adam Optimizer. We use PGD attack as adversary with random start, number of steps of 40, step size of 0.01, $m=10$ in the attack step, and adversarial budget $\epsilon=0.3$. 

\noindent\textbf{Evaluation Attacks}.
For attacking inputs to change model predictions, we use PGD attack with random start, number of steps of $100$, adversarial budget $\epsilon$ of 0.3 and step size of $0.01$. For attacking inputs to change interpretations, we use Iterative Feature Importance Attacks (IFIA) proposed by ~\cite{GAZ17}. We use their top-k attack with $k=200$, adversarial budget $\epsilon=0.3$, step size $\alpha=0.01$ and number of iterations $P=100$. We set the feature importance function as Integrated Gradients(IG) and dissimilarity function $D$ as Kendall's rank order correlation. We find that IFIA is not stable if we use GPU parallel computing (non-deterministic is a behavior of GPU), so we run IFIA three times on each test example and use the best result with the lowest Kendall's rank order correlation. 

\subsubsection{Fashion-MNIST}
\noindent\textbf{Data}.
The Fashion-MNIST dataset ~\cite{xiao2017fashion} contains images depicting wearables
such as shirts and boots instead of digits, which is more complex than MNIST dataset. The image format,
the number of classes, as well as the number of examples are
all identical to MNIST.

\noindent\textbf{Model}.
We use a network consisting of two convolutional layers with 32 and 64 filters respectively, each followed by $2 \times 2$ max-pooling, and a fully connected layer of size 1024. % We compare our RAR model with their naturally trained model and adversarially trained model. 

\noindent\textbf{Training hyper-parameters}.
The hyper-parameters to train different models are listed below:

\noindent\emph{NATURAL}.
We set learning rate as $10^{-4}$, batch size as 50, training steps as 25,000, and use Adam Optimizer.

\noindent\emph{Madry et al.}.
We set learning rate as $10^{-4}$, batch size as 50, training steps as 100,000, and use Adam Optimizer. We use PGD attack as adversary with random start, number of steps of 20, step size of 0.01, and adversarial budget $\epsilon$ of 0.1. 

\noindent\emph{IG-NORM}.
We set $\lambda=1$, $m=50$ for gradient step, learning rate as $10^{-4}$, batch size as 50, training steps as 100,000, and use Adam Optimizer. We use PGD attack as adversary with random start, number of steps of 20, step size of 0.01, $m=10$ for attack step, and adversarial budget $\epsilon=0.1$. 

\noindent\emph{IG-SUM-NORM}.
We set $\beta$ as 0.1, $m$ in the gradient step as 50, learning rate as $10^{-4}$, batch size as 50, training steps as 100,000, and use Adam Optimizer. We use PGD attack as adversary with random start, number of steps of 20, step size of 0.01, $m=10$ in the attack step, and adversarial budget $\epsilon=0.1$. 

\noindent\textbf{Evaluation Attacks}.
For attacking inputs to change model predictions, we use PGD attack with random start, number of steps of $100$, adversarial budget $\epsilon$ of 0.1 and step size of $0.01$. For attacking inputs to change interpretations, we use Iterative Feature Importance Attacks (IFIA) proposed by ~\cite{GAZ17}. We use their top-k attack with $k=100$, adversarial budget $\epsilon=0.1$, step size $\alpha=0.01$ and number of iterations $P=100$. We set the feature importance function as Integrated Gradients(IG) and dissimilarity function $D$ as Kendall's rank order correlation. We find that IFIA is not stable if we use GPU parallel computing (non-deterministic is a behavior of GPU), so we run IFIA three times on each test example and use the best result with the lowest Kendall's rank order correlation. 

\subsubsection{GTSRB}
\noindent\textbf{Data}.
The German Traffic Sign Recognition Benchmark (GTSRB) \cite{stallkamp2012man} is a dataset of color images depicting 43 different traffic signs. The images are not of a fixed dimensions and have rich background and varying light conditions as would be expected of photographed images of traffic signs. There are about 34,799 training images, 4,410 validation images and 12,630 test images. We resize each image to $32 \times 32$. The pixel values are in range of $[0, 255]$. The dataset has a large imbalance in the number of sample occurrences across classes. We use data augmentation techniques to enlarge the training data and make the number of samples in each class balanced. We construct a class preserving data augmentation pipeline consisting of rotation, translation, and projection transforms and apply this pipeline to images in the training set until each class contained 10,000 training examples. We use this new augmented training data set containing 430,000 samples in total to train models. We also preprocess images via image brightness normalization.

\noindent\textbf{Model} .
We use the Resnet model \cite{he2016deep}. We perform per image standardization before feeding images to the neural network. The network has 5 residual units with (16, 16, 32, 64) filters each. The model is adapted from CIFAR-10 model of \cite{madry2017towards}. Refer to our codes for details. 

\noindent\textbf{Training hyper-parameters}.
The hyper-parameters to train different models are listed below:

\noindent\emph{NATURAL}.
We use Momentum Optimizer with weight decay. We set momentum rate as 0.9, weight decay rate as 0.0002, batch size as 64, and training steps as 70,000.  We use learning rate schedule: the first 500 steps, we use learning rate of $10^{-3}$; after 500 steps and before 60,000 steps, we use learning rate of $10^{-2}$; after 60,000 steps, we use learning rate of $10^{-3}$.

\noindent\emph{Madry et al.}.
We use Momentum Optimizer with weight decay. We set momentum rate as 0.9, weight decay rate as 0.0002, batch size as 64, and training steps as 70,000.  We use learning rate schedule: the first 500 steps, we use learning rate of $10^{-3}$; after 500 steps and before 60,000 steps, we use learning rate of $10^{-2}$; after 60,000 steps, we use learning rate of $10^{-3}$.  We use PGD attack as adversary with random start, number of steps of 7, step size of 2, and adversarial budget $\epsilon$ of 8. 

\noindent\emph{IG-NORM}.
We set $\lambda$ as 1, $m$ in the gradient step as 50. We use Momentum Optimizer with weight decay. We set momentum rate as 0.9, weight decay rate as 0.0002, batch size as 64, and training steps as 70,000.  We use learning rate schedule: the first 500 steps, we use learning rate of $10^{-6}$; after 500 steps and before 60,000 steps, we use learning rate of $10^{-4}$; after 60,000 steps, we use learning rate of $10^{-5}$. We use PGD attack as adversary with random start, number of steps of 7, step size of 2, $m$ in the attack step of 5, and adversarial budget $\epsilon$ of 8.

\noindent\emph{IG-SUM-NORM}.
We set $\beta$ as 1, $m$ in the gradient step as 50. We use Momentum Optimizer with weight decay. We set momentum rate as 0.9, weight decay rate as 0.0002, batch size as 64, and training steps as 70,000.  We use learning rate schedule: the first 500 steps, we use learning rate of $10^{-5}$; after 500 steps and before 60,000 steps, we use learning rate of $10^{-4}$; after 60,000 steps, we use learning rate of $10^{-5}$. We use PGD attack as adversary with random start, number of steps of 7, step size of 2, $m$ in the attack step of 5, and adversarial budget $\epsilon$ of 8.

\noindent\textbf{Evaluation Attacks}.
For attacking inputs to change model predictions, we use PGD attack with number of steps of $40$, adversarial budget $\epsilon$ of 8 and step size of $2$. For attacking inputs to change interpretations, we use Iterative Feature Importance Attacks (IFIA) proposed by ~\cite{GAZ17}. We use their top-k attack with $k=100$, adversarial budget $\epsilon=8$, step size $\alpha=1$ and number of iterations $P=50$. We set the feature importance function as Integrated Gradients(IG) and dissimilarity function $D$ as Kendall's rank order correlation. We find that IFIA is not stable if we use GPU parallel computing (non-deterministic is a behavior of GPU), so we run IFIA three times on each test example and use the best result with the lowest Kendall's rank order correlation.

\subsubsection{Flower}
\noindent\textbf{Data}.
Flower dataset \cite{nilsback2006visual} is a dataset of 17 category flowers with 80 images for each class (1,360 image in total). The flowers chosen are some common flowers in the UK. The images have large scale, pose and light variations and there are also classes with large variations of images within the class and close similarity to other classes. We randomly split the dataset into training and test sets. The training set has totally 1,224 images with 72 images per class. The test set has totally 136 images with 8 images per class. We resize each image to $128 \times 128$. The pixel values are in range of $[0,255]$. We use data augmentation techniques to enlarge the training data. We construct a class preserving data augmentation pipeline consisting of rotation, translation, and projection transforms and apply this pipeline to images in the training set until each class contained 1,000 training examples. We use this new augmented training data set containing 17,000 samples in total to train models. 

\noindent\textbf{Model}.
We use the Resnet model \cite{he2016deep}. We perform per image standardization before feeding images to the neural network. The network has 5 residual units with (16, 16, 32, 64) filters each. The model is adapted from CIFAR-10 model of \cite{madry2017towards}. Refer to our codes for details.

\noindent\textbf{Training hyper-parameters}.
The hyper-parameters to train different models are listed below:

\noindent\emph{NATURAL}.
We use Momentum Optimizer with weight decay. We set momentum rate as 0.9, weight decay rate as 0.0002, batch size as 16, and training steps as 70,000.  We use learning rate schedule: the first 500 steps, we use learning rate of $10^{-3}$; after 500 steps and before 60,000 steps, we use learning rate of $10^{-2}$; after 60,000 steps, we use learning rate of $10^{-3}$.

\noindent\emph{Madry et al.}.
We use Momentum Optimizer with weight decay. We set momentum rate as 0.9, weight decay rate as 0.0002, batch size as 16, and training steps as 70,000.  We use learning rate schedule: the first 500 steps, we use learning rate of $10^{-3}$; after 500 steps and before 60,000 steps, we use learning rate of $10^{-2}$; after 60,000 steps, we use learning rate of $10^{-3}$.  We use PGD attack as adversary with random start, number of steps of 7, step size of 2, and adversarial budget $\epsilon$ of 8.

\noindent\emph{IG-NORM}.
We set $\lambda$ as 0.1, $m$ in the gradient step as 50. We use Momentum Optimizer with weight decay. We set momentum rate as 0.9, weight decay rate as 0.0002, batch size as 16, and training steps as 70,000.  We use learning rate schedule: the first 500 steps, we use learning rate of $10^{-4}$; after 500 steps and before 60,000 steps, we use learning rate of $10^{-3}$; after 60,000 steps, we use learning rate of $10^{-4}$. We use PGD attack as adversary with random start, number of steps of 7, step size of 2, $m$ in the attack step of 5, and adversarial budget $\epsilon$ of 8.

\noindent\emph{IG-SUM-NORM}.
We set $\beta$ as 0.1, $m$ in the gradient step as 50. We use Momentum Optimizer with weight decay. We set momentum rate as 0.9, weight decay rate as 0.0002, batch size as 16, and training steps as 70,000.  We use learning rate schedule: the first 500 steps, we use learning rate of $10^{-4}$; after 500 steps and before 60,000 steps, we use learning rate of $10^{-3}$; after 60,000 steps, we use learning rate of $10^{-4}$. We use PGD attack as adversary with random start, number of steps of 7, step size of 2, $m$ in the attack step of 5, and adversarial budget $\epsilon$ of 8.

\noindent\textbf{Evaluation Attacks}.
For attacking inputs to change model predictions, we use PGD attack with number of steps of $40$, adversarial budget $\epsilon$ of 8 and step size of $2$. For attacking inputs to change interpretations, we use Iterative Feature Importance Attacks (IFIA) proposed by ~\cite{GAZ17}. We use their top-k attack with $k=1000$, adversarial budget $\epsilon=8$, step size $\alpha=1$ and number of iterations $P=100$. We set the feature importance function as Integrated Gradients(IG) and dissimilarity function $D$ as Kendall's rank order correlation. We find that IFIA is not stable if we use GPU parallel computing (non-deterministic is a behavior of GPU), so we run IFIA three times on each test example and use the best result with the lowest Kendall's rank order correlation. 

\subsection{Why a different $m$ in the Attack Step?}
From our experiments, we find that the most time consuming part during training is using adversary $\calA$ to find $\bfx^*$. It is because we need to run several PGD steps to find $\bfx^*$. To speed it up, we set a smaller $m$ (no more than 10) in the attack step. 

\subsection{Choosing Hyper-parameters}
Our IG-NORM (or IG-SUM-NORM) objective contains hyper-parameters $m$ in the attack step, $m$ in the gradient step and $\lambda$ (or $\beta$). From our experiments, we find that if $\lambda$ (or $\beta$) is too large, the training cannot converge. And if $\lambda$ (or $\beta$) is too small, we cannot get good attribution robustness. To select best $\lambda$ (or $\beta$), we try three values: 1, 0.1, and 0.01, and use the one with the best attribution robustness. For $m$ in the attack step, due to the limitation of computing power, we usually set a small value, typically 5 or 10.
We study how $m$ in the gradient step affects results on MNIST using IG-NORM objective. We try $m \in \{10, 20, 30, \cdots, 100\}$, and set $\lambda=1$ and $m$ in the attack step as 10. Other training settings are the same. The results are summarized in Table~\ref{table:different-m}. 

\begin{table}[htb]
  \centering
  \begin{tabular}{ c | c | c | c | c }
    \hline
    $m$ & {\tt NA} & {\tt AA} &  {\tt IN} & {\tt CO} \\ \hline
    10  & 98.54\% & 78.05\% & 67.14\% & 0.2574 \\ \hline
    20 & 98.72\% & 80.29\% & 70.78\% & 0.2699 \\ \hline
    30 & 98.70\% & 80.44\% & 71.06\% & 0.2640 \\ \hline
    40 & 98.79\% & 73.41\% & 64.76\%  & 0.2733  \\ \hline
    50 & 98.74\% & 81.43\% & {\bf 71.36\%}  & {\bf 0.2841} \\ \hline
    60 & 98.78\% & 89.25\% & 63.55\%  & 0.2230 \\ \hline
    70 & 98.80\% & 74.78\% & 67.37\%  & 0.2556 \\ \hline
    80 & 98.75\% & 80.26\% & 69.90\%  & 0.2633 \\ \hline
    90 & 98.61\% & 78.54\% & 70.88\%  & 0.2787 \\ \hline
    100 & 98.59\% & 89.36\% & 59.70\%  & 0.2210 \\ \hline
  \end{tabular}
  \vspace{1mm}
  \caption{Experiment results for different $m$ in gradient step on MNIST.}
  \label{table:different-m}
  \vspace{-7mm}
\end{table}

From the results, we can see when $m = 50$, we can get the best attribution robustness. 
For objective IG-SUM-NORM and other datasets, we do similar search for $m$ in the gradient step.
We find that usually, $m = 50$ can give good attribution robustness. 

\subsection{Dimensionality and effectiveness of attribution attack}
% \noindent\textbf{Dimensionality and effectiveness of attribution attack}.
Similar to~\cite{GAZ17}, we observe that IFIA is not so successful when
number of dimensions is relatively small. For example, on GTSRB
dataset the number of dimensions is relatively small ($32\times 32 \times 3$),
and if one uses small adversarial budget ($8/255\approx 0.031$),
the attacks become not very effective.
On the other hand, even though MNIST dimension is small ($28\times 28 \times 1$)
, the attack remains effective for large budget ($0.3$).
On Flower dataset the number of dimension is large ($128 \times 128 \times 3$),
and the attack is very effective on this dataset. 

\subsection{Use Simple Gradient to Compute Feature Importance Maps}

We also experiment with Simple Gradient (SG)~\cite{simonyan2013deep}
instead of Integrated Gradients (IG) to compute feature importance map.
\change{The experiment settings are the same as previous ones except that
  we use SG to compute feature importance map in order to compute
  rank correlation and top intersection, and also in the Iterative Feature
  Importance Attacks (IFIA) (evaluation attacks). The results
  are summarized in Table \ref{table:simple-gradient}.
  Our method produces significantly better attribution robustness
  than both natural training and adversarial training, except being
  slightly worse than adversarial training on Fashion-MNIST.
  We note that Fashion-MNIST is also the only data set in our
  experiments where IG results are significantly different from
  that of SG (where under IG, IG-SUM-NORM is significantly better).
  Note that IG is a \emph{princpled sommothed verison} of SG and so
  this result highlights differences between these two attribution methods
  on a particular data set. More investigation into this phenomenon
  seems warranted.
}

\begin{table}[htb]
	\centering
	\begin{tabular}{ c | c | c | c | c | c }
		\hline
		Dataset & Approach & {\tt NA} & {\tt AA} &  {\tt IN} & {\tt CO} \\ \hline
		\multirow{3}{*}{MNIST} 
		& NATURAL & 99.17\% & 0.00\% & 16.64\% & 0.0107 \\ \cline{2-6}
		& Madry et al. & 98.40\% & 92.47\% & 47.95\% & 0.2524 \\ \cline{2-6}
		& IG-SUM-NORM & 98.34\% & 88.17\% & {\bf 61.67\%} & {\bf 0.2918} \\ \hline \hline
		\multirow{3}{*}{Fashion-MNIST} 
		& NATURAL & 90.86\% & 0.01\% & 21.55\% & 0.0734 \\ \cline{2-6}
		& Madry et al. & 85.73\% & 73.01\% & {\bf 58.37\%} & {\bf 0.3947} \\ \cline{2-6}
		& IG-SUM-NORM & 85.44\% & 70.26\% & 54.91\% & 0.3674 \\ \hline \hline
		\multirow{3}{*}{GTSRB} 
		& NATURAL & 98.57\% & 21.05\% & 51.31\% & 0.6000 \\ \cline{2-6}
		& Madry et al. & 97.59\% & 83.24\% & 70.27\% & 0.6965 \\ \cline{2-6}
		& IG-SUM-NORM & 95.68\%  & 77.12\% & {\bf 75.03\%} & {\bf 0.7151} \\ \hline \hline
		\multirow{3}{*}{Flower} 
		& NATURAL & 86.76\% & 0.00\% & 6.72\% &  0.2996 \\ \cline{2-6}
		& Madry et al. & 83.82\% & 41.91\% & 54.10\% & 0.7282 \\ \cline{2-6}
		& IG-SUM-NORM & 82.35\% & 47.06\% & {\bf 65.59\%} & {\bf 0.7503} \\ \hline
	\end{tabular}
	\vspace{1mm}
	\caption{Experiment results for using Simple Gradient to compute feature importance maps.}
	\label{table:simple-gradient}
	\vspace{-8mm}
\end{table}

\subsection{Additional Visualization Results}
\label{sec:additional-visualization-results}

\begin{figure}[htb]
  \centering 
  \begin{minipage}{\linewidth}
    \centering
    NATURAL \hspace{2.5cm} IG-NORM \hspace{2.5cm} IG-SUM-NORM
  \end{minipage} 
  \begin{subfigure}{\textwidth}
    \centering
    \begin{subfigure}[b]{.32\textwidth}
      \centering
      \includegraphics[width=0.48\linewidth,bb=0 0 449 464]{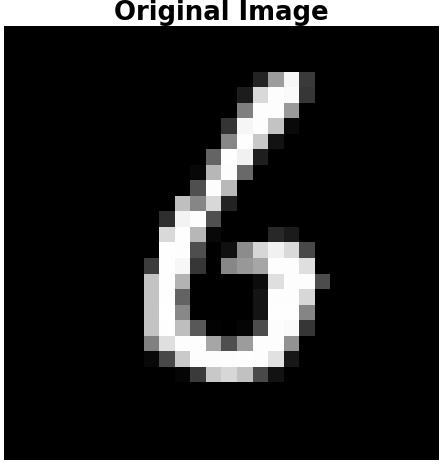} 
      \includegraphics[width=0.48\linewidth,bb=0 0 449 464]{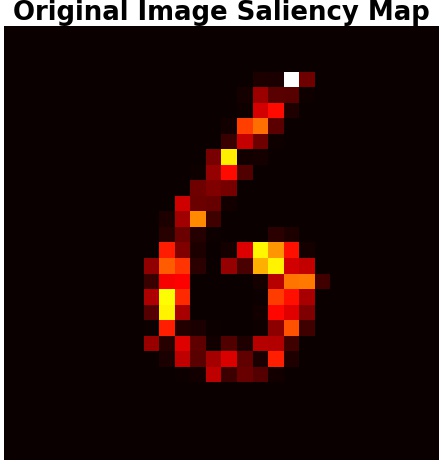} \\
      \includegraphics[width=0.48\linewidth,bb=0 0 449 464]{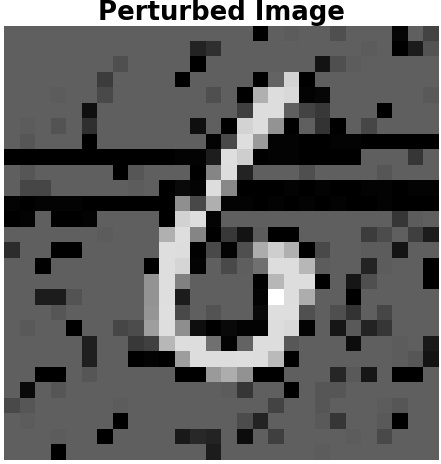} 
      \includegraphics[width=0.48\linewidth,bb=0 0 449 464]{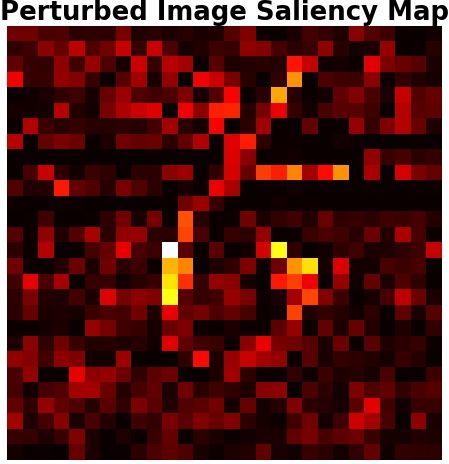} 
      \captionsetup{justification=centering}
      \caption*{Top-100 Intersection: 37.0\% \\  Kendall's Correlation: 0.0567}
    \end{subfigure}
    \begin{subfigure}[b]{.32\textwidth}
      \centering
      \includegraphics[width=0.48\linewidth,bb=0 0 449 464]{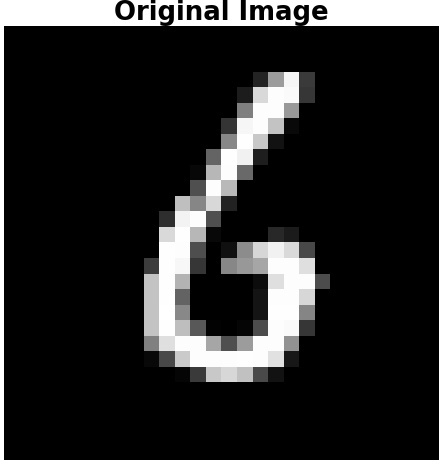} 
      \includegraphics[width=0.48\linewidth,bb=0 0 449 464]{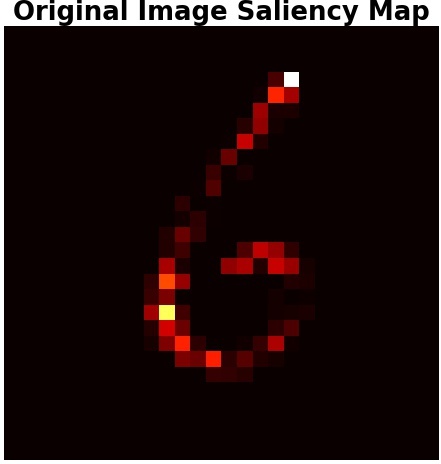} \\
      \includegraphics[width=0.48\linewidth,bb=0 0 449 464]{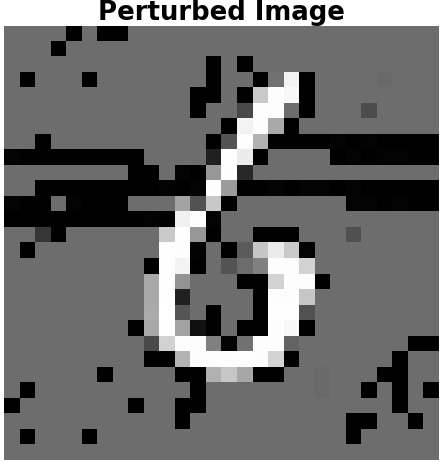} 
      \includegraphics[width=0.48\linewidth,bb=0 0 449 464]{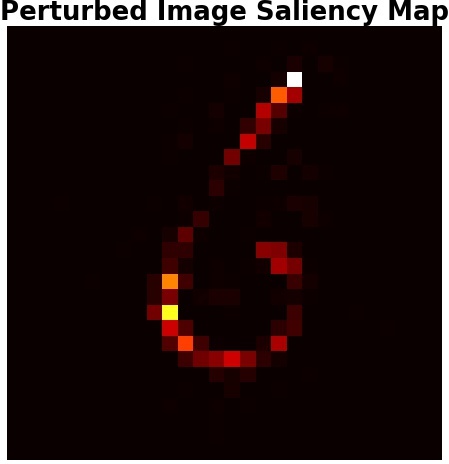} 
      \captionsetup{justification=centering}
      \caption*{Top-100 Intersection: 64.0\% \\ Kendall's Correlation: 0.1823}
    \end{subfigure}
    \begin{subfigure}[b]{.32\textwidth}
      \centering
      \includegraphics[width=0.48\linewidth,bb=0 0 449 464]{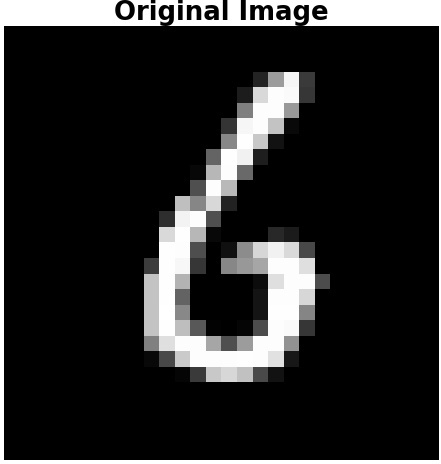} 
      \includegraphics[width=0.48\linewidth,bb=0 0 449 464]{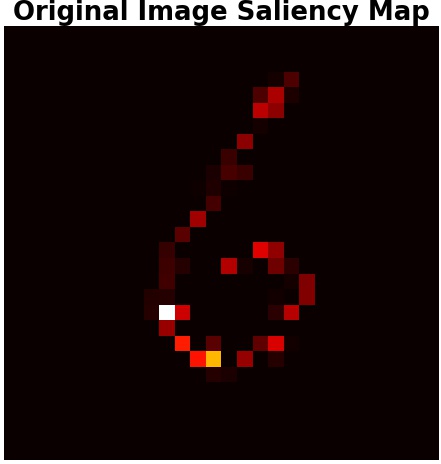} \\
      \includegraphics[width=0.48\linewidth,bb=0 0 449 464]{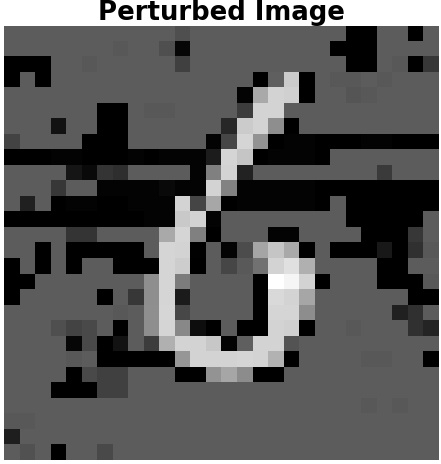} 
      \includegraphics[width=0.48\linewidth,bb=0 0 449 464]{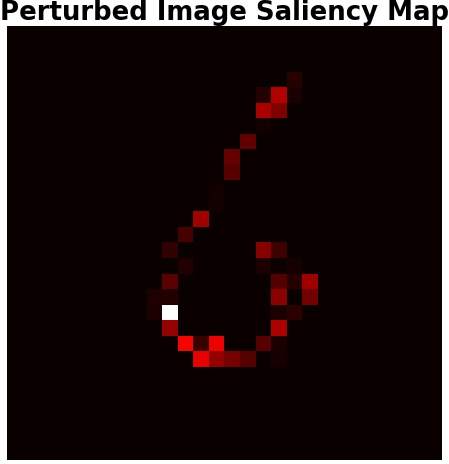} 
      \captionsetup{justification=centering}
      \caption*{Top-100 Intersection: 67.0\% \\ Kendall's Correlation: 0.2180}
    \end{subfigure}
	\caption{For all images, the models give \emph{correct} prediction -- 6.}
  \end{subfigure}
	\begin{subfigure}{\textwidth}
		\centering
		\begin{subfigure}[b]{.32\textwidth}
			\centering
			\includegraphics[width=0.48\linewidth,bb=0 0 449 464]{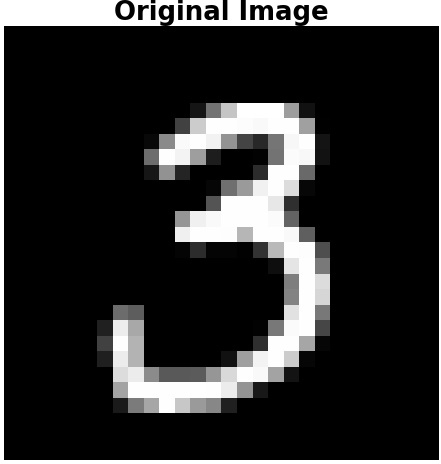} 
			\includegraphics[width=0.48\linewidth,bb=0 0 449 464]{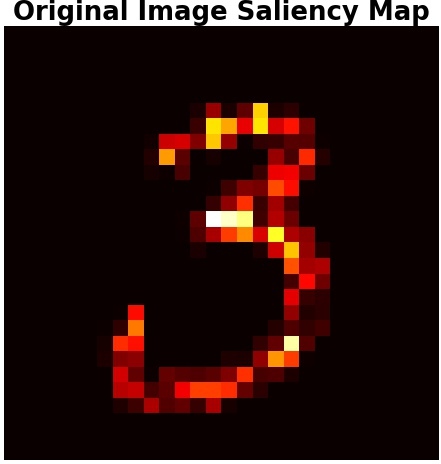} \\
			\includegraphics[width=0.48\linewidth,bb=0 0 449 464]{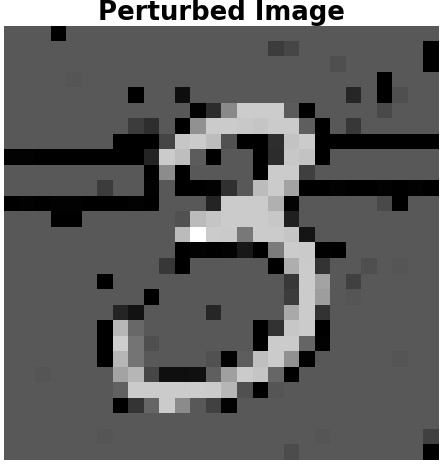} 
			\includegraphics[width=0.48\linewidth,bb=0 0 449 464]{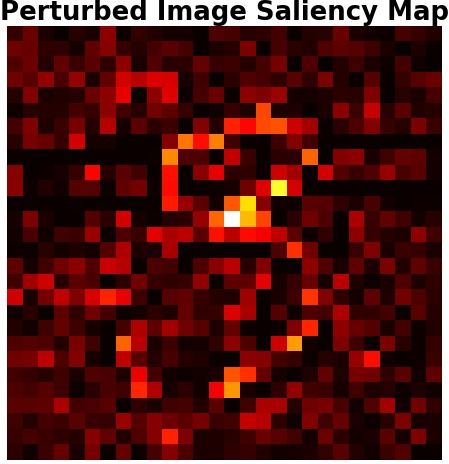} 
			\captionsetup{justification=centering}
			\caption*{Top-100 Intersection: 43.0\% \\  Kendall's Correlation: 0.0563}
		\end{subfigure}
		\begin{subfigure}[b]{.32\textwidth}
			\centering
			\includegraphics[width=0.48\linewidth,bb=0 0 449 464]{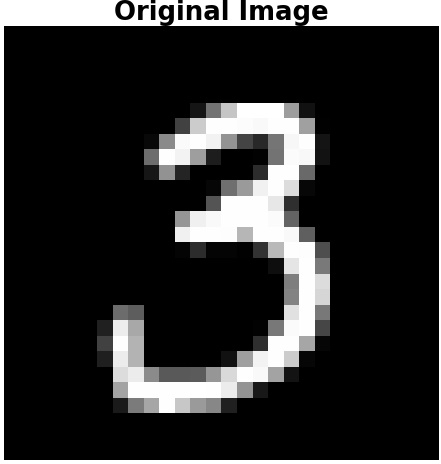} 
			\includegraphics[width=0.48\linewidth,bb=0 0 449 464]{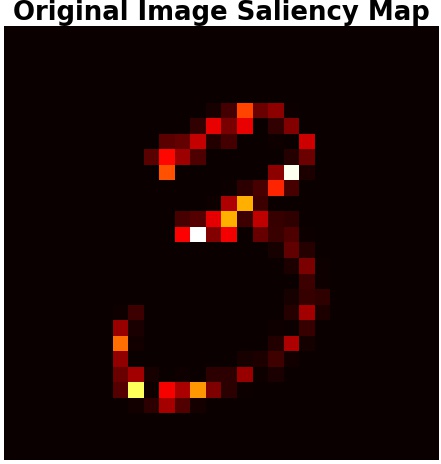} \\
			\includegraphics[width=0.48\linewidth,bb=0 0 449 464]{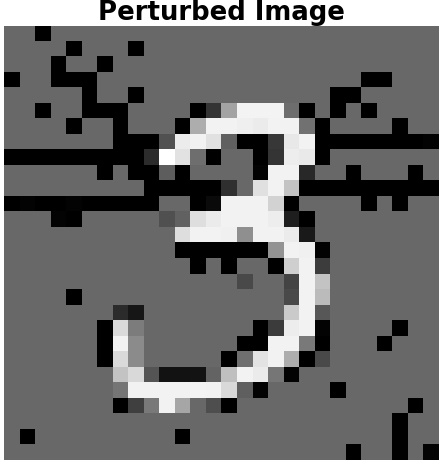} 
			\includegraphics[width=0.48\linewidth,bb=0 0 449 464]{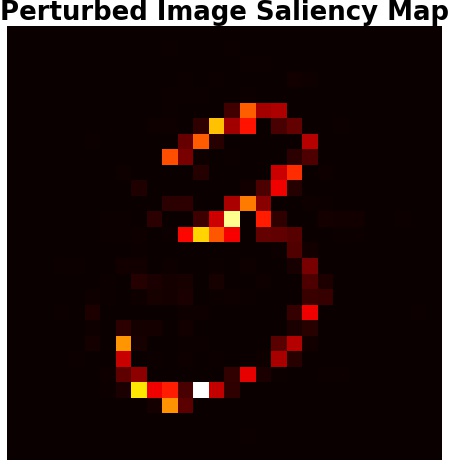} 
			\captionsetup{justification=centering}
			\caption*{Top-100 Intersection: 74.0\% \\ Kendall's Correlation: 0.1718}
		\end{subfigure}
		\begin{subfigure}[b]{.32\textwidth}
			\centering
			\includegraphics[width=0.48\linewidth,bb=0 0 449 464]{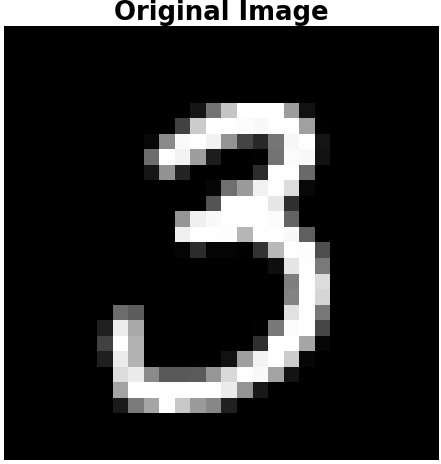} 
			\includegraphics[width=0.48\linewidth,bb=0 0 449 464]{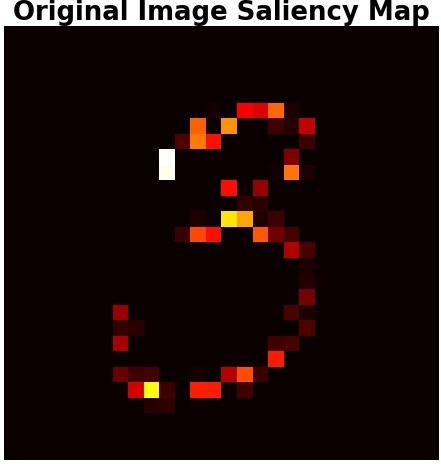} \\
			\includegraphics[width=0.48\linewidth,bb=0 0 449 464]{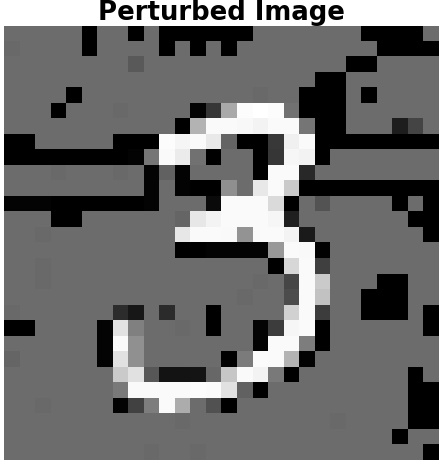} 
			\includegraphics[width=0.48\linewidth,bb=0 0 449 464]{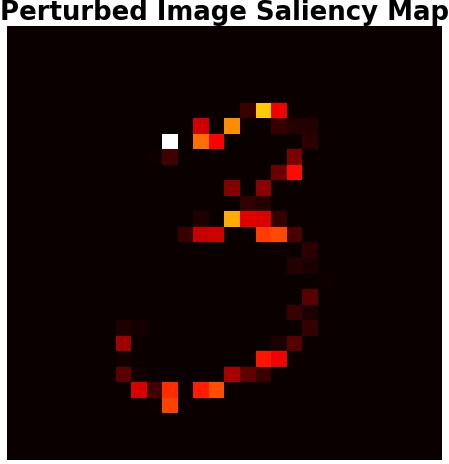} 
			\captionsetup{justification=centering}
			\caption*{Top-100 Intersection: 84.0\% \\ Kendall's Correlation: 0.2501}
		\end{subfigure}
	\caption{For all images, the models give \emph{correct} prediction -- 3.}
	\end{subfigure}
	\begin{subfigure}{\textwidth}
		\centering
		\begin{subfigure}[b]{.32\textwidth}
			\centering
			\includegraphics[width=0.48\linewidth,bb=0 0 449 464]{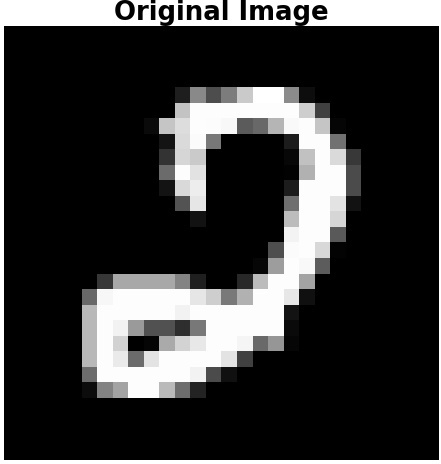} 
			\includegraphics[width=0.48\linewidth,bb=0 0 449 464]{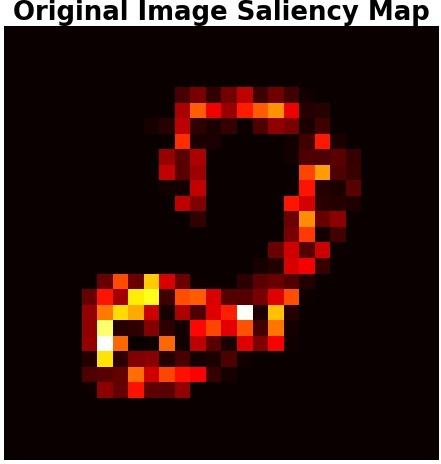} \\
			\includegraphics[width=0.48\linewidth,bb=0 0 449 464]{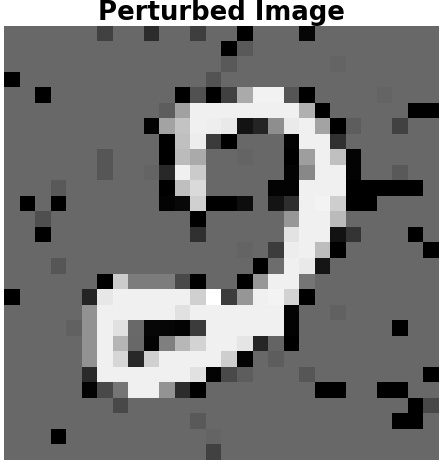} 
			\includegraphics[width=0.48\linewidth,bb=0 0 449 464]{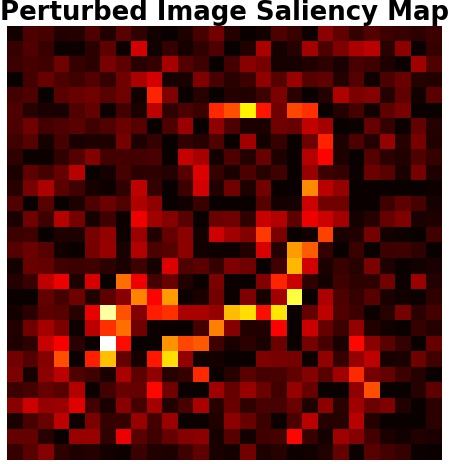} 
			\captionsetup{justification=centering}
			\caption*{Top-100 Intersection: 41.0\% \\  Kendall's Correlation: 0.1065}
		\end{subfigure}
		\begin{subfigure}[b]{.32\textwidth}
			\centering
			\includegraphics[width=0.48\linewidth,bb=0 0 449 464]{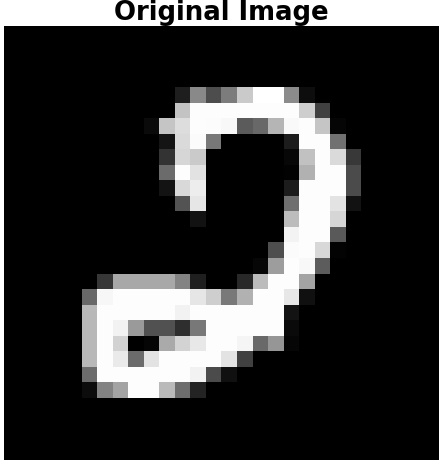} 
			\includegraphics[width=0.48\linewidth,bb=0 0 449 464]{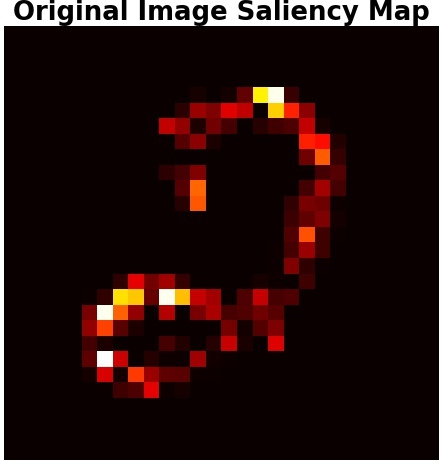} \\
			\includegraphics[width=0.48\linewidth,bb=0 0 449 464]{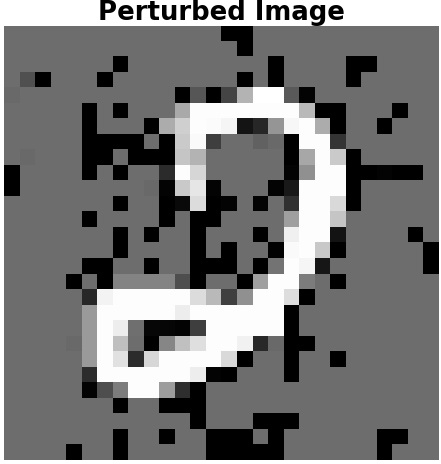} 
			\includegraphics[width=0.48\linewidth,bb=0 0 449 464]{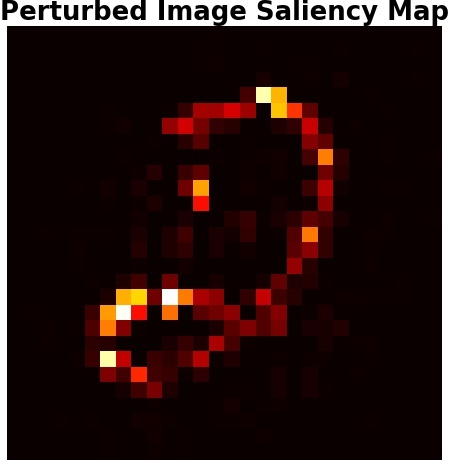} 
			\captionsetup{justification=centering}
			\caption*{Top-100 Intersection: 83.0\% \\ Kendall's Correlation: 0.2837}
		\end{subfigure}
		\begin{subfigure}[b]{.32\textwidth}
			\centering
			\includegraphics[width=0.48\linewidth,bb=0 0 449 464]{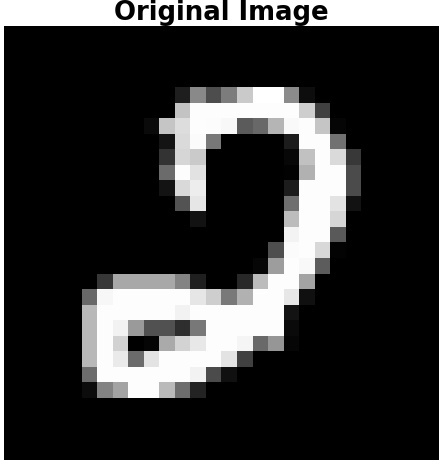} 
			\includegraphics[width=0.48\linewidth,bb=0 0 449 464]{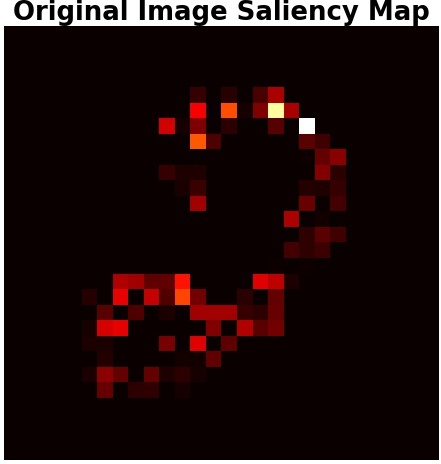} \\
			\includegraphics[width=0.48\linewidth,bb=0 0 449 464]{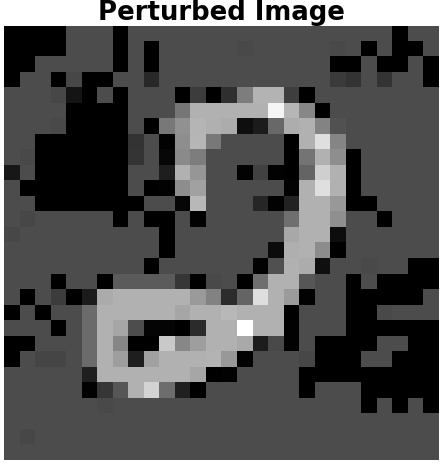} 
			\includegraphics[width=0.48\linewidth,bb=0 0 449 464]{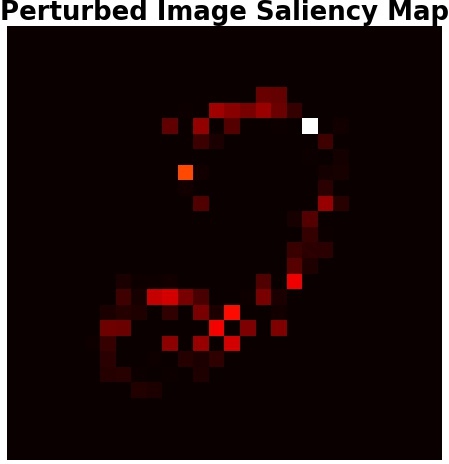} 
			\captionsetup{justification=centering}
			\caption*{Top-100 Intersection: 84.0\% \\ Kendall's Correlation: 0.3151}
		\end{subfigure}
		\caption{For all images, the models give \emph{correct} prediction -- 2.}
	\end{subfigure}
  \caption{Top-100 and Kendall's Correlation are rank correlations
  	between original and perturbed saliency maps.
  	NATURAL is the naturally trained model, IG-NORM and IG-SUM-NORM are models trained using our
  	robust attribution method. We use attribution attacks described in~\cite{GAZ17} to perturb
  	the attributions while keeping predictions intact. For all images, the models give
  	\emph{correct} predictions. However, the saliency maps
  	(also called feature importance maps), computed via IG, show that attributions of
  	the naturally trained model are very fragile, either visually or quantitatively as measured
  	by correlation analysis, while models trained using our method are much more robust
  	in their attributions.}
  \label{fig:ex1}
\end{figure}

\begin{figure}[htb]
	\centering 
	\begin{minipage}{\linewidth}
		\centering
		NATURAL \hspace{2.5cm} IG-NORM \hspace{2.5cm} IG-SUM-NORM
	\end{minipage} 
	\begin{subfigure}{\textwidth}
		\centering
		\begin{subfigure}[b]{.32\textwidth}
			\centering
			\includegraphics[width=0.48\linewidth,bb=0 0 449 464]{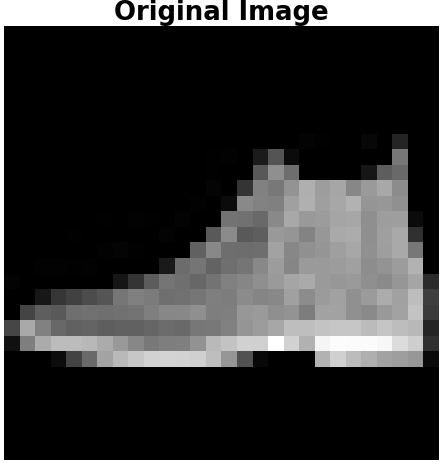} 
			\includegraphics[width=0.48\linewidth,bb=0 0 449 464]{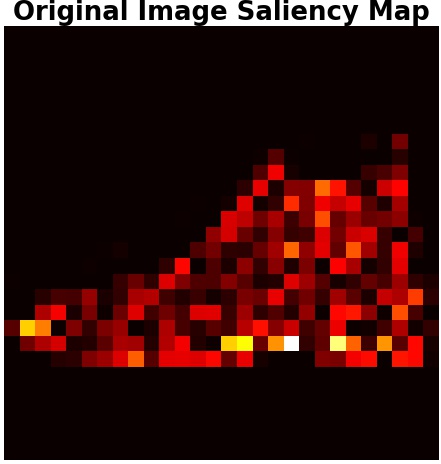} \\
			\includegraphics[width=0.48\linewidth,bb=0 0 449 464]{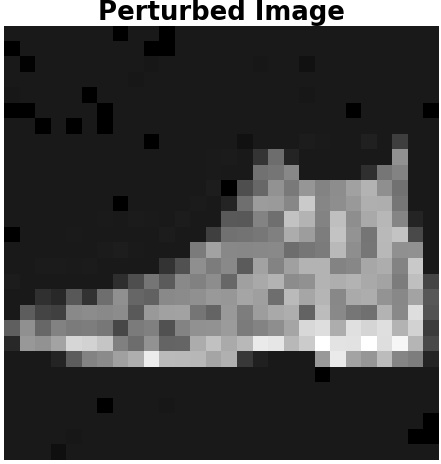} 
			\includegraphics[width=0.48\linewidth,bb=0 0 449 464]{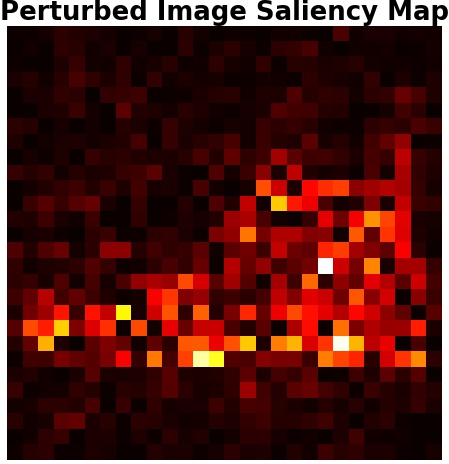} 
			\captionsetup{justification=centering}
			\caption*{Top-100 Intersection: 50.0\% \\  Kendall's Correlation: 0.4595}
		\end{subfigure}
		\begin{subfigure}[b]{.32\textwidth}
			\centering
			\includegraphics[width=0.48\linewidth,bb=0 0 449 464]{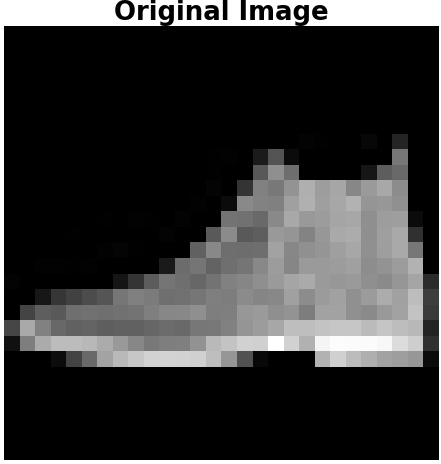} 
			\includegraphics[width=0.48\linewidth,bb=0 0 449 464]{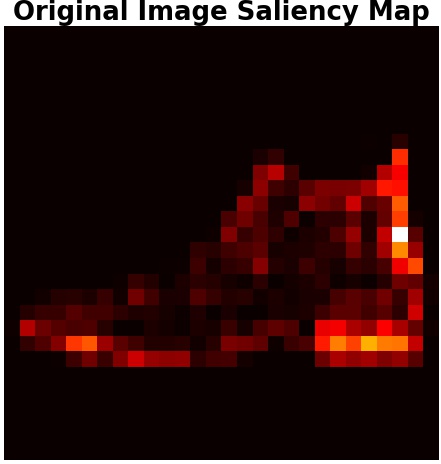} \\
			\includegraphics[width=0.48\linewidth,bb=0 0 449 464]{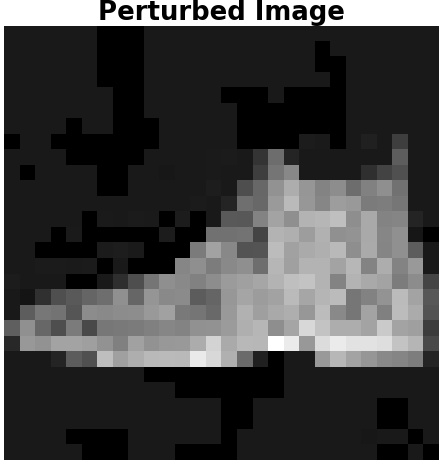} 
			\includegraphics[width=0.48\linewidth,bb=0 0 449 464]{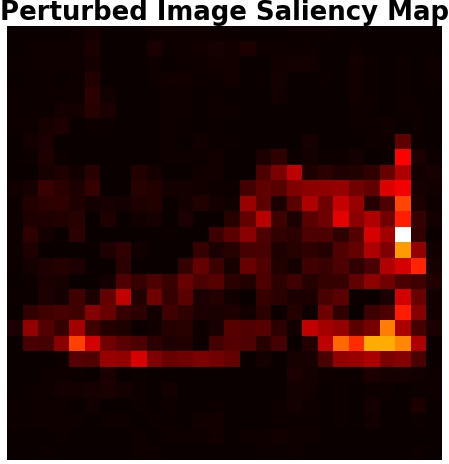} 
			\captionsetup{justification=centering}
			\caption*{Top-100 Intersection: 63.0\% \\ Kendall's Correlation: 0.6099}
		\end{subfigure}
		\begin{subfigure}[b]{.32\textwidth}
			\centering
			\includegraphics[width=0.48\linewidth,bb=0 0 449 464]{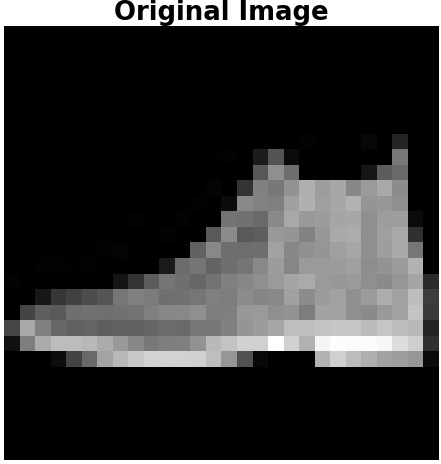} 
			\includegraphics[width=0.48\linewidth,bb=0 0 449 464]{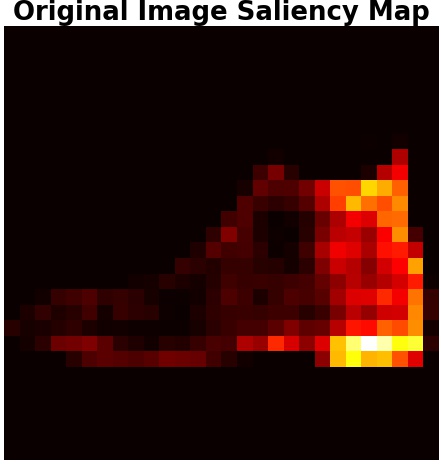} \\
			\includegraphics[width=0.48\linewidth,bb=0 0 449 464]{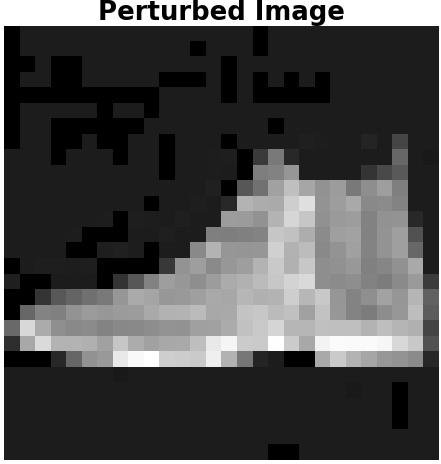} 
			\includegraphics[width=0.48\linewidth,bb=0 0 449 464]{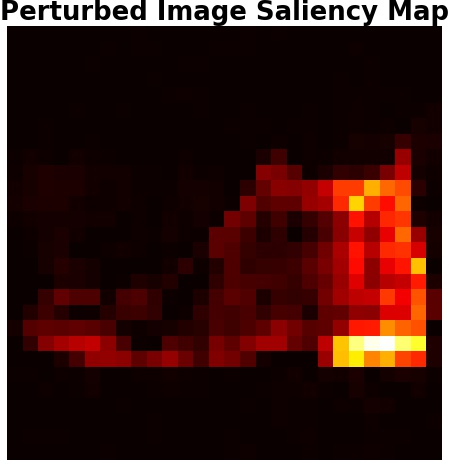} 
			\captionsetup{justification=centering}
			\caption*{Top-100 Intersection: 87.0\% \\ Kendall's Correlation: 0.6607}
		\end{subfigure}
		\caption{For all images, the models give \emph{correct} prediction -- Ankle boot.}
	\end{subfigure}
	\begin{subfigure}{\textwidth}
		\centering
		\begin{subfigure}[b]{.32\textwidth}
			\centering
			\includegraphics[width=0.48\linewidth,bb=0 0 449 464]{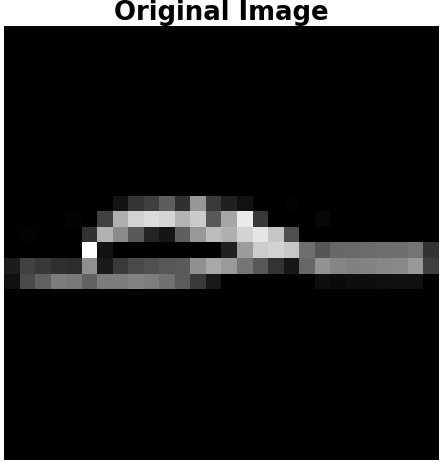} 
			\includegraphics[width=0.48\linewidth,bb=0 0 449 464]{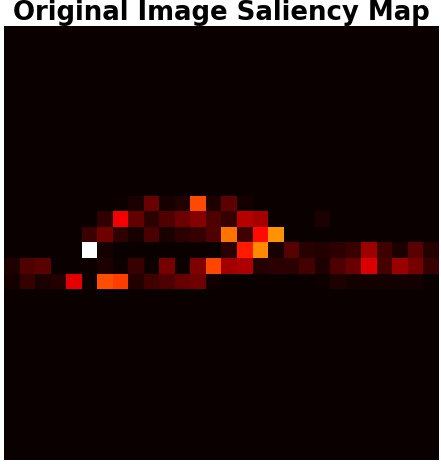} \\
			\includegraphics[width=0.48\linewidth,bb=0 0 449 464]{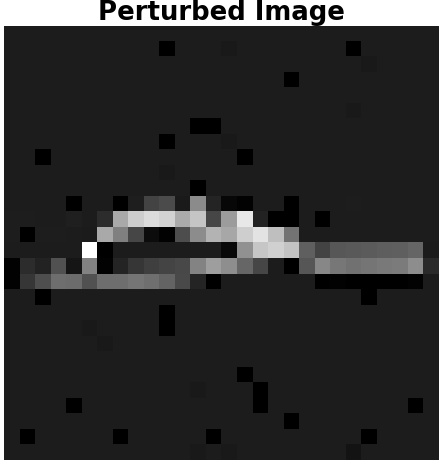} 
			\includegraphics[width=0.48\linewidth,bb=0 0 449 464]{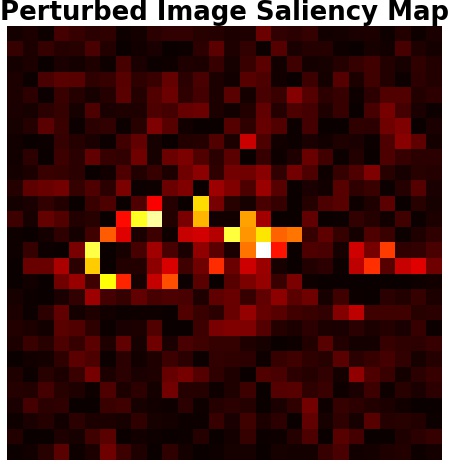} 
			\captionsetup{justification=centering}
			\caption*{Top-100 Intersection: 47.0\% \\  Kendall's Correlation: 0.1293}
		\end{subfigure}
		\begin{subfigure}[b]{.32\textwidth}
			\centering
			\includegraphics[width=0.48\linewidth,bb=0 0 449 464]{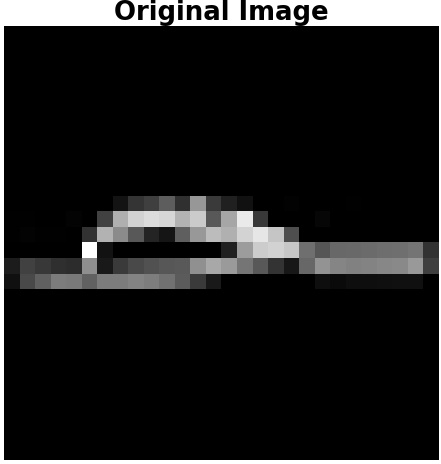} 
			\includegraphics[width=0.48\linewidth,bb=0 0 449 464]{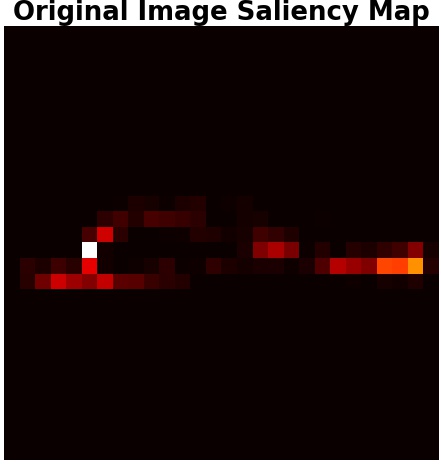} \\
			\includegraphics[width=0.48\linewidth,bb=0 0 449 464]{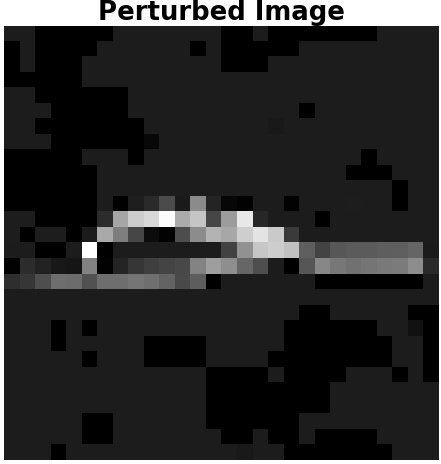} 
			\includegraphics[width=0.48\linewidth,bb=0 0 449 464]{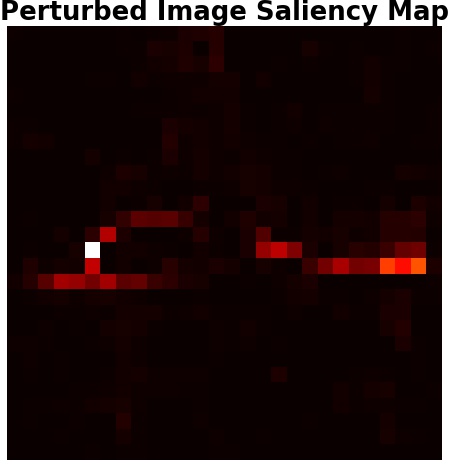} 
			\captionsetup{justification=centering}
			\caption*{Top-100 Intersection: 54.0\% \\ Kendall's Correlation: 0.2508}
		\end{subfigure}
		\begin{subfigure}[b]{.32\textwidth}
			\centering
			\includegraphics[width=0.48\linewidth,bb=0 0 449 464]{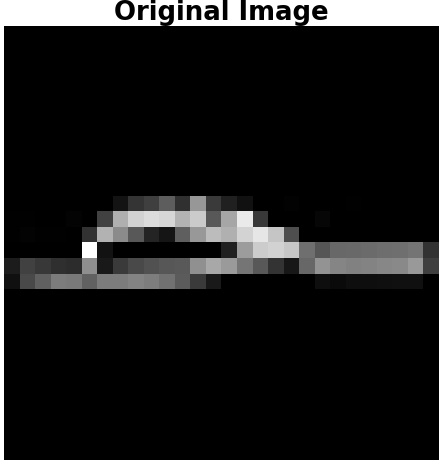} 
			\includegraphics[width=0.48\linewidth,bb=0 0 449 464]{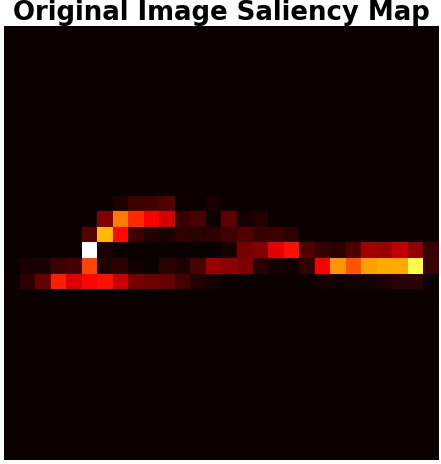} \\
			\includegraphics[width=0.48\linewidth,bb=0 0 449 464]{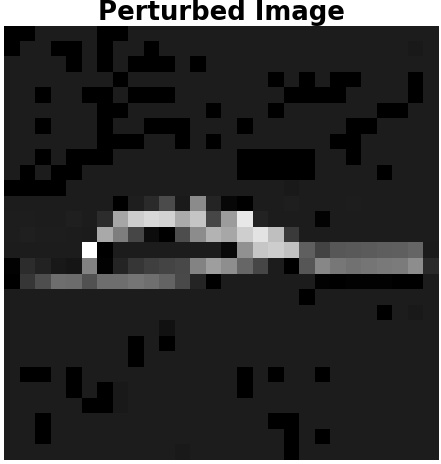} 
			\includegraphics[width=0.48\linewidth,bb=0 0 449 464]{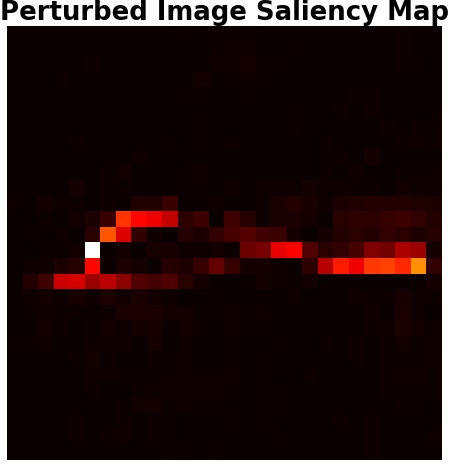} 
			\captionsetup{justification=centering}
			\caption*{Top-100 Intersection: 65.0\% \\ Kendall's Correlation: 0.3136}
		\end{subfigure}
		\caption{For all images, the models give \emph{correct} prediction -- Sandal.}
	\end{subfigure}
	\begin{subfigure}{\textwidth}
		\centering
		\begin{subfigure}[b]{.32\textwidth}
			\centering
			\includegraphics[width=0.48\linewidth,bb=0 0 449 464]{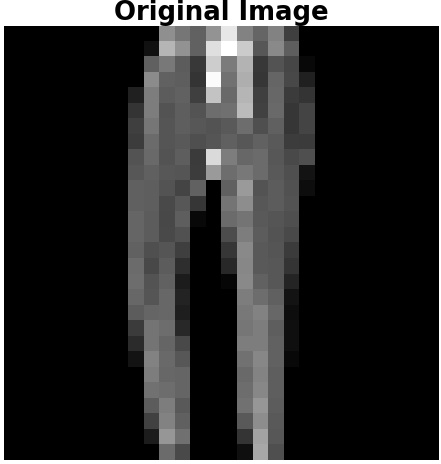} 
			\includegraphics[width=0.48\linewidth,bb=0 0 449 464]{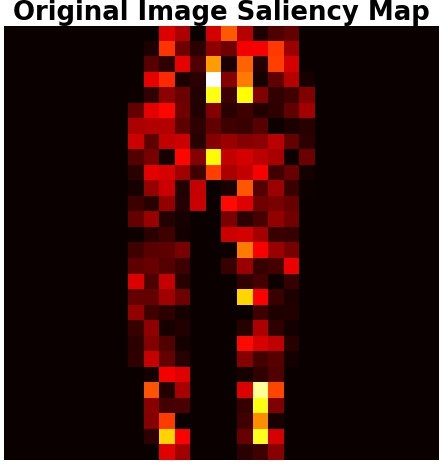} \\
			\includegraphics[width=0.48\linewidth,bb=0 0 449 464]{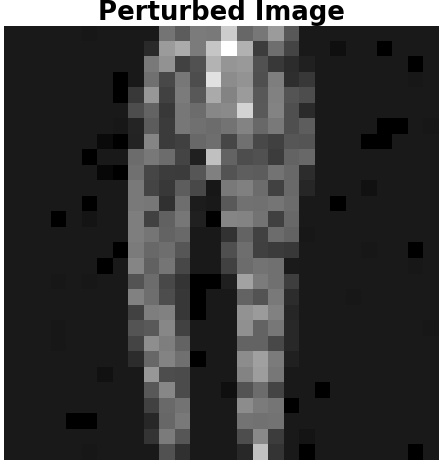} 
			\includegraphics[width=0.48\linewidth,bb=0 0 449 464]{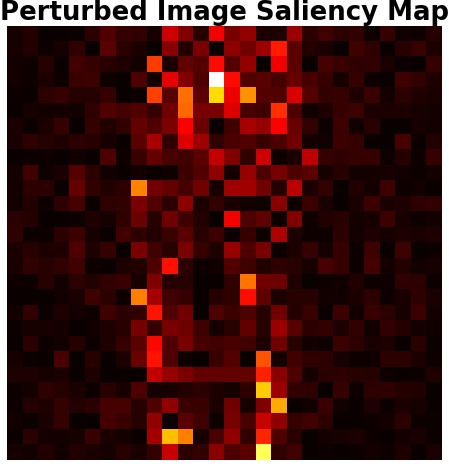} 
			\captionsetup{justification=centering}
			\caption*{Top-100 Intersection: 39.0\% \\  Kendall's Correlation: 0.4129}
		\end{subfigure}
		\begin{subfigure}[b]{.32\textwidth}
			\centering
			\includegraphics[width=0.48\linewidth,bb=0 0 449 464]{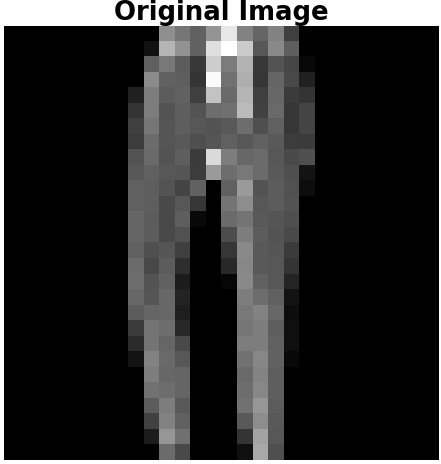} 
			\includegraphics[width=0.48\linewidth,bb=0 0 449 464]{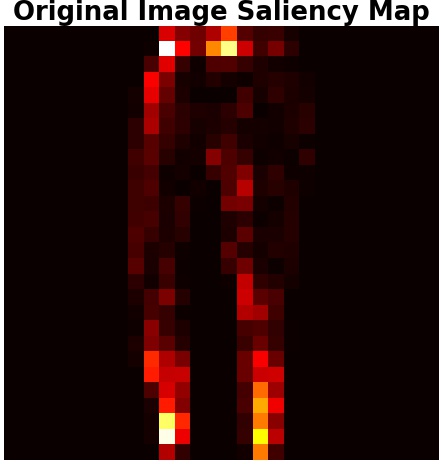} \\
			\includegraphics[width=0.48\linewidth,bb=0 0 449 464]{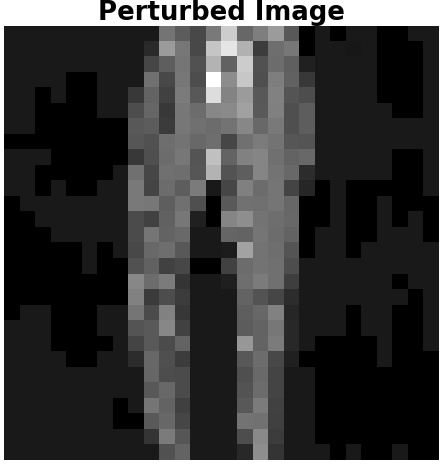} 
			\includegraphics[width=0.48\linewidth,bb=0 0 449 464]{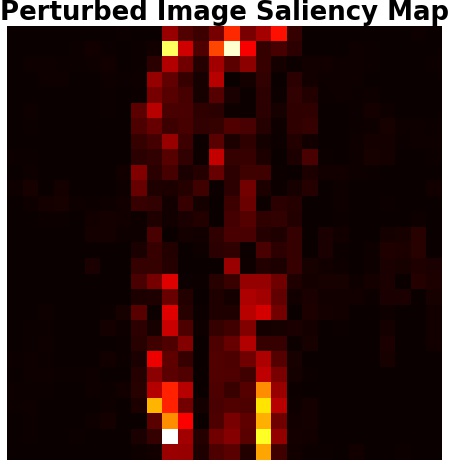} 
			\captionsetup{justification=centering}
			\caption*{Top-100 Intersection: 61.0\% \\ Kendall's Correlation: 0.5983}
		\end{subfigure}
		\begin{subfigure}[b]{.32\textwidth}
			\centering
			\includegraphics[width=0.48\linewidth,bb=0 0 449 464]{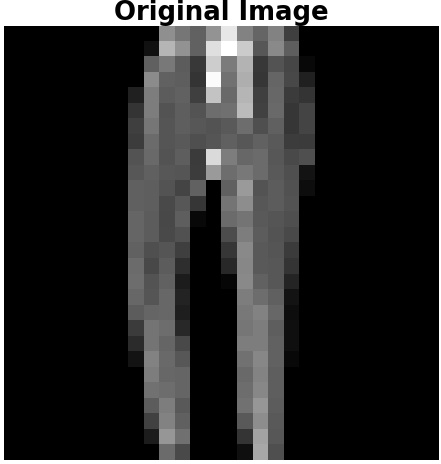} 
			\includegraphics[width=0.48\linewidth,bb=0 0 449 464]{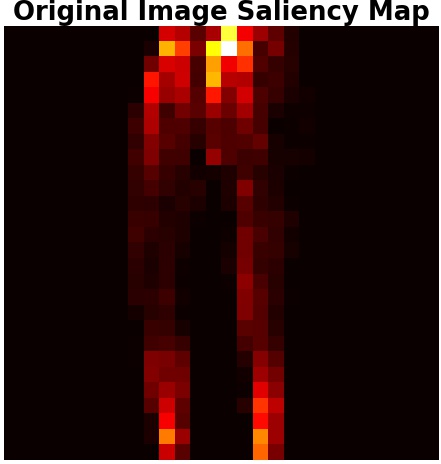} \\
			\includegraphics[width=0.48\linewidth,bb=0 0 449 464]{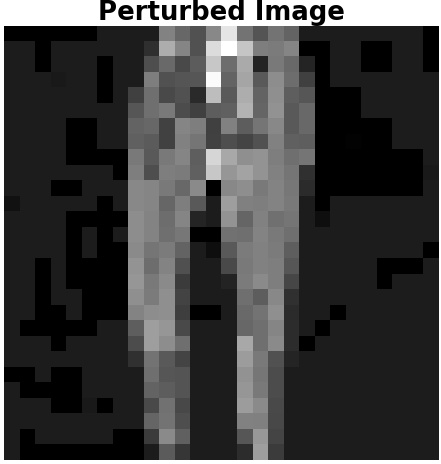} 
			\includegraphics[width=0.48\linewidth,bb=0 0 449 464]{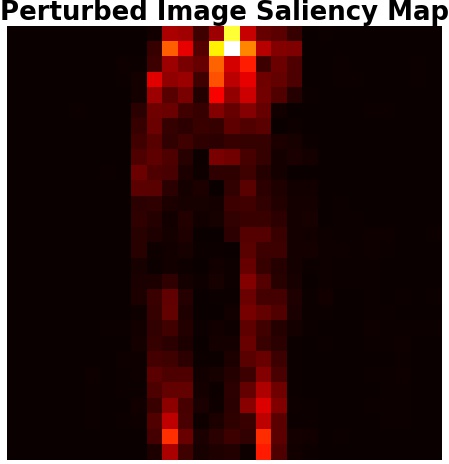} 
			\captionsetup{justification=centering}
			\caption*{Top-100 Intersection: 71.0\% \\ Kendall's Correlation: 0.6699}
		\end{subfigure}
		\caption{For all images, the models give \emph{correct} prediction -- Trouser.}
	\end{subfigure}
	\caption{Top-100 and Kendall's Correlation are rank correlations
		between original and perturbed saliency maps.
		NATURAL is the naturally trained model, IG-NORM and IG-SUM-NORM are models trained using our
		robust attribution method. We use attribution attacks described in~\cite{GAZ17} to perturb
		the attributions while keeping predictions intact. For all images, the models give
		\emph{correct} predictions. However, the saliency maps
		(also called feature importance maps), computed via IG, show that attributions of
		the naturally trained model are very fragile, either visually or quantitatively as measured
		by correlation analysis, while models trained using our method are much more robust
		in their attributions.}
	\label{fig:ex2}
\end{figure}

\begin{figure}[htb]
	\centering 
	\begin{minipage}{\linewidth}
		\centering
		NATURAL \hspace{2.5cm} IG-NORM \hspace{2.5cm} IG-SUM-NORM
	\end{minipage} 
	\begin{subfigure}{\textwidth}
		\centering
		\begin{subfigure}[b]{.32\textwidth}
			\centering
			\includegraphics[width=0.48\linewidth,bb=0 0 449 464]{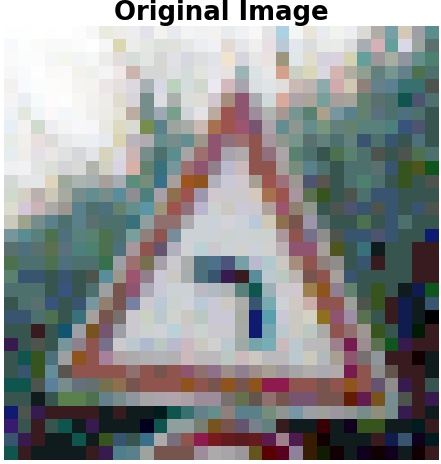} 
			\includegraphics[width=0.48\linewidth,bb=0 0 449 464]{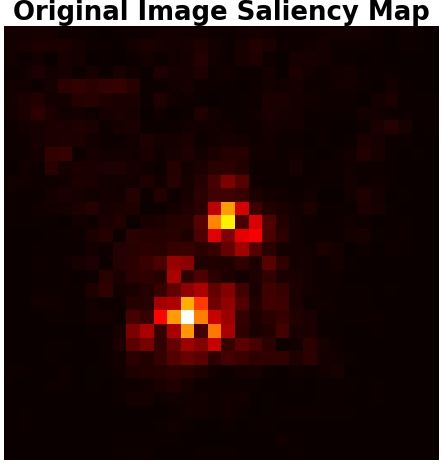} \\
			\includegraphics[width=0.48\linewidth,bb=0 0 449 464]{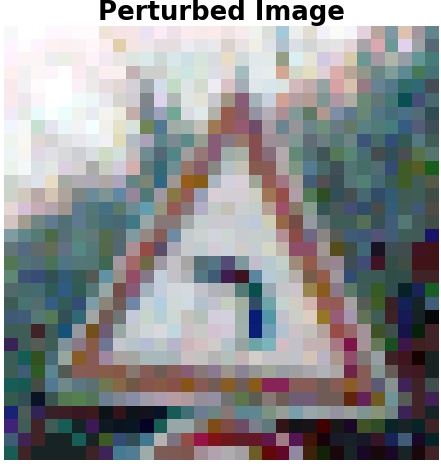} 
			\includegraphics[width=0.48\linewidth,bb=0 0 449 464]{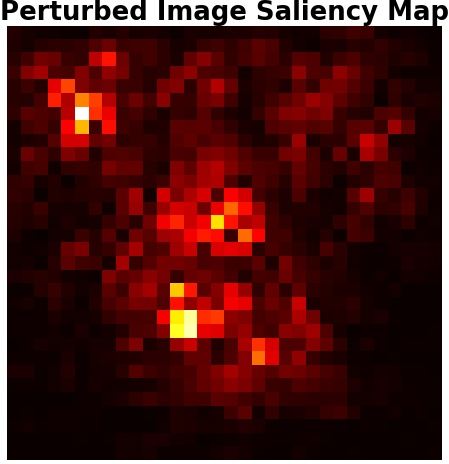} 
			\captionsetup{justification=centering}
			\caption*{Top-100 Intersection: 45.0\% \\  Kendall's Correlation: 0.5822}
		\end{subfigure}
		\begin{subfigure}[b]{.32\textwidth}
			\centering
			\includegraphics[width=0.48\linewidth,bb=0 0 449 464]{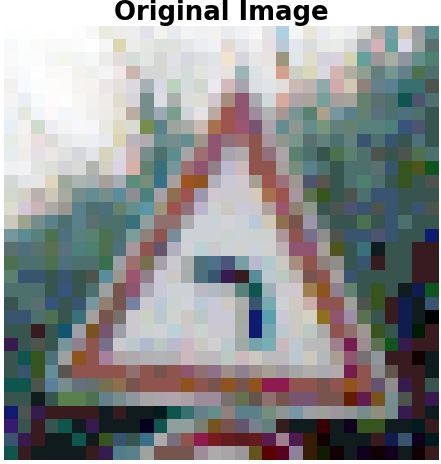} 
			\includegraphics[width=0.48\linewidth,bb=0 0 449 464]{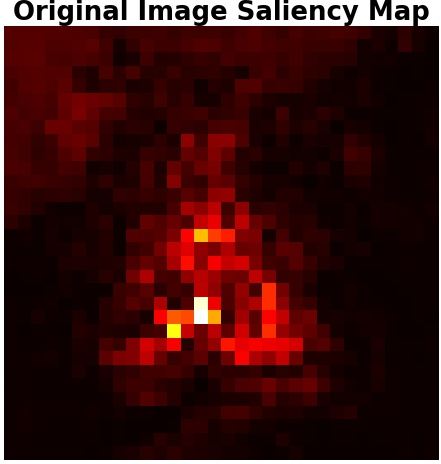} \\
			\includegraphics[width=0.48\linewidth,bb=0 0 449 464]{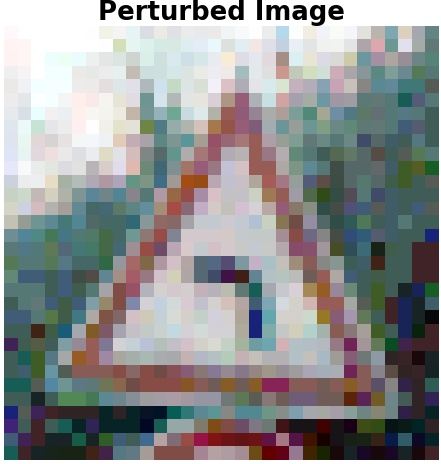} 
			\includegraphics[width=0.48\linewidth,bb=0 0 449 464]{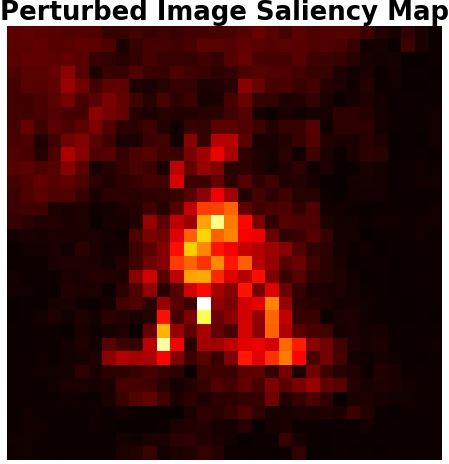} 
			\captionsetup{justification=centering}
			\caption*{Top-100 Intersection: 78.0\% \\ Kendall's Correlation: 0.7471}
		\end{subfigure}
		\begin{subfigure}[b]{.32\textwidth}
			\centering
			\includegraphics[width=0.48\linewidth,bb=0 0 449 464]{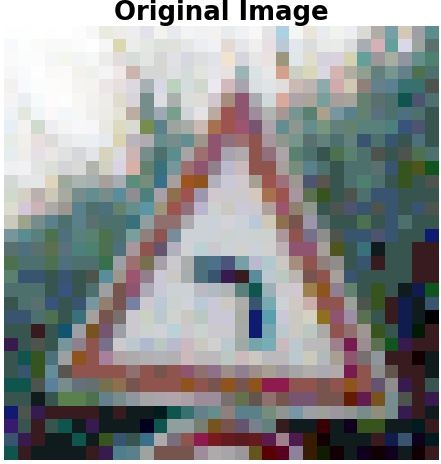} 
			\includegraphics[width=0.48\linewidth,bb=0 0 449 464]{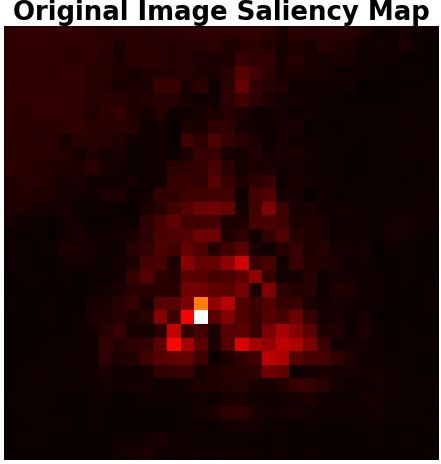} \\
			\includegraphics[width=0.48\linewidth,bb=0 0 449 464]{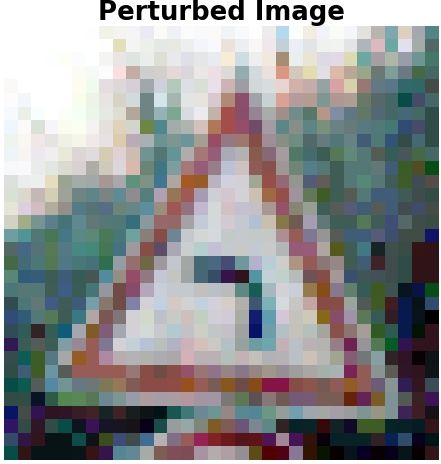} 
			\includegraphics[width=0.48\linewidth,bb=0 0 449 464]{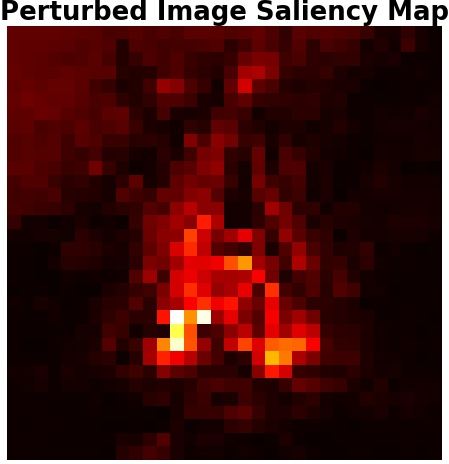} 
			\captionsetup{justification=centering}
			\caption*{Top-100 Intersection: 80.0\% \\ Kendall's Correlation: 0.7886}
		\end{subfigure}
		\caption{For all images, the models give \emph{correct} prediction -- Dangerous Curve to The Left.}
	\end{subfigure}
	\begin{subfigure}{\textwidth}
		\centering
		\begin{subfigure}[b]{.32\textwidth}
			\centering
			\includegraphics[width=0.48\linewidth,bb=0 0 449 464]{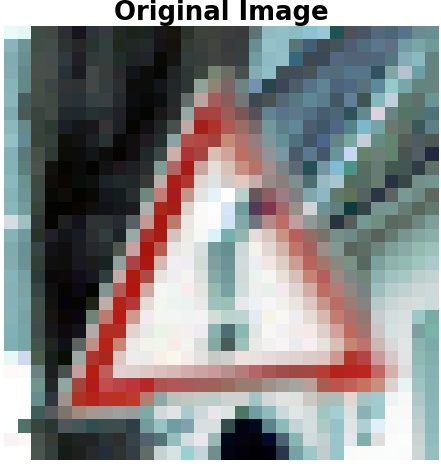} 
			\includegraphics[width=0.48\linewidth,bb=0 0 449 464]{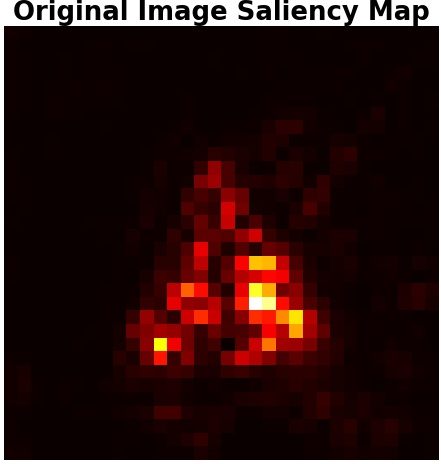} \\
			\includegraphics[width=0.48\linewidth,bb=0 0 449 464]{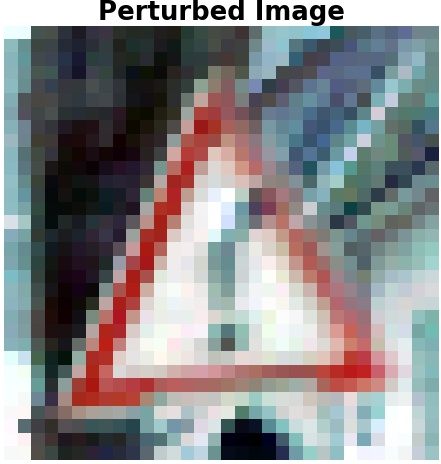} 
			\includegraphics[width=0.48\linewidth,bb=0 0 449 464]{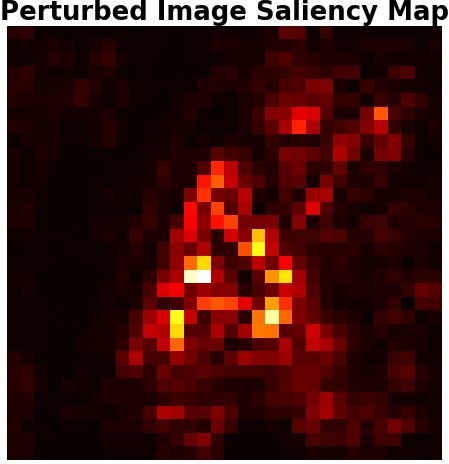} 
			\captionsetup{justification=centering}
			\caption*{Top-100 Intersection: 56.0\% \\  Kendall's Correlation: 0.6679}
		\end{subfigure}
		\begin{subfigure}[b]{.32\textwidth}
			\centering
			\includegraphics[width=0.48\linewidth,bb=0 0 449 464]{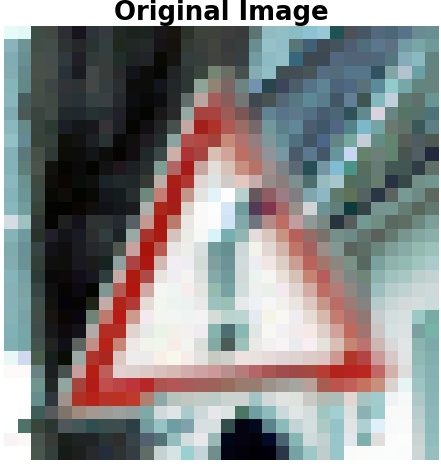} 
			\includegraphics[width=0.48\linewidth,bb=0 0 449 464]{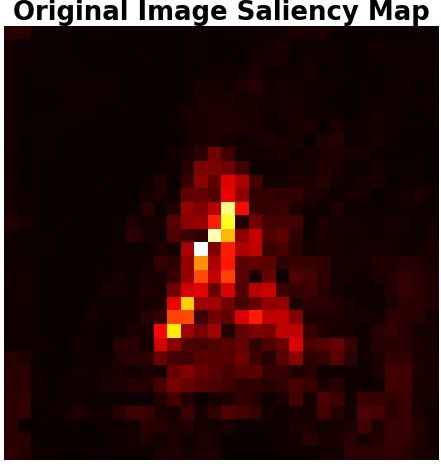} \\
			\includegraphics[width=0.48\linewidth,bb=0 0 449 464]{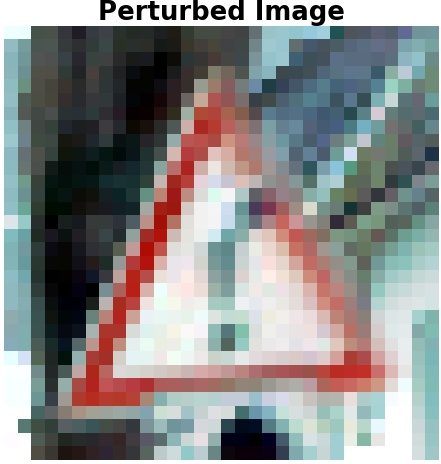} 
			\includegraphics[width=0.48\linewidth,bb=0 0 449 464]{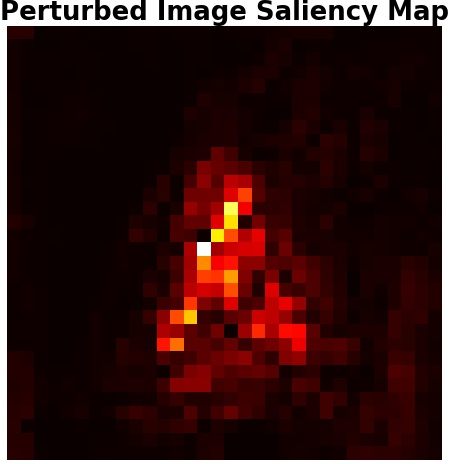} 
			\captionsetup{justification=centering}
			\caption*{Top-100 Intersection: 85.0\% \\ Kendall's Correlation: 0.7963}
		\end{subfigure}
		\begin{subfigure}[b]{.32\textwidth}
			\centering
			\includegraphics[width=0.48\linewidth,bb=0 0 449 464]{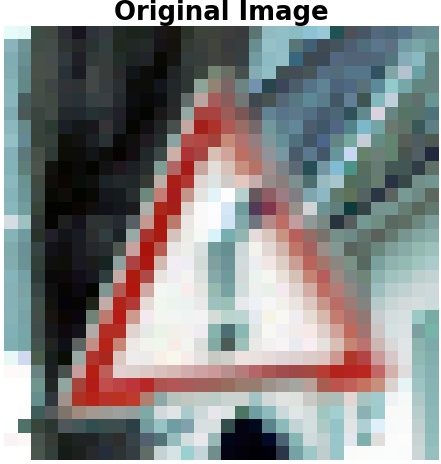} 
			\includegraphics[width=0.48\linewidth,bb=0 0 449 464]{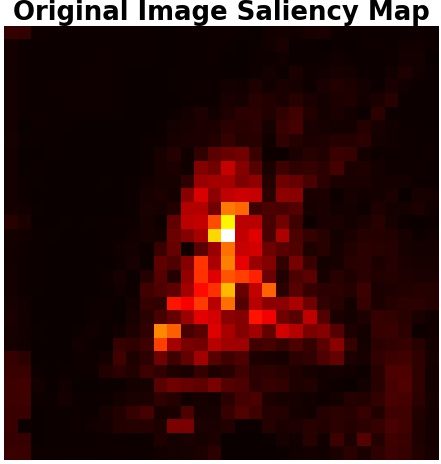} \\
			\includegraphics[width=0.48\linewidth,bb=0 0 449 464]{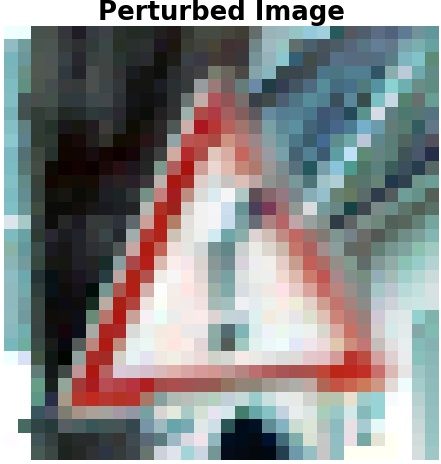} 
			\includegraphics[width=0.48\linewidth,bb=0 0 449 464]{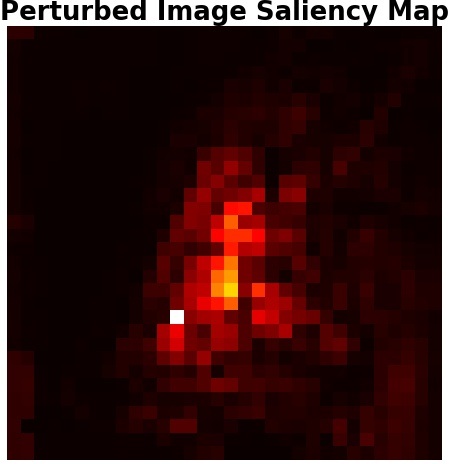} 
			\captionsetup{justification=centering}
			\caption*{Top-100 Intersection: 83.0\% \\ Kendall's Correlation: 0.8338}
		\end{subfigure}
		\caption{For all images, the models give \emph{correct} prediction -- General Caution.}
	\end{subfigure}
	\begin{subfigure}{\textwidth}
		\centering
		\begin{subfigure}[b]{.32\textwidth}
			\centering
			\includegraphics[width=0.48\linewidth,bb=0 0 449 464]{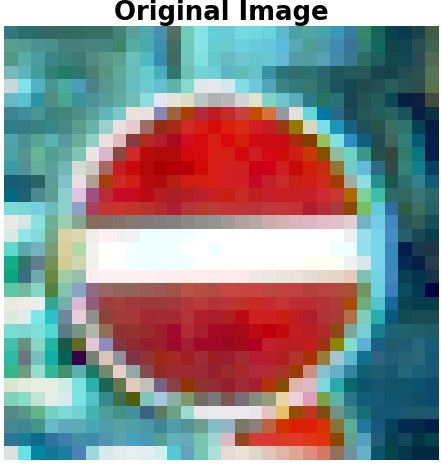} 
			\includegraphics[width=0.48\linewidth,bb=0 0 449 464]{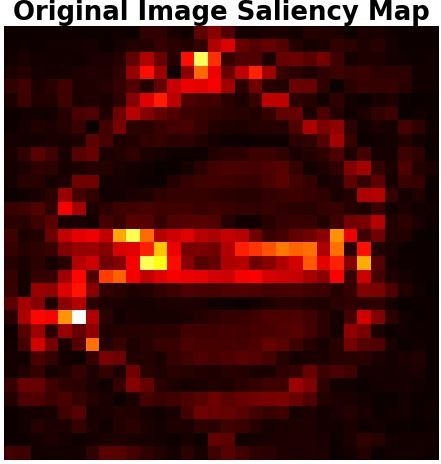} \\
			\includegraphics[width=0.48\linewidth,bb=0 0 449 464]{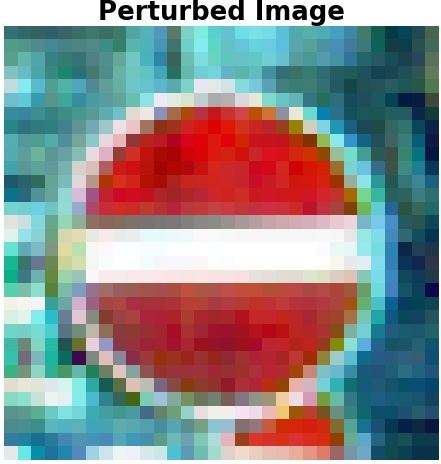} 
			\includegraphics[width=0.48\linewidth,bb=0 0 449 464]{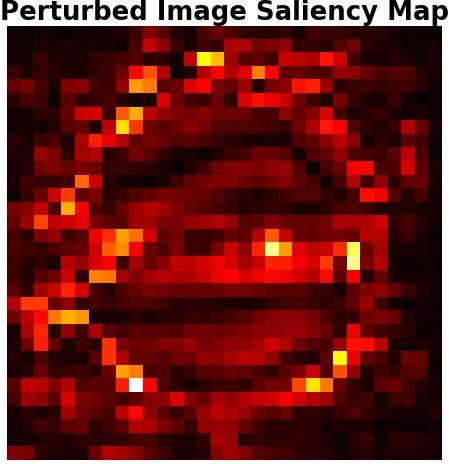} 
			\captionsetup{justification=centering}
			\caption*{Top-100 Intersection: 43.0\% \\  Kendall's Correlation: 0.6160}
		\end{subfigure}
		\begin{subfigure}[b]{.32\textwidth}
			\centering
			\includegraphics[width=0.48\linewidth,bb=0 0 449 464]{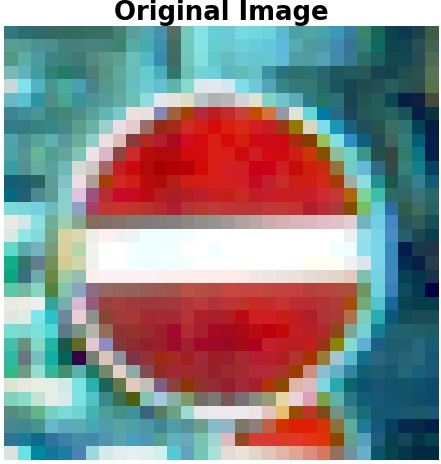} 
			\includegraphics[width=0.48\linewidth,bb=0 0 449 464]{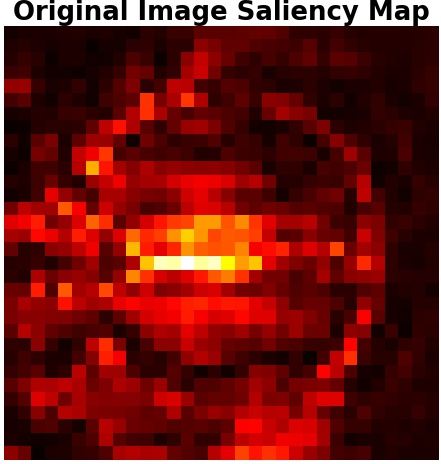} \\
			\includegraphics[width=0.48\linewidth,bb=0 0 449 464]{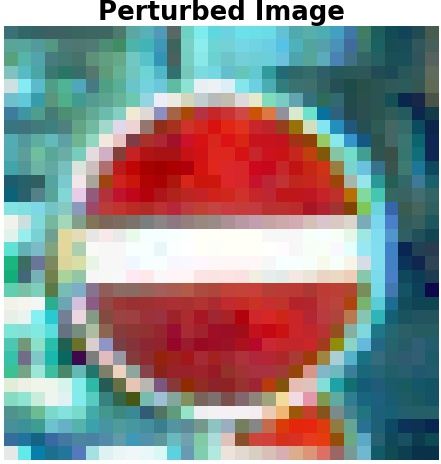} 
			\includegraphics[width=0.48\linewidth,bb=0 0 449 464]{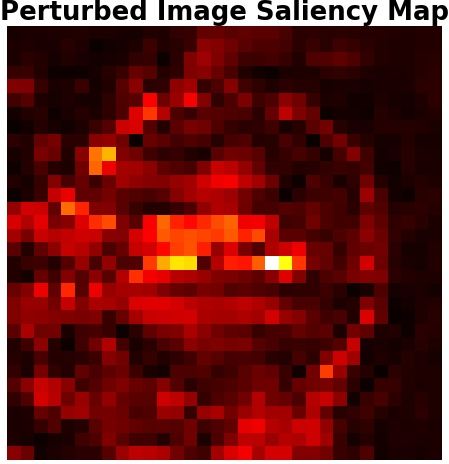} 
			\captionsetup{justification=centering}
			\caption*{Top-100 Intersection: 67.0\% \\ Kendall's Correlation: 0.7595}
		\end{subfigure}
		\begin{subfigure}[b]{.32\textwidth}
			\centering
			\includegraphics[width=0.48\linewidth,bb=0 0 449 464]{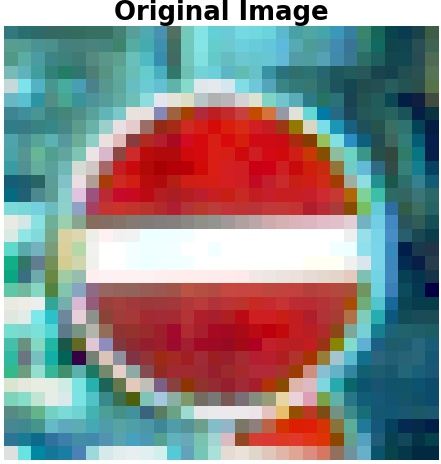} 
			\includegraphics[width=0.48\linewidth,bb=0 0 449 464]{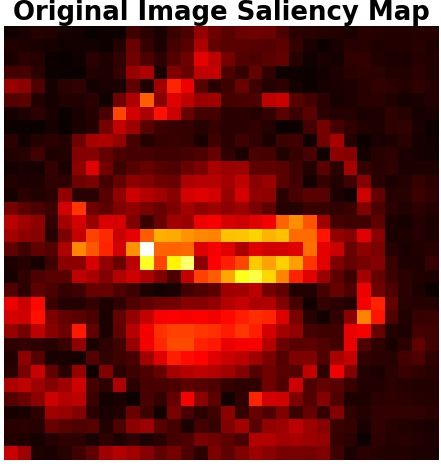} \\
			\includegraphics[width=0.48\linewidth,bb=0 0 449 464]{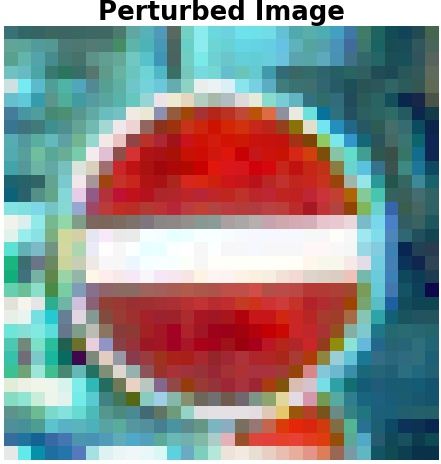} 
			\includegraphics[width=0.48\linewidth,bb=0 0 449 464]{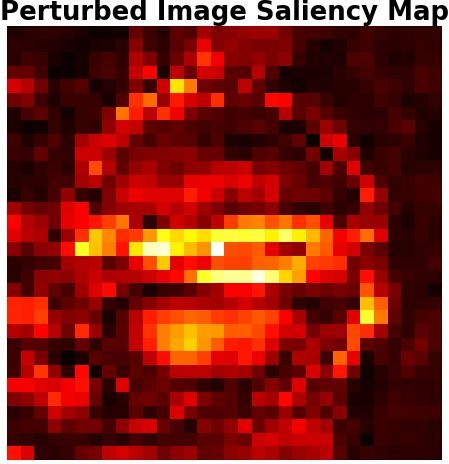} 
			\captionsetup{justification=centering}
			\caption*{Top-100 Intersection: 81.0\% \\ Kendall's Correlation: 0.8128}
		\end{subfigure}
		\caption{For all images, the models give \emph{correct} prediction -- No Entry.}
	\end{subfigure}
	\caption{Top-100 and Kendall's Correlation are rank correlations
		between original and perturbed saliency maps.
		NATURAL is the naturally trained model, IG-NORM and IG-SUM-NORM are models trained using our
		robust attribution method. We use attribution attacks described in~\cite{GAZ17} to perturb
		the attributions while keeping predictions intact. For all images, the models give
		\emph{correct} predictions. However, the saliency maps
		(also called feature importance maps), computed via IG, show that attributions of
		the naturally trained model are very fragile, either visually or quantitatively as measured
		by correlation analyses, while models trained using our method are much more robust
		in their attributions.}
	\label{fig:ex3}
\end{figure}

\begin{figure}[htb]
	\centering 
	\begin{minipage}{\linewidth}
		\centering
		NATURAL \hspace{2.5cm} IG-NORM \hspace{2.5cm} IG-SUM-NORM
	\end{minipage} 
	\begin{subfigure}{\textwidth}
		\centering
		\begin{subfigure}[b]{.32\textwidth}
			\centering
			\includegraphics[width=0.48\linewidth,bb=0 0 449 464]{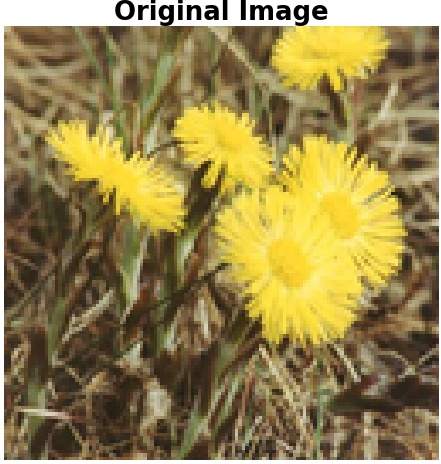} 
			\includegraphics[width=0.48\linewidth,bb=0 0 449 464,bb=0 0 449 464]{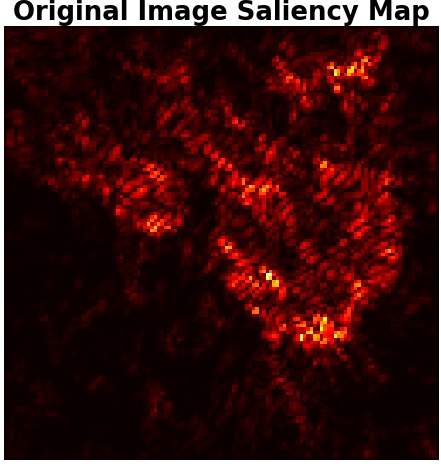} \\
			\includegraphics[width=0.48\linewidth,bb=0 0 449 464]{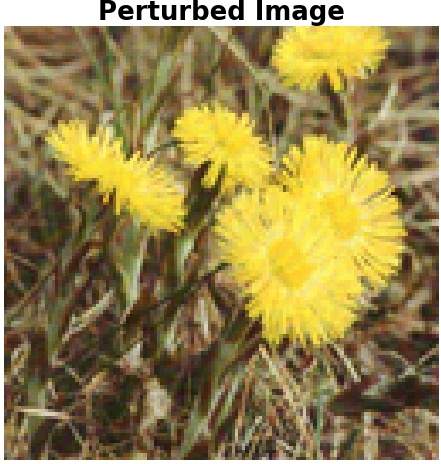} 
			\includegraphics[width=0.48\linewidth,bb=0 0 449 464]{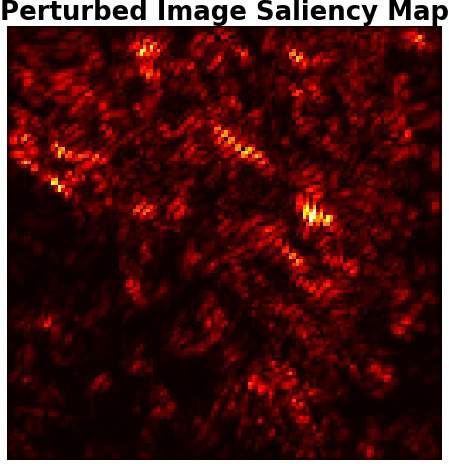} 
			\captionsetup{justification=centering}
			\caption*{Top-1000 Intersection: 1.0\% \\  Kendall's Correlation: 0.4601}
		\end{subfigure}
		\begin{subfigure}[b]{.32\textwidth}
			\centering
			\includegraphics[width=0.48\linewidth,bb=0 0 449 464]{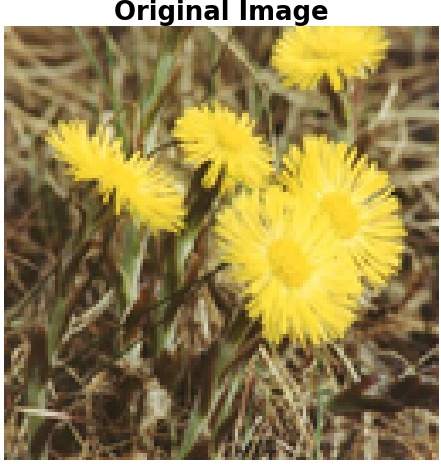} 
			\includegraphics[width=0.48\linewidth,bb=0 0 449 464]{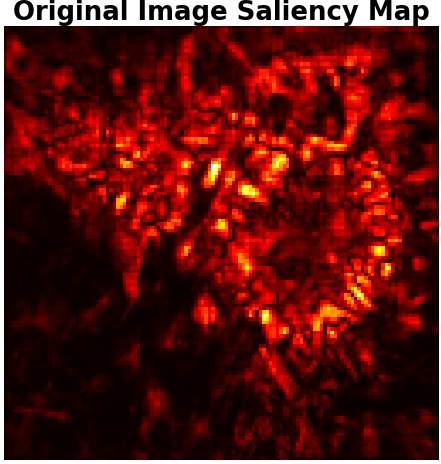} \\
			\includegraphics[width=0.48\linewidth,bb=0 0 449 464]{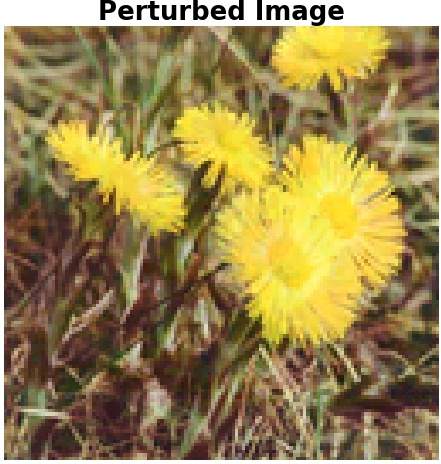} 
			\includegraphics[width=0.48\linewidth,bb=0 0 449 464]{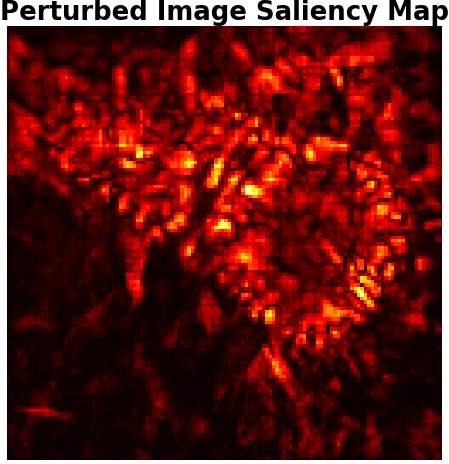} 
			\captionsetup{justification=centering}
			\caption*{Top-1000 Intersection: 65.4\% \\ Kendall's Correlation: 0.7248}
		\end{subfigure}
		\begin{subfigure}[b]{.32\textwidth}
			\centering
			\includegraphics[width=0.48\linewidth,bb=0 0 449 464]{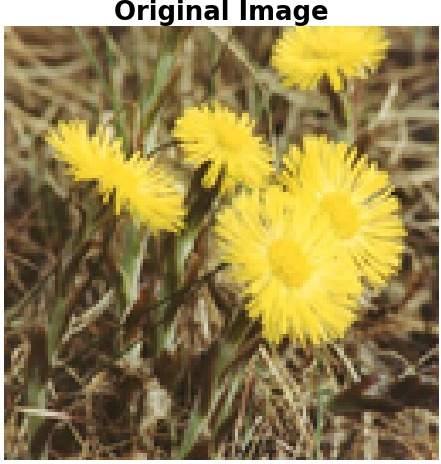} 
			\includegraphics[width=0.48\linewidth,bb=0 0 449 464]{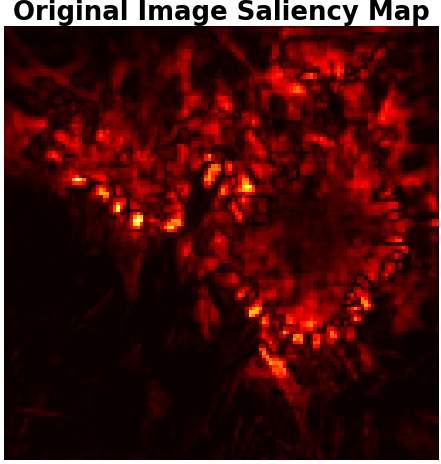} \\
			\includegraphics[width=0.48\linewidth,bb=0 0 449 464]{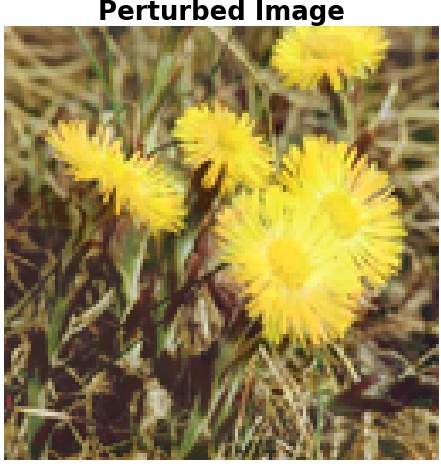} 
			\includegraphics[width=0.48\linewidth,bb=0 0 449 464]{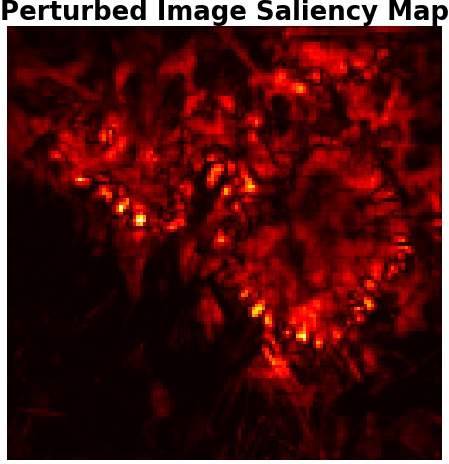} 
			\captionsetup{justification=centering}
			\caption*{Top-1000 Intersection: 63.9\% \\ Kendall's Correlation: 0.8036}
		\end{subfigure}
		\caption{For all images, the models give \emph{correct} prediction -- Bluebell.}
	\end{subfigure}
	\begin{subfigure}{\textwidth}
		\centering
		\begin{subfigure}[b]{.32\textwidth}
			\centering
			\includegraphics[width=0.48\linewidth,bb=0 0 449 464]{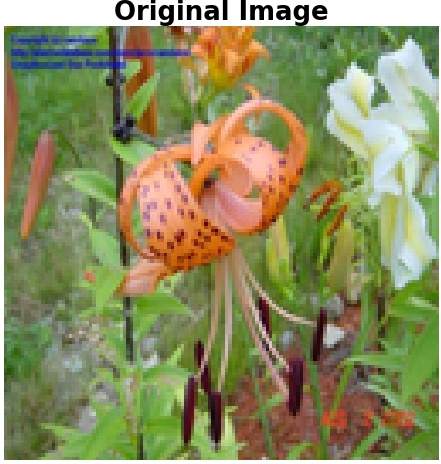} 
			\includegraphics[width=0.48\linewidth,bb=0 0 449 464]{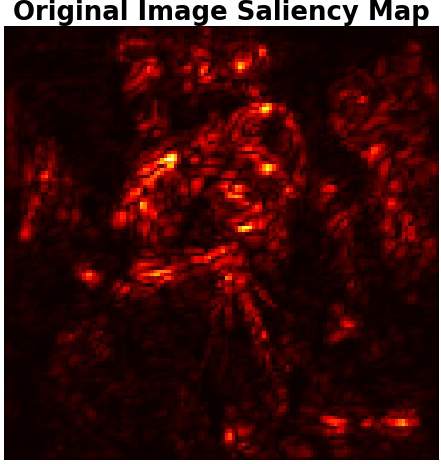} \\
			\includegraphics[width=0.48\linewidth,bb=0 0 449 464]{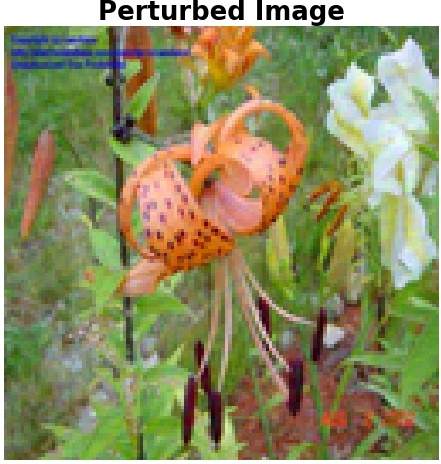} 
			\includegraphics[width=0.48\linewidth,bb=0 0 449 464]{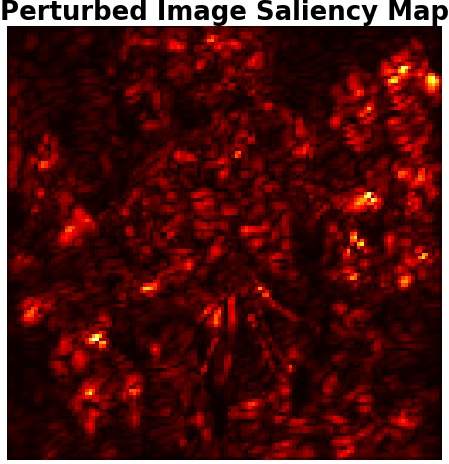} 
			\captionsetup{justification=centering}
			\caption*{Top-1000 Intersection: 6.2\% \\  Kendall's Correlation: 0.3863}
		\end{subfigure}
		\begin{subfigure}[b]{.32\textwidth}
			\centering
			\includegraphics[width=0.48\linewidth,bb=0 0 449 464]{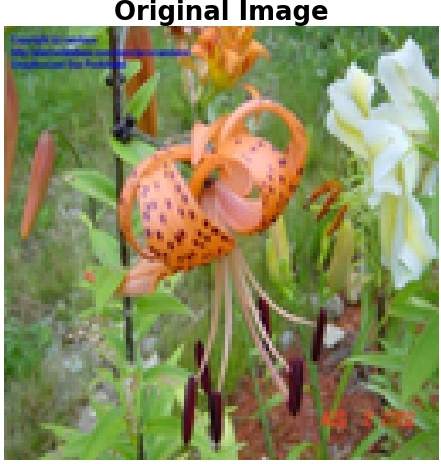} 
			\includegraphics[width=0.48\linewidth,bb=0 0 449 464]{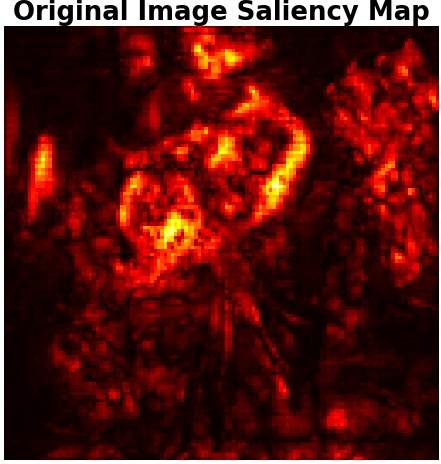} \\
			\includegraphics[width=0.48\linewidth,bb=0 0 449 464]{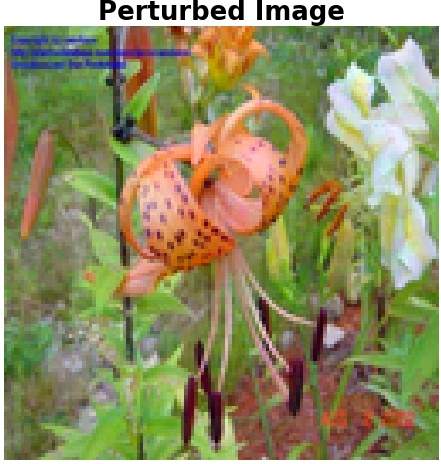} 
			\includegraphics[width=0.48\linewidth,bb=0 0 449 464]{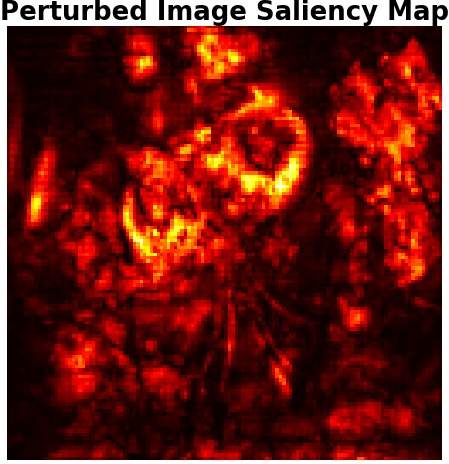} 
			\captionsetup{justification=centering}
			\caption*{Top-1000 Intersection: 58.20\% \\ Kendall's Correlation: 0.6694}
		\end{subfigure}
		\begin{subfigure}[b]{.32\textwidth}
			\centering
			\includegraphics[width=0.48\linewidth,bb=0 0 449 464]{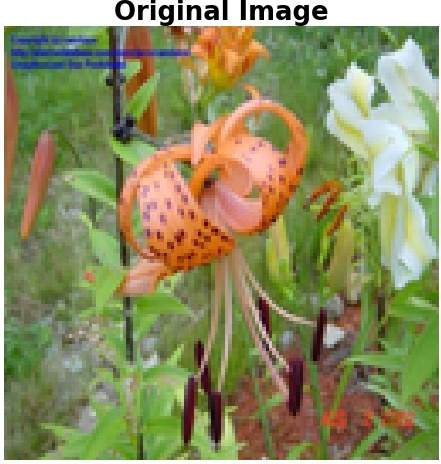} 
			\includegraphics[width=0.48\linewidth,bb=0 0 449 464]{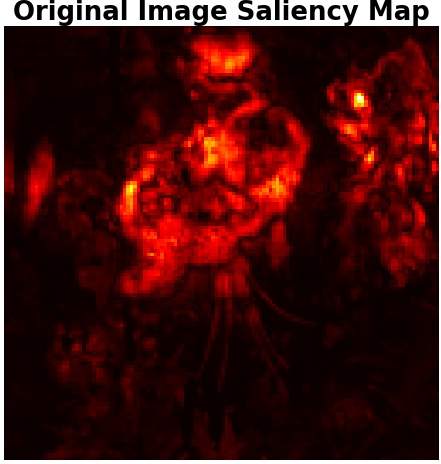} \\
			\includegraphics[width=0.48\linewidth,bb=0 0 449 464]{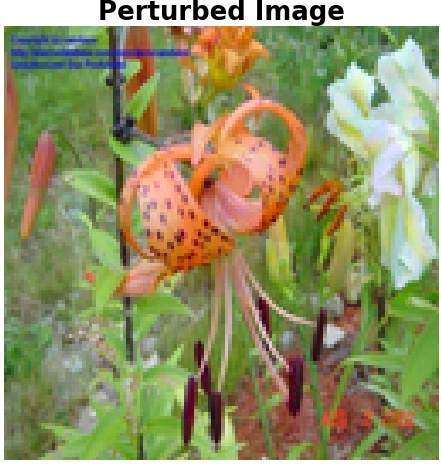} 
			\includegraphics[width=0.48\linewidth,bb=0 0 449 464]{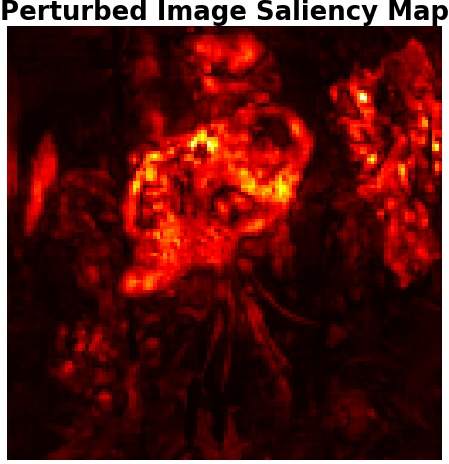} 
			\captionsetup{justification=centering}
			\caption*{Top-1000 Intersection: 65.9\% \\ Kendall's Correlation: 0.7970}
		\end{subfigure}
		\caption{For all images, the models give \emph{correct} prediction -- Cowslip.}
	\end{subfigure}
	\begin{subfigure}{\textwidth}
		\centering
		\begin{subfigure}[b]{.32\textwidth}
			\centering
			\includegraphics[width=0.48\linewidth,bb=0 0 449 464]{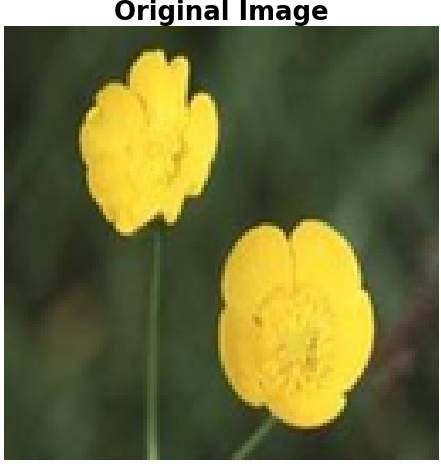} 
			\includegraphics[width=0.48\linewidth,bb=0 0 449 464]{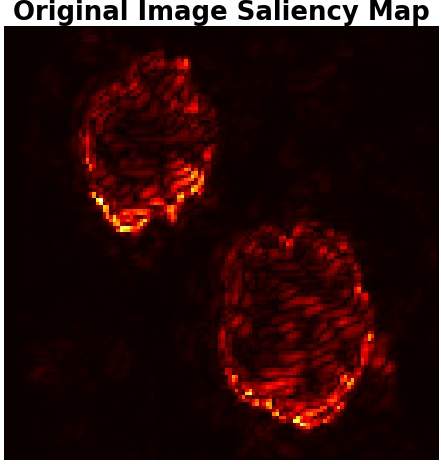} \\
			\includegraphics[width=0.48\linewidth,bb=0 0 449 464]{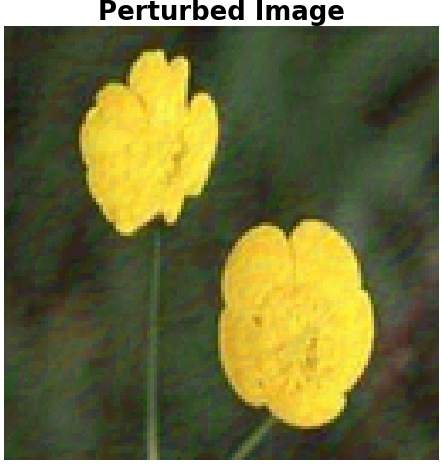} 
			\includegraphics[width=0.48\linewidth,bb=0 0 449 464]{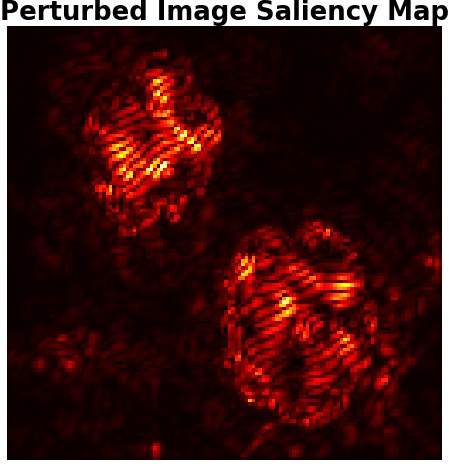} 
			\captionsetup{justification=centering}
			\caption*{Top-1000 Intersection: 6.8\% \\  Kendall's Correlation: 0.4653}
		\end{subfigure}
		\begin{subfigure}[b]{.32\textwidth}
			\centering
			\includegraphics[width=0.48\linewidth,bb=0 0 449 464]{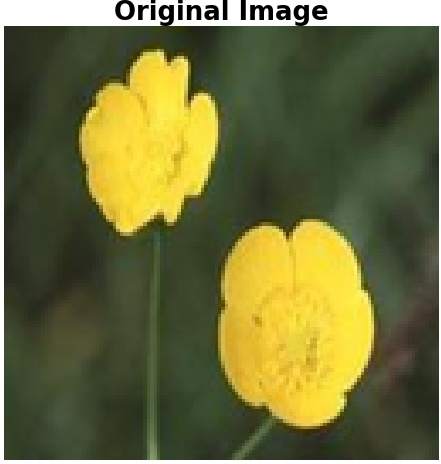} 
			\includegraphics[width=0.48\linewidth,bb=0 0 449 464]{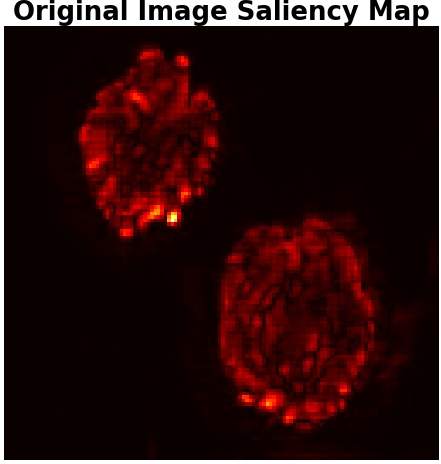} \\
			\includegraphics[width=0.48\linewidth,bb=0 0 449 464]{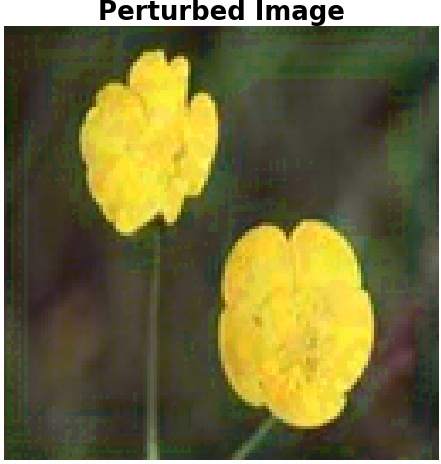} 
			\includegraphics[width=0.48\linewidth,bb=0 0 449 464]{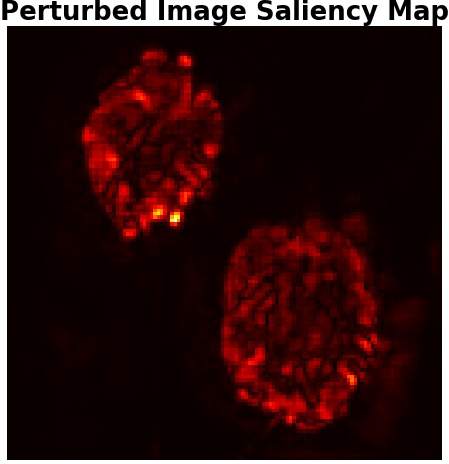} 
			\captionsetup{justification=centering}
			\caption*{Top-1000 Intersection: 58.0\% \\ Kendall's Correlation: 0.7165}
		\end{subfigure}
		\begin{subfigure}[b]{.32\textwidth}
			\centering
			\includegraphics[width=0.48\linewidth,bb=0 0 449 464]{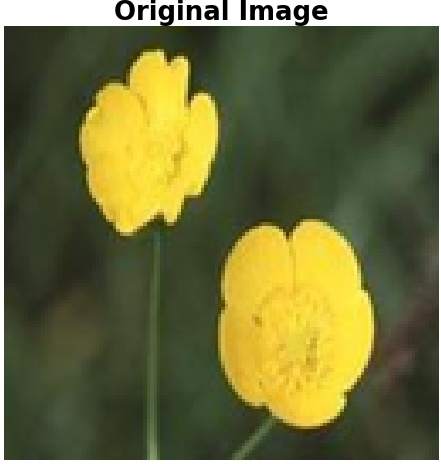} 
			\includegraphics[width=0.48\linewidth,bb=0 0 449 464]{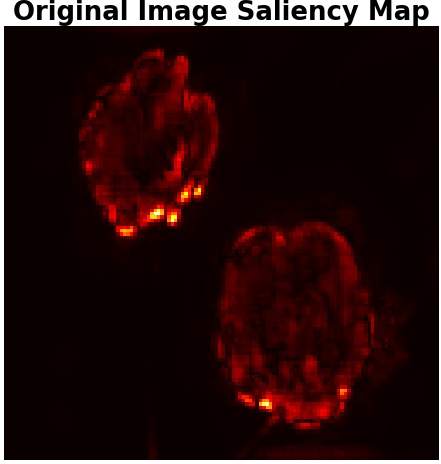} \\
			\includegraphics[width=0.48\linewidth,bb=0 0 449 464]{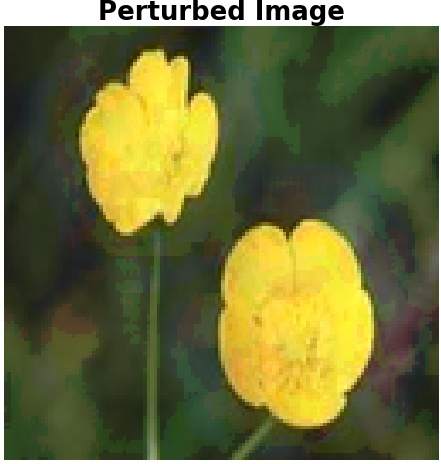} 
			\includegraphics[width=0.48\linewidth,bb=0 0 449 464]{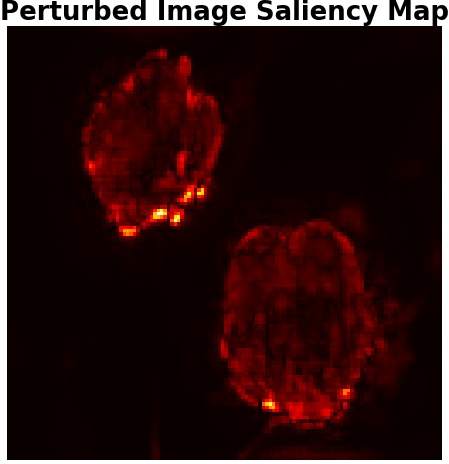} 
			\captionsetup{justification=centering}
			\caption*{Top-1000 Intersection: 63.4\% \\ Kendall's Correlation: 0.8201}
		\end{subfigure}
		\caption{For all images, the models give \emph{correct} prediction -- Tigerlily.}
	\end{subfigure}
	\caption{Top-1000 and Kendall's Correlation are rank correlations
		between original and perturbed saliency maps.
		NATURAL is the naturally trained model, IG-NORM and IG-SUM-NORM are models trained using our
		robust attribution method. We use attribution attacks described in~\cite{GAZ17} to perturb
		the attributions while keeping predictions intact. For all images, the models give
		\emph{correct} predictions. However, the saliency maps
		(also called feature importance maps), computed via IG, show that attributions of
		the naturally trained model are very fragile, either visually or quantitatively as measured
		by correlation analyses, while models trained using our method are much more robust
		in their attributions.}
	\label{fig:ex4}
\end{figure}

Here we provide more visualization results for MNIST in Figure~\ref{fig:ex1}, for Fashion-MNIST in Figure~\ref{fig:ex2}, for GTSRB in Figure~\ref{fig:ex3}, and for Flower in Figure~\ref{fig:ex4}. 

%%% Local Variables:
%%% mode: latex
%%% TeX-master: "make"
%%% End:

\end{document}